\documentclass{article}
\usepackage{graphicx}
\usepackage[margin=1in]{geometry}
\usepackage{authblk}
\usepackage{float}
\usepackage{amsfonts}
\usepackage{amsmath}
\usepackage{amssymb}
\usepackage{amsthm}
\usepackage{booktabs}
\usepackage{makecell}
\usepackage{multirow}
\usepackage{bm}
\usepackage{xcolor}
\usepackage{hyperref}

\title{
A Physics-Conditioned Neural Operator for Generalization of Atrioventricular Valve Mechanics across Pressure and Tissue Properties
}

\newcommand{\bs}[1]{\boldsymbol{#1}}
\newcommand{\pd}[2]{\frac{\partial #1}{\partial #2}}

\author[1,*]{Shawn Koohy}
\author[1,3]{Wensi Wu}
\author[2,3]{Matthew A. Jolley}
\author[1]{Paris Perdikaris}

\affil[1]{Department of Mechanical Engineering and Applied Mechanics, University of Pennsylvania, Philadelphia, PA, USA.}
\affil[2]{Department of Anesthesiology and Critical Care Medicine, Children’s Hospital of Philadelphia, Philadelphia, PA, USA.}
\affil[3]{Division of Pediatric Cardiology, Children’s Hospital of Philadelphia, Philadelphia, PA, USA.}
\affil[*]{Corresponding author: \texttt{skoohy@seas.upenn.edu}}

\date{}

\begin{document}
\maketitle



\newpage
\section{Abstract}

Mitral and tricuspid regurgitation are the most common regurgitant valvular lesions, yet only a minority of severe cases undergo corrective surgery. Rapid assessment of valve mechanics could enable earlier, more precise intervention, but finite element (FE) analysis is slow to repeat across the many loading and tissue-property values of interest, which for a given valve are not known in advance. We introduce the Physics-Conditioned Neural Operator (PCNO), a transformer-based neural operator predicting leaflet displacement, strain, and stress fields conditioned on systolic blood pressure and tissue properties. PCNO is trained on, and evaluated against, FEBio simulations of functional and regurgitant mitral and tricuspid valves and of three mitral pathologies. With pressure and all material parameters simultaneously outside the training support, displacement error against FE reaches 4.48\% and the mean errors of unsupervised geometric measures of valve function stay within 3.5\%, indicating a conditioned solution operator over parameter space rather than an interpolator of the training set. Under this shift, PCNO is also more accurate than graph neural network and graph neural operator baselines trained on the same simulations, with the largest margins in stress.

\newpage

\section{Introduction}

Valvular heart disease affected an estimated 74 million people worldwide in 2019 and ranks among the leading contributors to global cardiovascular mortality \cite{hibino2024valvular,roth2020global,nkomo2006burden}. Among the four cardiac valves, the mitral and tricuspid are the most commonly affected by regurgitant lesions \cite{dziadzko2018outcome}, and the prevalence of both mitral and tricuspid regurgitation rises with age \cite{figlioli2025global,pierard2010ischaemic,topilsky2019burden}. Treatment, however, falls far short of this burden. Only 2.6\% of patients with moderate or severe tricuspid regurgitation ever undergo tricuspid valve surgery \cite{topilsky2019burden}, and the procedure carries an in-hospital mortality reported as high as 24\% \cite{zack2017national}, driven largely by late referral after right heart failure and end-organ damage have already developed \cite{kawsara2021determinants}. Notably, the number of isolated tricuspid valve operations nearly tripled between 2004 and 2013, yet in-hospital mortality remained unchanged at 8.8\% \cite{zack2017national}, suggesting that gains in surgical volume alone are insufficient without earlier detection and better-informed intervention. These limitations motivate the rapid and accurate quantification of atrioventricular valve structure and mechanics pursued in this study.

Understanding these mechanics, and ultimately improving outcomes, has been propelled by advances in cardiac imaging that now allow quantitative, computational evaluation of both valve structure and the mechanical state of the leaflets in living patients. Image-based assessment of valve function has historically advanced along two complementary tracks. Structural assessment, enabled by three-dimensional echocardiography and other tomographic imaging, focuses on extracting geometric measures such as annular height, leaflet billow, tenting depth, coaptation area, and regurgitant orifice area that associate valve morphology with the degree of regurgitation \cite{vergnat2012influence,herz2021segmentation,jassar2014regional}, clinical outcome, and may better inform strategies for valve repair \cite{pingitore20223d,muraru2021right,oguz2019quantitative}. In parallel, finite element (FE) modeling has been developed to quantify the mechanical environment of the leaflets \cite{wu2023effects,wu2022computational}, providing access to stress and strain measures \cite{laurence2024febio,el2021valve,el2021mitral} that are increasingly linked to valve failure \cite{kong2020finite,biffi2019workflow} but cannot be measured in vivo by current clinical imaging. In addition to these, FE modeling has enabled investigations into optimizing patient-specific surgical strategies before they are performed in real-world operations \cite{choi2014novel,choi2017neochordoplasty}; given the number of candidate procedures and the duration of the studies required to compare them, such optimization is unlikely to be achieved through prospective clinical trials alone \cite{sacks2019simulation}. Together, these geometric and mechanical measures provide a quantitative basis for patient-specific assessment of valve function.

However, traditional computational approaches such as the finite element method, while accurate, remain costly for patient-specific use, with individual analyses reported to take from minutes to days \cite{liang2018deep} and the cost compounding in the many-query workloads that parametric analysis and design optimization demand \cite{balu2019deep}. In response to these challenges, machine learning surrogates have emerged as a promising direction for personalized healthcare, disease diagnosis, and clinical decision-making \cite{marchal2025applications,ahsan2022machine,habehh2021machine,aslan2025simulation}. As an example, a deep learning approach for cardiac mechanics achieved prediction times roughly 600 times faster than conventional FE simulation, reducing simulation of a full pressure--volume loop from 5 hours to 30 seconds \cite{motiwale2024neural}. Such speedups have been argued to be particularly valuable in time-sensitive clinical settings, where the delay of conventional analysis can preclude prompt feedback to clinicians \cite{liang2018deep}. These frameworks can more broadly capture complex behaviors of physical and biological systems while providing real-time inference and evaluation at substantially lower cost \cite{sel2024building,sun2020surrogate,du2022deep}.

Transformers \cite{vaswani2017attention} have emerged as a general-purpose architecture across domains as varied as vision, video, audio, and weather prediction \cite{dosovitskiy2020image,liu2021swin,arnab2021vivit,baevski2020wav2vec,bodnar2024aurora}. Their strength lies in the attention mechanism, which captures long-range dependencies and, because it imposes few inductive biases on the input structure, extends naturally across diverse data modalities \cite{jaegle2021perceiver,jaegle2021perceiver2}. Alongside these developments, neural operators \cite{kovachki2023neural} have been introduced as a data-driven approach for learning mappings between function spaces rather than point-wise input--output relationships. Though developed independently, the two are closely related: attention can be viewed as a special case of the integral-kernel layer underlying neural operators \cite{cao2021choose,kovachki2023neural}. Such architectures admit flexible conditioning on auxiliary information, either through spatially varying fields (local conditioning) or through scalar and low-dimensional context variables (global conditioning), the latter typically implemented by modulating intermediate feature statistics \cite{perez2018film,peebles2023scalable} or by cross-attention \cite{wang2024cvit}. The valve mechanics problem considered here is of the second kind: the loading and material state of a given valve is described by a small set of scalars that act on the valve as a whole. Conditioning of this kind enters through the inputs to the model rather than through the training objective \cite{raissi2019physics,li2024physics}.

In this work, we introduce the Physics-Conditioned Neural Operator (PCNO), a neural operator architecture that predicts the loaded mid-systolic configuration and the corresponding leaflet stress and strain fields of atrioventricular valves across a heterogeneous set of geometries, spanning variations in mesh resolution, annular diameter, valve type (mitral and tricuspid), and boundary condition treatment (fixed and prescribed). The model takes nodal coordinates and associated local features as spatially varying input fields, with coarse global geometric features identifying the geometry under consideration. Separately, it is conditioned on the physical parameters that define the problem: peak systolic blood pressure and the material parameters of the leaflet tissue. The ``physics'' in PCNO refers to the physical parameters used to condition the model rather than to physics-based supervision, with the governing equations entering only through the FE solutions on which it is trained. It is therefore distinct from physics-informed neural operators \cite{li2024physics}, which impose the governing equations through partial differential equation (PDE) residuals in the training loss. Generalization is evaluated over this parameter space, on both in-distribution and out-of-distribution (OOD) regimes, the latter testing whether the model has learned a conditioned solution operator rather than interpolating its training set. 

We further evaluate the predicted configurations using established geometric measures of valve function, which are recovered to within 3.5\% mean error under joint OOD shift despite never being supervised during training. Under the same shift, PCNO also achieves lower error than a message-passing graph neural network and a graph neural operator trained on the same simulations on every predicted field, with relative reductions of 3--6\% in displacement, 9--12\% in strain, and 14--32\% in stress, and retains this advantage at a comparable parameter count and less than a fifth the training time. Once trained, PCNO evaluates a query in 6.01~ms on the coarse mesh and 53.38--55.57~ms on the fine mesh on a single GPU, against mean FE run times of 13.59~s and 847.98~s, and we report these per-query costs together with the FE run time required to generate the datasets and training splits (Table~\ref{tab:datagen_cost}). Throughout, FE solutions serve as the ground truth, and all reported accuracy is relative to them. These results establish PCNO as a surrogate whose validity extends beyond the parameter regime it was trained on, toward higher pressure and stiffer tissue, enabling dense exploration of valve mechanics across loading and tissue properties at a fraction of the per-query cost of direct simulation.

\newpage
\section{Results}

\subsection{Problem Setup and Evaluation Framework}

Throughout, the peak systolic blood pressure (SBP) $p$ and the material parameters $c_0$ and $c_1$ are expressed in kPa and $c_2$ is dimensionless, displacements are reported in mm, stresses in kPa, and strains are dimensionless, while heights, distances, and diameters are reported in mm, volumes in mL, the volume-height ratios in mL/mm, and diameter ratios are dimensionless.

We evaluate the proposed PCNO on two FE datasets of atrioventricular valve mechanics that share a common modeling setup. In all simulations, the leaflet tissue is modeled as isotropic and hyperelastic, with the constitutive behavior governed by the Lee--Sacks \cite{lee2014inverse,wu2018anisotropic} strain energy density (SED) in the uncoupled formulation. Let $\boldsymbol{C}$ denote the right Cauchy--Green deformation tensor, $J=\sqrt{\det(\boldsymbol{C})}$ the Jacobian determinant, and $\boldsymbol{C}^*=J^{-2/3}\boldsymbol{C}$ the isochoric right Cauchy--Green deformation tensor. The SED separates into an isochoric term depending only on $\boldsymbol{C}^*$ and a volumetric term depending only on $J$, 
\begin{equation}\label{eq:sed_split}
    \Psi(\boldsymbol{C}) = \Psi^*(\boldsymbol{C}^*) + U(J), 
\end{equation} 
with 
\begin{equation}\label{eq:sed_dev} 
    \Psi^* = \frac{c_0}{2}(I^*_1 - 3) + \frac{c_1}{2}\left(e^{c_2(I^*_1-3)^2} - 1\right) 
\end{equation} 
and 
\begin{equation}\label{eq:sed_vol}
    U(J) = \frac{\kappa}{2}\left(\ln J\right)^2,
\end{equation} 
where $I^*_1=\text{tr}(\boldsymbol{C}^*)$ denotes the first invariant of $\boldsymbol{C}^*$, $(c_0, c_1, c_2)$ are the material parameters, and $\kappa$ is the bulk modulus. The annulus is treated under one of two boundary conditions, fixed or prescribed, where the prescribed case applies a uniform $2$ mm annular displacement directed radially inward rather than a physiologically deformed annular trajectory, and is applied to the mitral and tricuspid geometries alike. The chordae tendineae are modeled as linear truss elements connecting each papillary muscle tip to a set of leaflet insertion points, with the papillary muscle tips fixed in space; both are present in every simulation, and their generation is described in the Methods. The reference configuration is treated as stress-free, so no in vivo prestrain is imposed and no calibration against paired end-diastolic and end-systolic configurations is performed. Each simulation loads the valve from this unloaded reference configuration to a loaded mid-systolic current configuration under an SBP $p$, and records the resulting displacement, strain, and stress fields. A finite bulk modulus was used in all simulations.

Across both datasets, PCNO receives the reference configuration geometry, together with local features derived from it, and the physical conditioning variables $p$ and $(c_0, c_1, c_2)$, and from these jointly predicts the displacement, strain, and stress fields in a single forward pass. No information from the loaded configuration is available to the network at inference; in particular, the systolic coaptation geometry, which is the region least reliably resolved in clinical imaging, is predicted by the model rather than supplied to it. The input features are detailed in the Methods. The chordae tendineae and papillary muscles are present in the simulations that generate the training data, but their connectivity is withheld from the network inputs, so the model must recover their mechanical effect from the reference geometry and conditioning variables alone, matching the clinical setting in which chordal anatomy is not resolved by routine imaging. Predicted displacements are added to the input reference configuration to obtain the predicted current configuration.

The first dataset, referred to as the base dataset, comprises 331{,}777 simulations spanning 11 mitral and 11 tricuspid valve geometries, including both functional and regurgitant valves. One mitral and one tricuspid geometry are meshed at fine resolution, with 11{,}917 and 12{,}436 nodes, and are simulated under the fixed annular boundary condition, while the remaining 10 mitral and 10 tricuspid geometries are meshed at coarse resolution with 1{,}029 nodes and are simulated under both the fixed and the prescribed annular boundary conditions. The SBP $p$ ranges from 2--15\,kPa, equivalently 15--112.5\,mmHg, while the material parameters $c_0$ and $c_1$ range from 0--1000\,kPa and $c_2$ from 0--10. This heterogeneity is what a single PCNO must accommodate, since one model is trained across all geometries, valve types, mesh resolutions, and annular boundary treatments.

The second dataset, referred to as the diseased dataset, contains 41{,}052 simulations of three mitral valve pathologies, tethering, P2 prolapse, and annular dilation, represented by one geometry per pathology. These three geometries are distinct from the 22 geometries in the base dataset. Each is meshed at a single resolution, with 8{,}651, 7{,}440, and 7{,}471 nodes, respectively, and the tethering and prolapse geometries are simulated under the fixed annular boundary condition, while the dilation geometry is simulated under the prescribed condition. The SBP $p$ ranges from 8--15\,kPa, equivalently 60--112.5\,mmHg, while the material parameters $c_0$ and $c_1$ range from 200--800\,kPa and $c_2$ from 2--8, each a subrange of the corresponding base dataset range. On this dataset, the model additionally predicts a disease class label, performing pathology classification alongside the mechanical field regression in the same forward pass. A complete description of both the base and diseased datasets is provided in the Methods.

We derive a set of geometric measures of valve function specific to each pathology, namely tenting height (TH) and tenting volume (TV) for tethering, prolapse height (PH), prolapse volume (PV), and their ratio for P2 prolapse, and anteroposterior (AP) diameter, intercommissural (IC) diameter, and their ratio for annular dilation. Each measure is computed by applying the same functional to the ground truth current configuration and to the predicted current configuration. Since none of the measures are supervised during training, they serve as an independent check on the predicted configuration against the FE solution. 

\begin{figure}[H]
    \centering
    \includegraphics[width=1\linewidth]{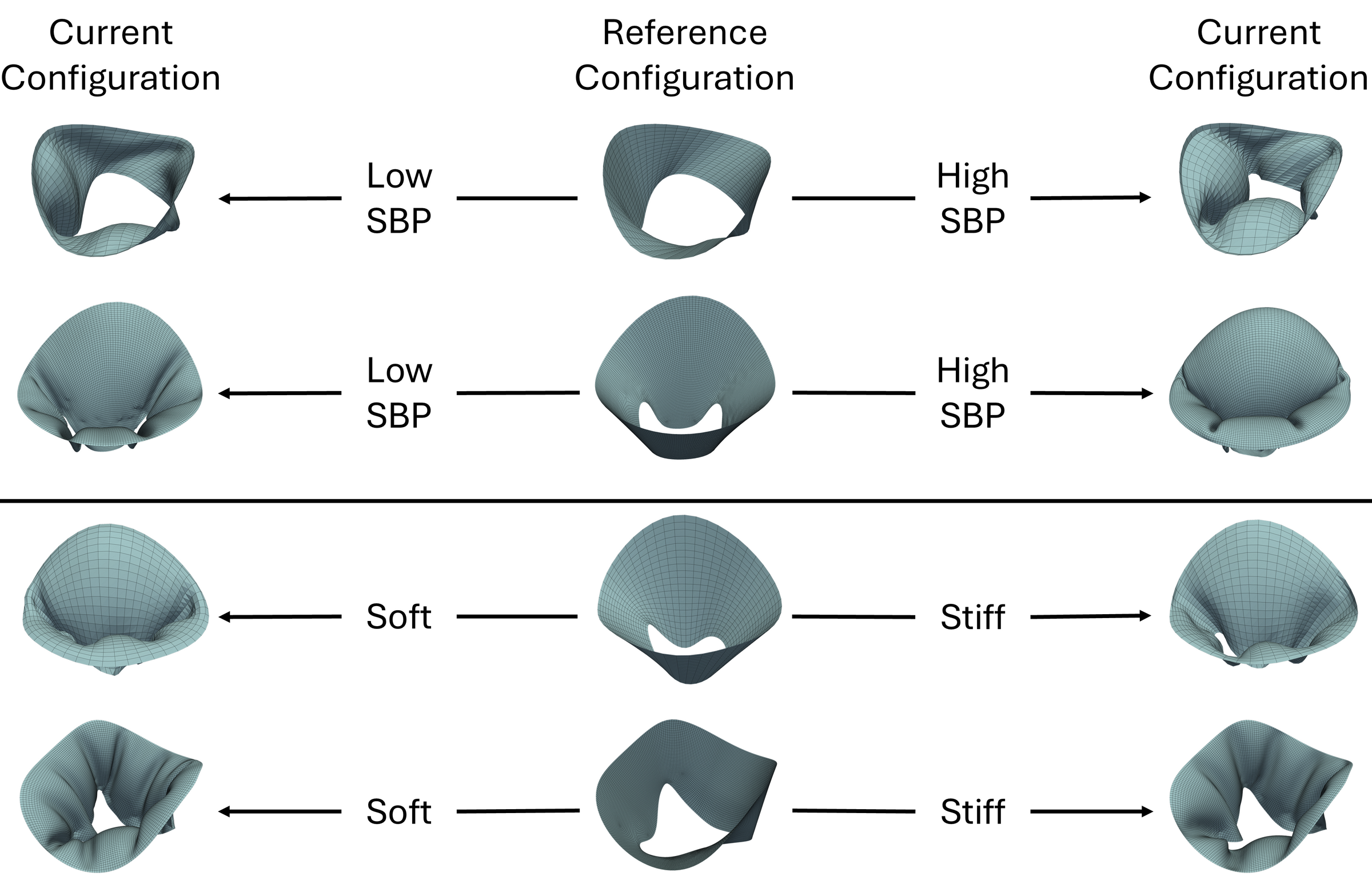}
    \caption{\textbf{Top:} Valve deformation under varying SBP with the same material parameters. \textbf{Bottom:} Valve deformation under varying material parameters with the same SBP.}
    \label{fig:sbp_stiff}
\end{figure}

\begin{figure}[H]
    \centering
    \includegraphics[width=1\linewidth]{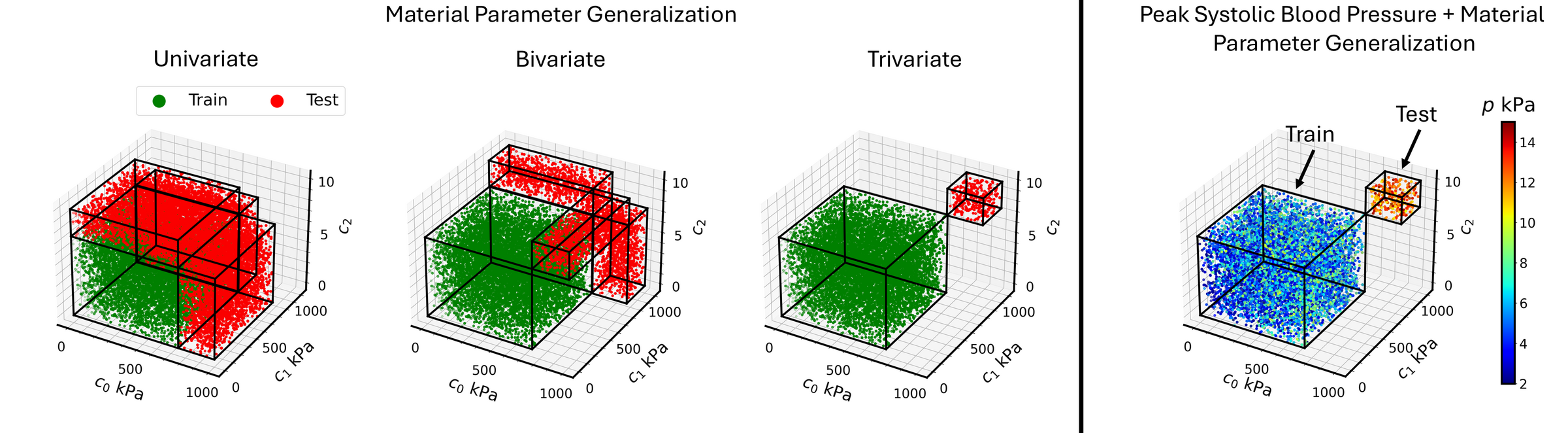}
    \caption{\textbf{Left:} Material parameter generalization with univariate, bivariate, and trivariate splits of increasing difficulty. \textbf{Right:} Joint SBP and material parameter generalization, where both the SBP and material parameters lie outside the training support simultaneously. Bounding boxes between the train and test sets are shown for clarity.} 
    \label{fig:parameter_spaces}
\end{figure}

We assess PCNO under both in-distribution and OOD testing regimes. The physical conditioning variables are not known in advance for a given valve, since the SBP is set by the condition being modeled and the Lee--Sacks parameters are obtained by inverse estimation \cite{ross2024bayesian,lee2014inverse}, so the values a user later wants to query are selected after the training set has been fixed and are not guaranteed to fall inside it. Characterizing how accuracy degrades outside the training range is therefore as informative as in-distribution accuracy. OOD settings are constructed by withholding specific parameter regions during training and evaluating the model on these held-out regions at test time, and a visual overview of the parameter spaces is provided in Fig.~\ref{fig:parameter_spaces}. No geometry is held out, so every test geometry also appears in training, and the results do not test generalization to unseen geometries. The cut-offs separating training from held-out values are 10 kPa for $p$ and 750 kPa, 750 kPa, and 7.5 for $c_0$, $c_1$, and $c_2$ on the base dataset, and 13 kPa, 620 kPa, 640 kPa, and 6.2 on the diseased dataset. Fig.~\ref{fig:sbp_stiff} shows ground-truth geometry configurations spanning a wide range of SBPs and soft to stiff materials, illustrating the qualitative differences in deformation leading to intrinsic variability of the mechanical responses that make these generalization tasks challenging. Our evaluation is organized in order of increasing difficulty: in-distribution testing on the base dataset, single-axis OOD generalization with respect to the SBP and then to the material parameters under univariate (Uni), bivariate (Bi), and trivariate (Tri) distribution shifts, and a joint SBP and trivariate material parameter setting, the most demanding regime on the base dataset. We then turn to the diseased dataset and test the easiest and most challenging of these settings, in-distribution prediction and joint generalization, evaluating not only the mechanical fields but also the recovery of pathology-specific geometric measures of valve function and disease classification.

Model performance on the mechanical fields is evaluated using a relative $L^2$ error, computed over all spatial nodes and averaged across output channels $D$
\begin{equation}
  \text{Relative } L^2(\square)
    = 100 \times \frac{1}{D}\sum_{i=1}^{D}
      \frac{\|\square_i - \tilde{\square}|_i\|_2}%
           {\|\square_i\|_2},
  \qquad \square \in \{\boldsymbol{u},\,\boldsymbol{E},\,\boldsymbol{\sigma}\},
  \label{eq:rel_l2}
\end{equation}
where $\tilde{\square}$ denotes the predicted field and $\square$ the corresponding ground truth. We denote displacements in the reference configuration by $\boldsymbol{u}$, the Cauchy stress tensor by $\boldsymbol{\sigma}$, and the Lagrangian strain tensor by $\boldsymbol{E}$. While the model jointly predicts all unique components of the Cauchy stress tensor $\boldsymbol{\sigma}$ and the Lagrangian strain tensor $\boldsymbol{E}$, we only visualize the principal stress and principal strain fields for clarity. These are obtained as the eigenvalues of the respective tensors, ordered as $\sigma_1 \geq \sigma_2 \geq \sigma_3$ for the principal stresses and $E_1 \geq E_2 \geq E_3$ for the principal strains. All quantitative error metrics are computed over the full tensor fields, not the principal values alone. 

\begin{figure}[H]
    \centering
    \includegraphics[width=1\linewidth]{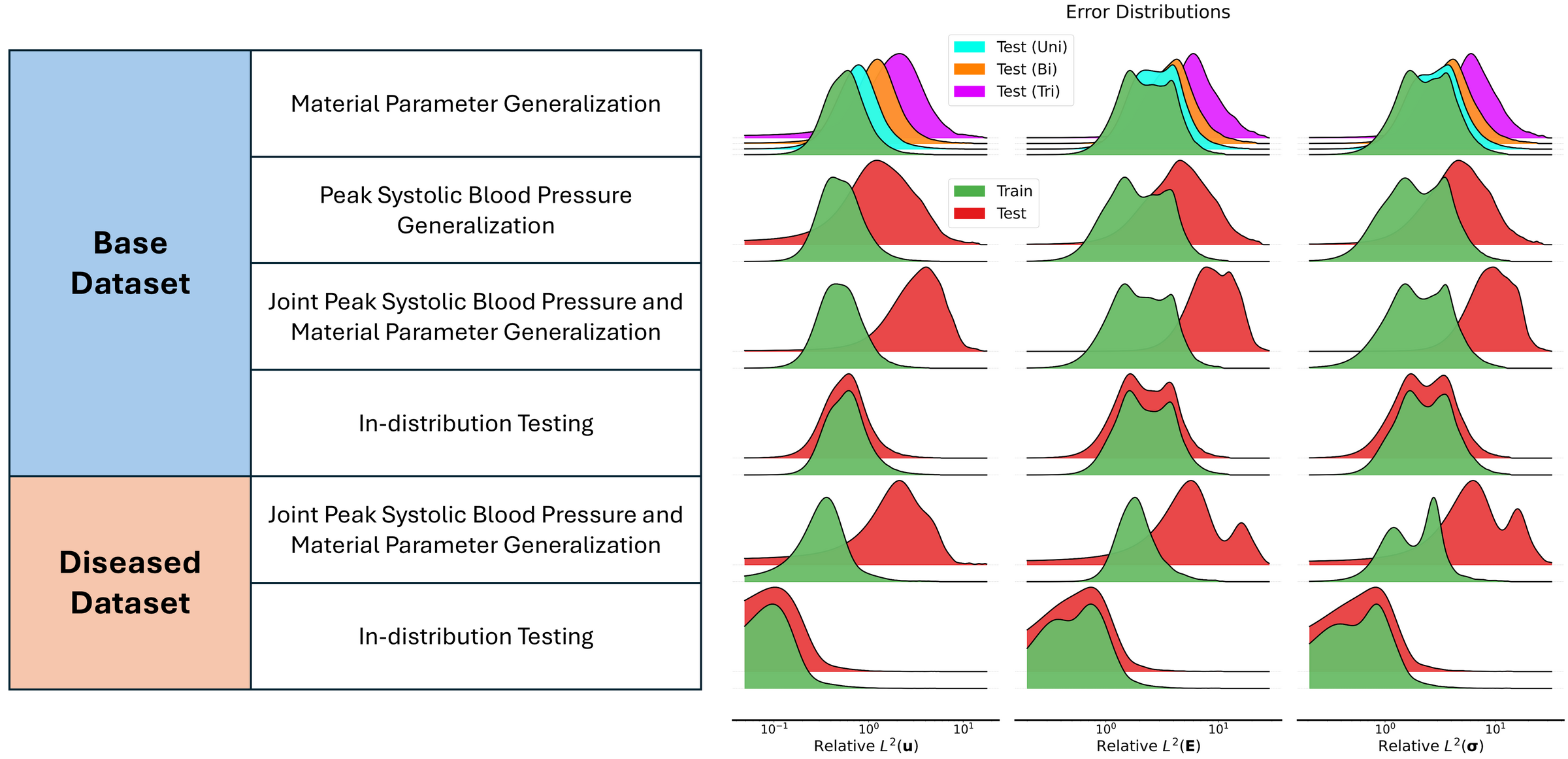}
    \caption{Distribution of relative $L^2$ errors (\%) across generalization settings on the base and diseased datasets. Material parameter generalization cases (Uni, Bi, Tri) share identical training data.}
    \label{fig:results_summary}
\end{figure}

For the geometric measures on the diseased dataset, which are scalar quantities, performance is reported using the relative error $100 \times |m - \tilde{m}| / |m|$, where $m$ and $\tilde{m}$ denote the ground-truth and predicted measure values, respectively. These functionals are nonlinear, and the relative $L^2$ error used for the mechanical fields averages over all mesh nodes. It is therefore insensitive both to error concentrated at the single extremal node that determines TH or PH and to a small systematic tilt of the fitted annular surface against which TV and PV are integrated, and the ratios inherit the sensitivity of both of their constituents. Agreement on the geometric measures is accordingly not implied by a low relative $L^2$ error, and the two are reported separately. Results over each generalization split are summarized in Table~\ref{tab:results_summary} and their distributions visualized in Fig.~\ref{fig:results_summary}. All reported errors, including these distributions and the errors of the geometric measures and their confidence intervals, are computed after excluding the samples whose error lies at or above the 99th percentile of their split, to limit the influence of a small number of samples with extreme localized stress and strain concentrations. The rule is applied identically to every field, setting, and model, and untrimmed mean errors for the displacement, strain, and stress fields are reported in Supplementary Table~6.

Each generalization setting is presented with the same set of figures: a representative test case in the main text, comparing the ground truth and predicted current configuration, stress, and strain fields; a supplementary figure expanding it into the individual stress and strain components with their point-wise absolute and relative $L^2$ errors; and a supplementary figure reporting relative errors as functions of the SBP and material parameters. The representative case is selected by a fixed rule rather than by inspection. All test samples are ranked by relative $L^2$ error separately for displacement, strain, and stress, and we select the sample whose largest distance from the median rank across the three fields is smallest, so that the case shown lies near the middle of the error distribution in every field at once. The resulting train and test splits are given in Supplementary Table~1, and the count of each geometry is given in Supplementary Table~5.

\begin{table}[H]
\centering
\begin{tabular}{lllccc}
\toprule
\textbf{Dataset} & \textbf{Setting} & \textbf{Split} & Rel. $L^2(\boldsymbol{u})$ & Rel. $L^2(\boldsymbol{E})$ & Rel. $L^2(\boldsymbol{\sigma})$ \\
\midrule
\multirow{11}{*}{Base}
 & \multirow{2}{*}{In-distribution}
   & Train & 0.92 & 3.45 & 3.65 \\
 & & Test  & 0.93 & 3.48 & 3.69 \\
\cmidrule(lr){2-6}
 & \multirow{2}{*}{SBP Generalization}
   & Train & 0.84 & 3.26 & 3.49 \\
 & & Test  & 3.17 & 7.65 & 7.76 \\
\cmidrule(lr){2-6}
 & \multirow{4}{*}{\makecell[l]{Material Parameter \\ Generalization}}
   & Train      & 0.88 & 3.43 & 3.68 \\
 & & Test (Uni) & 1.35 & 4.31 & 4.47 \\
 & & Test (Bi)  & 2.08 & 5.68 & 5.83 \\
 & & Test (Tri) & 3.56 & 8.88 & 8.83 \\
\cmidrule(lr){2-6}
 & \multirow{2}{*}{\makecell[l]{Joint SBP and Material \\ Parameter Generalization}}
   & Train & 1.00 & 3.65 & 3.91 \\
 & & Test  & 4.48 & 10.11 & 11.56 \\
\midrule
\multirow{4}{*}{Diseased}
 & \multirow{2}{*}{In-distribution}
   & Train & 0.21 & 0.95 & 1.09 \\
 & & Test  & 0.22 & 1.00 & 1.15 \\
\cmidrule(lr){2-6}
 & \multirow{2}{*}{\makecell[l]{Joint SBP and Material \\ Parameter Generalization}}
   & Train & 0.66 & 2.66 & 2.86 \\
 & & Test  & 3.57 & 10.92 & 12.06 \\
\bottomrule
\end{tabular}
\caption{Average relative $L^2$ errors (\%) across generalization settings on the base and diseased datasets. Material parameter generalization cases (Uni, Bi, Tri) share identical training data.}
\label{tab:results_summary}
\end{table}

\newpage

\subsection{In-distribution}

Accurate prediction within the training parameter range is a prerequisite for the OOD evaluations that follow, since a model that cannot interpolate reliably gives no baseline against which degradation outside the training range can be measured. Here the training and test sets are strictly disjoint, and both cover the full ranges of SBP and material parameters considered in this study. Interpolation is not trivial in this regime as the governing equations involve nonlinear hyperelastic constitutive laws and large deformations, so the map from input parameters to spatially resolved displacement, strain, and stress fields is correspondingly nonlinear.

\begin{figure}[H]
    \centering
    \includegraphics[width=1\linewidth]{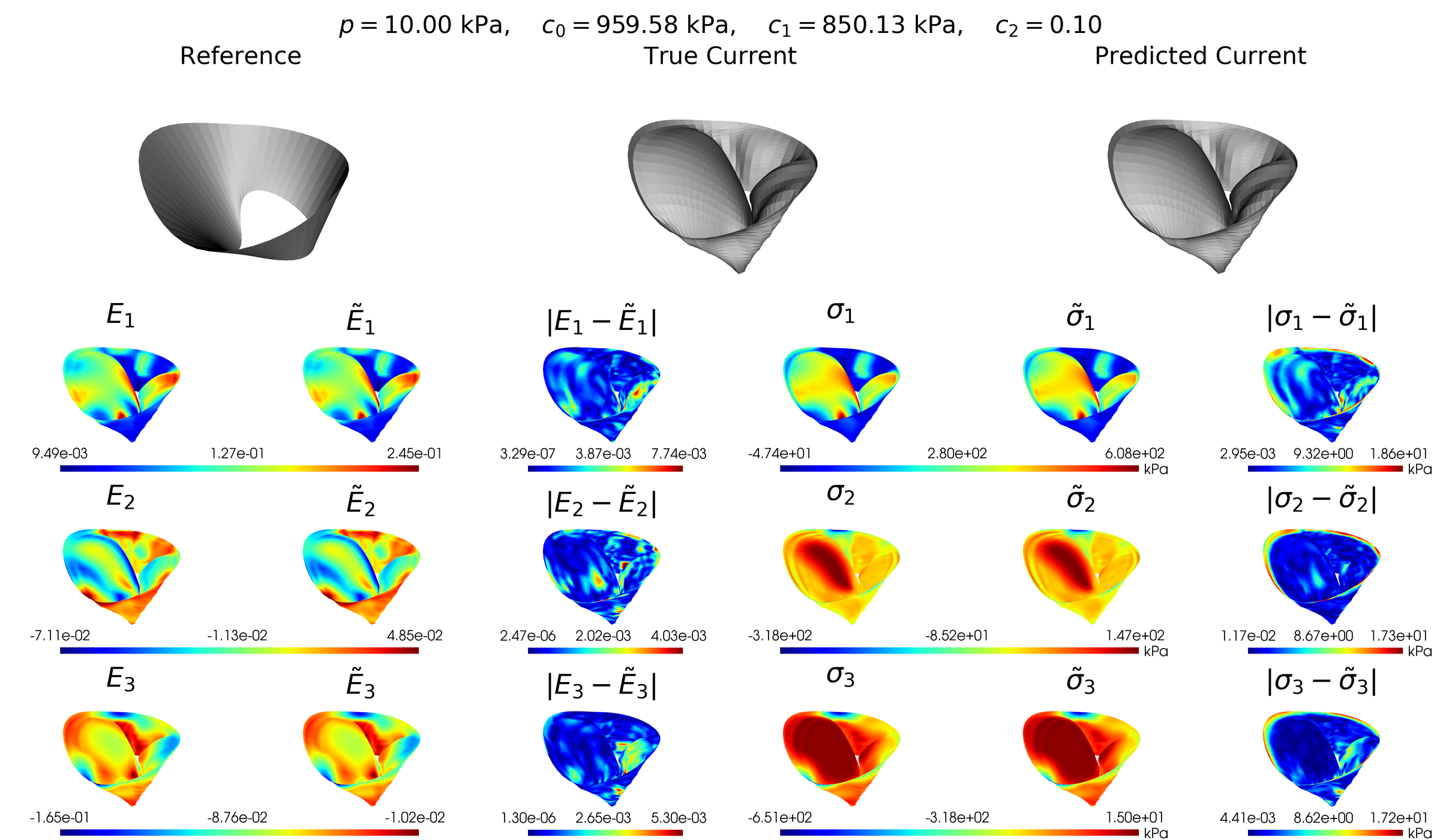}
    \caption{Test example from the in-distribution setting corresponding to a mitral valve.}
    \label{fig:id_example}
\end{figure}

Training and test errors closely align for all predicted quantities, indicating a smooth and stable mapping across the parameter space. Displacement predictions achieve 0.93\% mean relative error, while strain reaches 3.48\% and stress 3.69\%, reflecting their greater sensitivity to local geometric and material variations (Fig.~\ref{fig:id_example}, Supplementary Fig.~1). This is consistent with the nonlinear dependence of strain and stress on spatial gradients of displacement, resulting in these higher-order fields being more challenging to learn.

Errors remain approximately uniform across the full range of SBP and material parameter values, with modest increases near the bounds of the parameter ranges (Supplementary Fig.~6). These near-boundary increases are expected: at low SBP, the current configuration is close to the reference state, amplifying relative error denominators, while at the upper extremes, the mechanical response enters a more strongly nonlinear regime. The absence of sharp error gradients or localized failure regions across the parameter space suggests that the conditioning mechanism enables the model to adjust smoothly to changes in loading and material properties. Consistent performance is observed across the diversity of valve geometries, mesh resolutions, and boundary conditions included in the dataset, demonstrating that the model accommodates geometric and boundary variability within a unified predictive framework.

\subsection{Peak Systolic Blood Pressure Generalization}

The SBP $p$ sets the magnitude of the pressure load applied to the closed leaflets during systole and is therefore the conditioning variable tied most directly to the applied loading. Its value varies substantially across individuals and hemodynamic states, and elevated systolic pressure has been associated with mitral regurgitation \cite{katsi2019role,rahimi2017elevated}. Within our simulations, variation in SBP produces spatially heterogeneous deformation and stress responses whose structure depends nontrivially on geometry and material stiffness. Extrapolating beyond the SBP values observed during training therefore cannot be treated as a simple scaling problem and requires the model to capture how loading interacts with geometry and material stiffness. To assess this, the model is trained on low SBPs and tested on higher, unseen SBPs, and the Methods give full details of the split.

\begin{figure}[H]
    \centering
    \includegraphics[width=1\linewidth]{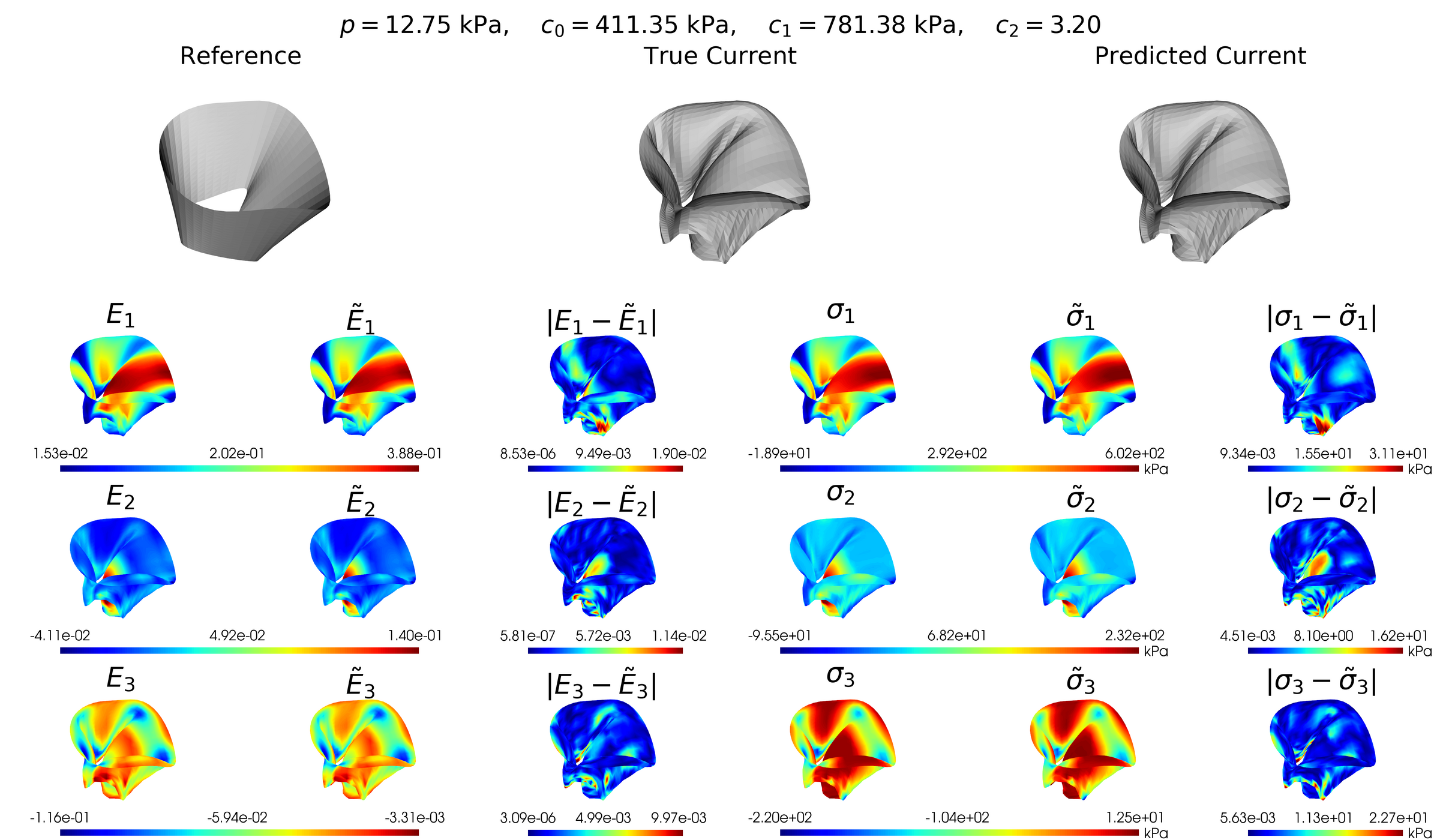}
    \caption{Test example from the SBP generalization setting corresponding to a tricuspid valve.}
    \label{fig:sbp_example}
\end{figure}

Test errors increase relative to the in-distribution setting, though the degradation is moderate: displacement predictions achieve 3.17\% mean relative error, while strain reaches 7.65\% and stress 7.76\%. Notably, the model captures regions of localized stress and strain concentrations in tricuspid valve examples, even in the OOD setting (Fig.~\ref{fig:sbp_example}, Supplementary Fig.~2).

Errors remain approximately uniform across the training range and increase smoothly beyond the training boundary (Supplementary Fig.~7). This suggests that the model retains robustness under moderate OOD shifts while still reflecting the increasing difficulty of the mechanical response at higher loading levels.

Having established that the model extrapolates reliably along the loading axis toward higher SBP, we next examine generalization with respect to the material parameters, where the constitutive response introduces a qualitatively different source of nonlinearity.

\newpage
\subsection{Material Parameter Generalization}
The Lee--Sacks material parameters $(c_0,c_1,c_2)$ enter the SED $\Psi^*$ in Eq.~\ref{eq:sed_dev} in a nonlinear and partially coupled fashion, where $c_0$ contributes a neo-Hookean term while the pair $(c_1,c_2)$ jointly controls both the magnitude and the onset of nonlinear stiffening through the exponential term. Consequently, the resulting mechanical response is governed by joint variations in $(c_0,c_1,c_2)$ rather than by independent effects of each parameter in isolation. Because these parameters are not measured directly but recovered by inverse estimation \cite{lee2014inverse,ross2024bayesian}, a surrogate conditioned on them will be queried across wide regions of the material parameter space, and its accuracy away from the values seen during training determines how far that space can be explored.

We construct three OOD test subsets of increasing difficulty by withholding progressively more parameters from their training ranges: a univariate split in which exactly one parameter lies outside its training range, a bivariate split in which exactly two are outside, and a trivariate split in which all three are simultaneously OOD. In each subset, the model is trained on softer materials and tested on stiffer ones. All three experiments share the same training set; only the test partition changes. Formal definitions of the parameter domain, training ranges, and OOD subsets are provided in the Methods. 

\begin{figure}[H]
    \centering
    \includegraphics[width=1\linewidth]{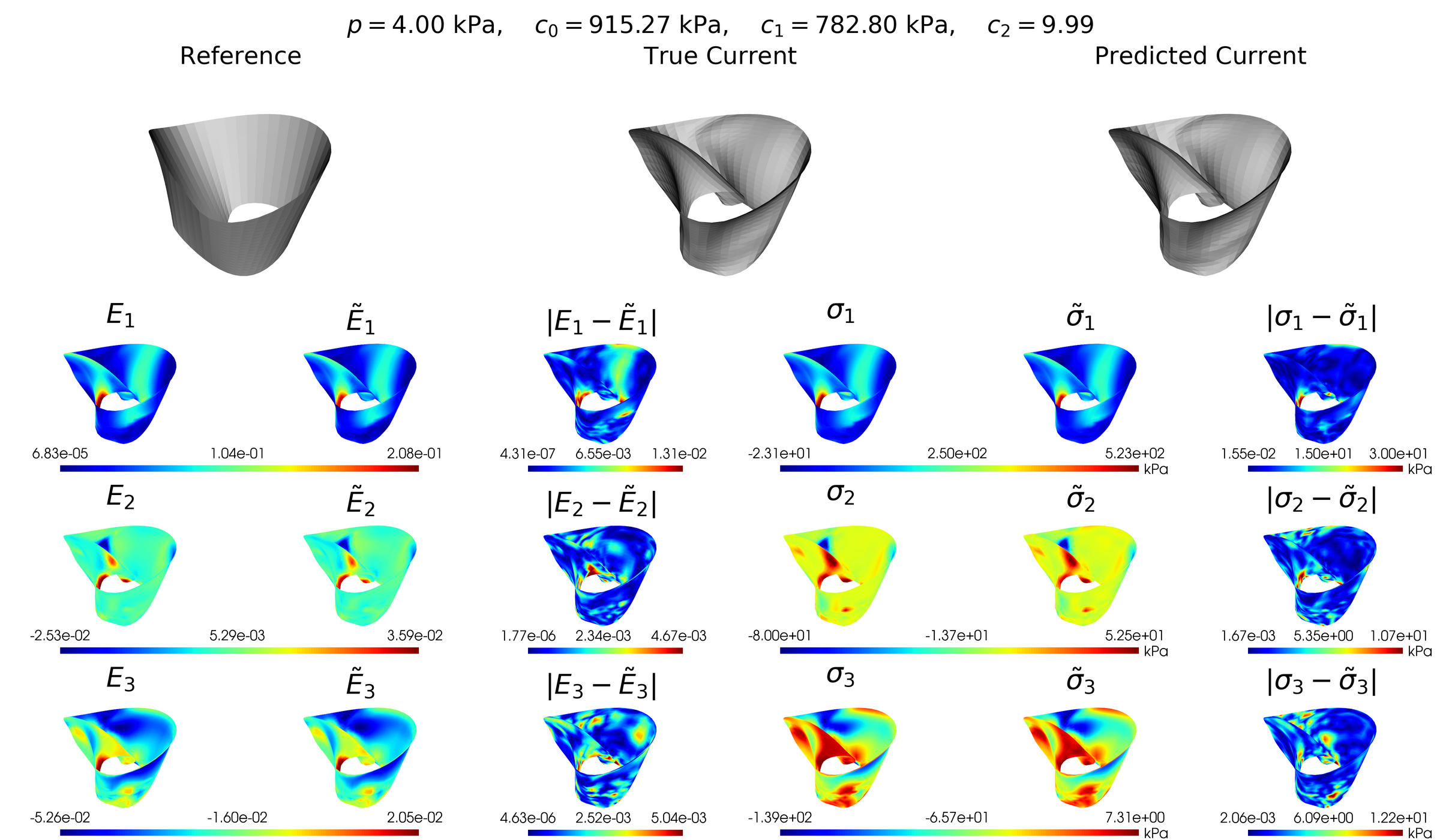}
    \caption{Test example from the trivariate material parameter generalization setting corresponding to a tricuspid valve.}
    \label{fig:mat_param_example}
\end{figure}

Test errors increase progressively across the three splits, reflecting the growing difficulty of each task. In the most demanding setting (Tri), displacement predictions achieve 3.56\% mean relative error, while strain reaches 8.88\% and stress 8.83\%. The close agreement between predicted and ground-truth fields confirms that the model captures the qualitative structure of the mechanical response even under trivariate OOD conditions (Fig.~\ref{fig:mat_param_example}, Supplementary Fig.~3).

In the univariate setting, errors near the training boundary remain comparable to in-distribution levels, followed by a gradual increase deeper into the OOD region, consistent with a distance-to-support effect in which mildly extrapolative values incur only modest additional error (Supplementary Fig.~8). In the bivariate setting, errors are higher and increase more rapidly; extrapolating in multiple constitutive directions simultaneously compounds sensitivity in the material response, confining low-error behavior to a smaller neighborhood near the boundary. In the trivariate setting, we observe a sharp increase in error upon crossing the training boundary, reflecting the immediate shift away from the training support when no constitutive parameters remain in range. Within the trivariate OOD region, however, errors plateau and remain approximately uniform, with only a gradual upward trend as $(c_0,c_1,c_2)$ move farther into their respective OOD intervals.

\subsection{Joint Peak Systolic Blood Pressure and Material Parameter Generalization}

In the most demanding OOD setting, the SBP and all three material parameters are simultaneously withheld from the training distribution, so that no physical conditioning variable remains in range at test time. That is, the model is trained on low SBPs and softer tissue, and tested on higher SBPs and stiffer tissue (see Methods for full details of the split). Errors in this setting are the highest among all OOD experiments: displacement predictions achieve 4.48\% mean relative error, while strain reaches 10.11\% and stress 11.56\%.

\begin{figure}[H]
    \centering
    \includegraphics[width=1\linewidth]{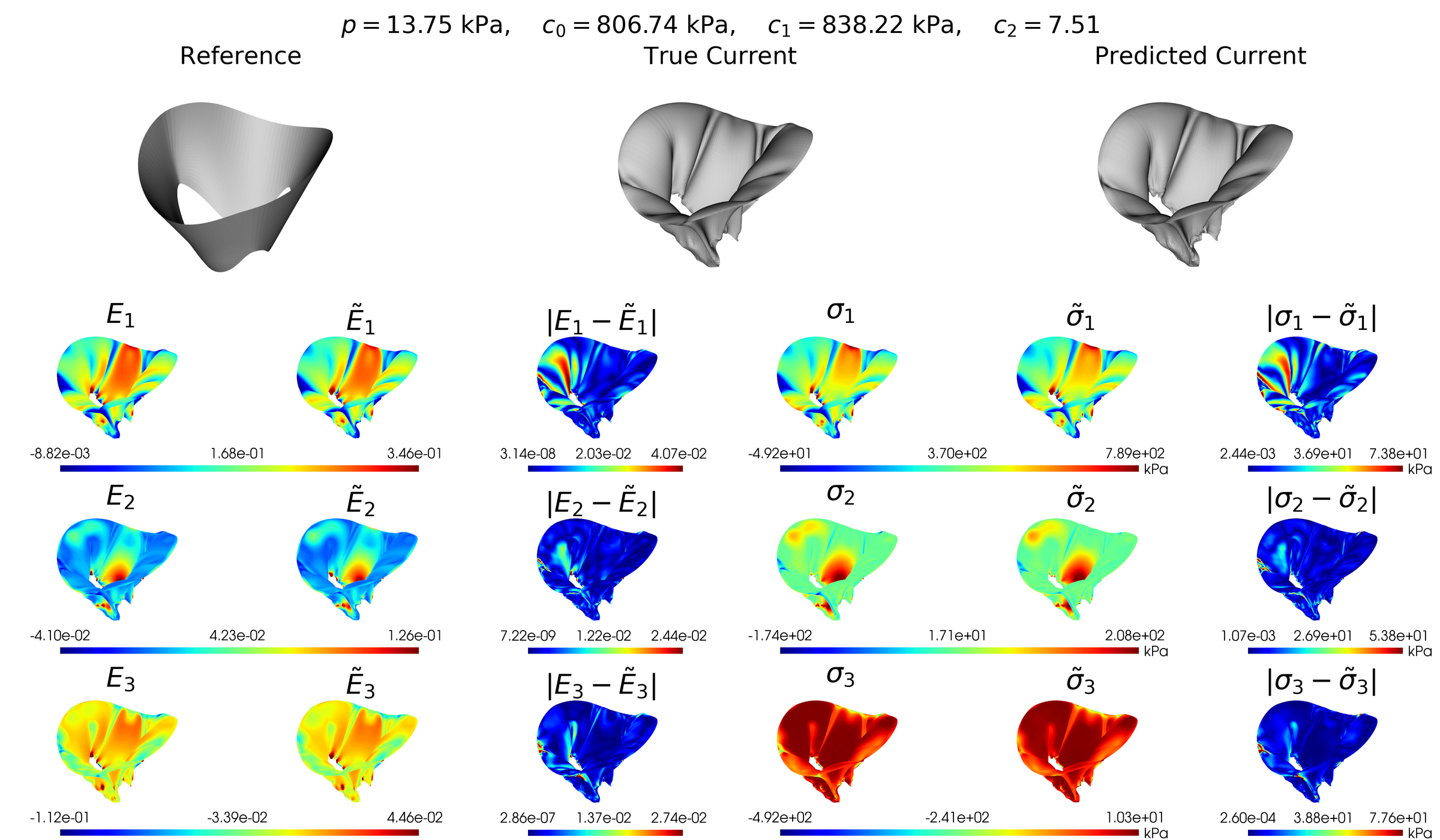}
    \caption{Test example from the joint SBP and material parameter generalization setting corresponding to a tricuspid valve.}
    \label{fig:mat_param_sbp_example}
\end{figure}

Despite the elevated quantitative errors, the predicted fields remain qualitatively consistent with the ground truth. The model correctly identifies the spatial distribution of high-strain and high-stress regions, preserves the overall deformed shape, and captures the dominant mechanical patterns, even though point-wise magnitudes deviate more than in the single-axis OOD settings (Fig.~\ref{fig:mat_param_sbp_example}, Supplementary Fig.~4). This suggests that the learned representation retains meaningful structural information about the mechanical response even when extrapolating simultaneously along all physical conditioning axes.

Within the training region, errors remain approximately uniform across all four physical conditioning variables, consistent with the in-distribution behavior (Supplementary Fig.~9). Beyond the training boundary, errors jump across all four axes; along the SBP axis, they continue to rise moderately with $p$, following the same gradual degradation observed in the SBP-only setting, while along each material parameter axis ($c_0$, $c_1$, $c_2$) they remain roughly flat across the OOD range, as in the trivariate material parameter setting. The joint-generalization error structure thus decomposes into the characteristic behaviors identified along each individual axis, rather than exhibiting a new failure mode unique to the simultaneous OOD shift. The difficulty of this task lies in the nonlinear dependence of the mechanical response on all four physical conditioning variables and on their interactions with geometry, none of which can be anchored by an in-range parameter.

\subsection{Diseased Valve Assessment}

Mitral valve disease encompasses a spectrum of pathologies, each characterized by distinct geometric deformations of the valve apparatus. Tenting height, tenting volume, and annular diameter are standard geometric measures of these deformations, and the mitral valve literature has used them to grade disease severity, guide treatment decisions, and predict outcomes following surgical repair \cite{wagner2014subvalvular,mufarrih2023geometric,delling2014epidemiology}. We test whether the model can recover these measures directly from the predicted current configuration on the three pathologies in the diseased dataset, while simultaneously classifying the underlying disease and predicting the full mechanical fields. Where applicable, measures are computed separately for the anterior and posterior leaflets to capture leaflet-specific deformation patterns.

\begin{figure}[H]
    \centering
    \includegraphics[width=1\linewidth]{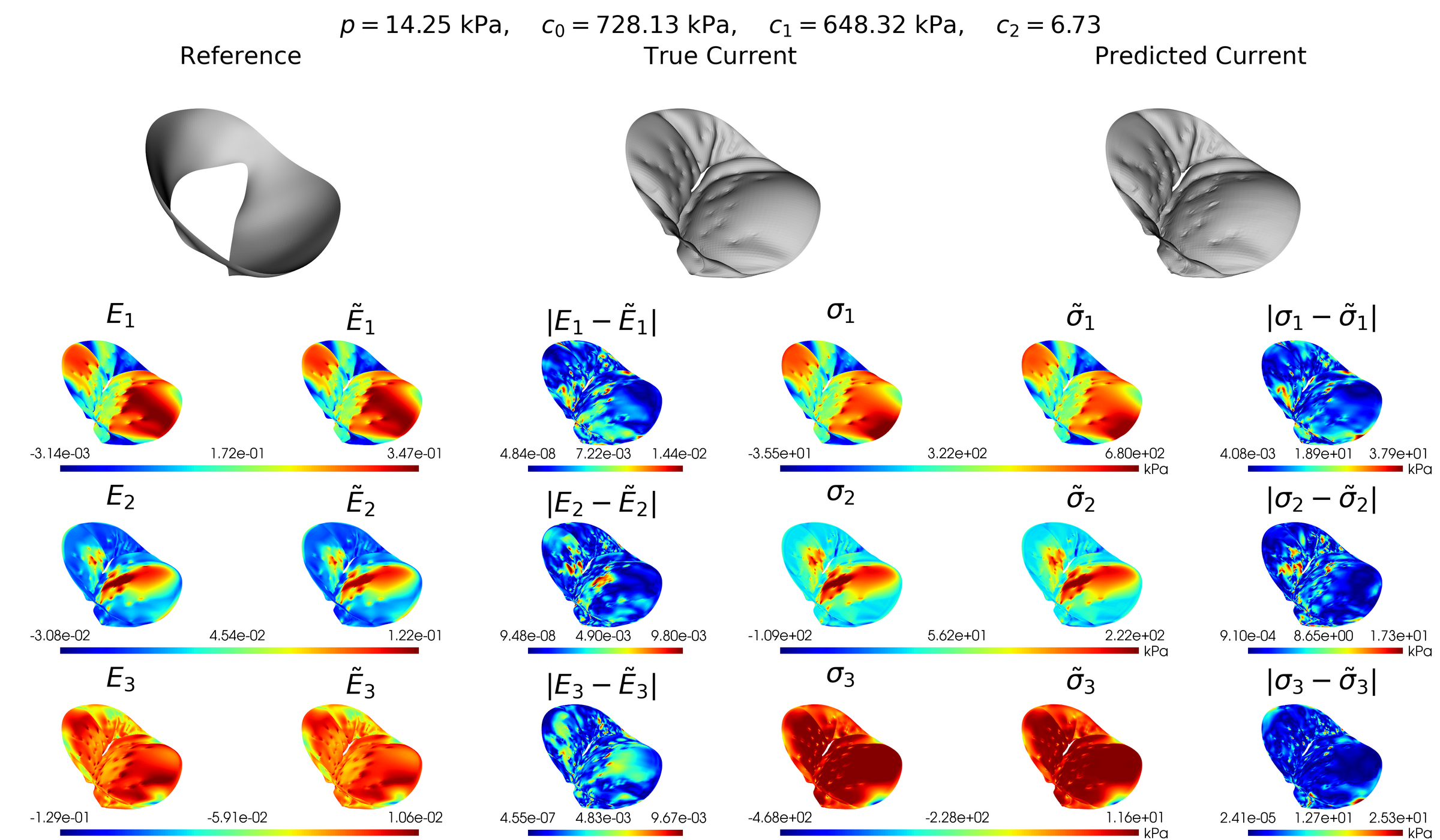}
    \caption{Test example from the diseased joint SBP and material parameter generalization setting corresponding to a tethered mitral valve.}
    \label{fig:diseases_mat_param_sbp_example}
\end{figure}

We evaluate the model under two settings: in-distribution with a random split covering the full parameter ranges, and joint SBP and material parameter generalization. Results for the geometric measures are summarized in Table~\ref{tab:clinical_errors}. Note that the AP and IC diameters are direct indicators of how well the model learns the prescribed displacement boundary condition, highlighting the effectiveness of conditioning on the boundary condition class described in the Methods. In the in-distribution setting, the model achieves low relative errors across all geometric measures for each pathology, with displacement predictions reaching 0.22\% mean relative error, while strain reaches 1.00\% and stress 1.15\%. These results indicate that the predicted displacement fields are sufficiently accurate to recover the measures. In the joint generalization setting, errors increase substantially: displacement predictions achieve 3.57\% mean relative error, while strain reaches 10.92\% and stress 12.06\%. However, the predicted fields degrade in the same manner seen in the base dataset joint generalization setting (Fig.~\ref{fig:diseases_mat_param_sbp_example}, Supplementary Fig.~5).

Within the training region, errors remain approximately uniform across all four physical conditioning variables (Supplementary Fig.~10). Beyond the training boundary, errors jump across all four axes; along the SBP axis, they rise slightly with $p$ over the narrow OOD range spanned in this setting, while along the material parameter axes they remain roughly flat in $c_1$ and $c_2$ and exhibit a mild upward trend in $c_0$ deeper into the OOD region. As in the base dataset joint generalization setting, the error structure decomposes into the characteristic behaviors identified along each individual axis, indicating that the model retains a consistent extrapolation pattern even on the diseased dataset.

The classification head achieves 100\% accuracy on both the training and test sets in both evaluation settings. Although each pathology corresponds to a distinct valve geometry, which makes classification appear geometrically straightforward, as described in the Methods, the encoder must produce a single shared latent representation that supports both classification and the spatially resolved mechanical fields, with no classification-specific encoding pathway. The classification head therefore serves as evidence that the shared representation supports an auxiliary task appended alongside the mechanical fields, rather than as a classification benchmark in its own right.

\begin{table}[H]
\centering
\begin{tabular}{lllccccc}
\toprule
& & & \multicolumn{2}{c}{\textbf{In-dist.}} & \multicolumn{3}{c}{\textbf{Joint Gen.}} \\
\cmidrule(lr){4-5}\cmidrule(lr){6-8}
\textbf{Pathology} & \textbf{Measure} & \textbf{Leaflet} & Train & Test & Train & Test & Test 95\% CI \\
\midrule
\multirow{4.5}{*}{Tethering}
 & \multirow{2}{*}{Tenting Height}   & Anterior  & 0.03 & 0.03 & 0.24 & 0.42 & [0.38, 0.46] \\
 &                                    & Posterior & 0.03 & 0.04 & 0.24 & 1.60 & [1.37, 1.81] \\
\cmidrule(lr){2-8}
 & \multirow{2}{*}{Tenting Volume}   & Anterior  & 0.04 & 0.04 & 0.28 & 2.69 & [2.55, 2.82] \\
 &                                    & Posterior & 0.06 & 0.06 & 0.47 & 2.26 & [2.15, 2.37] \\
\midrule
\multirow{6.75}{*}{P2 Prolapse} 
 & \multirow{2}{*}{Prolapse Height}  & Anterior  & 0.12 & 0.13 & 0.34 & 2.08 & [1.64, 2.46] \\
 &                                    & Posterior & 0.11 & 0.11 & 0.25 & 0.80 & [0.61, 0.96] \\
\cmidrule(lr){2-8}
 & \multirow{2}{*}{Prolapse Volume}  & Anterior  & 0.17 & 0.18 & 0.45 & 3.47 & [2.30, 4.53] \\
 &                                    & Posterior & 0.14 & 0.16 & 0.31 & 3.15 & [2.07, 4.12] \\
\cmidrule(lr){2-8}
 & \multirow{2}{*}{PV--PH Ratio}    & Anterior  & 0.12 & 0.13 & 0.36 & 1.76 & [0.93, 2.52] \\
 &                                    & Posterior & 0.13 & 0.15 & 0.23 & 2.56 & [1.60, 3.46] \\
\midrule
\multirow{3.75}{*}{Dilation} 
 & Anteroposterior Diameter     & -- & 0.01 & 0.01 & 0.09 & 0.32 & [0.29, 0.35] \\
\cmidrule(lr){2-8}
 & Intercommissural Diameter    & -- & 0.01 & 0.01 & 0.06 & 0.99 & [0.94, 1.04] \\
\cmidrule(lr){2-8}
 & AP--IC Ratio                 & -- & 0.02 & 0.02 & 0.14 & 1.30 & [1.23, 1.37] \\
\bottomrule
\end{tabular}
\caption{Average relative errors (\%) for geometric measures of valve function on the diseased dataset. For the joint generalization test split (103 tethering, 98 P2 prolapse, and 103 dilation samples), 95\% confidence intervals on the mean are obtained by nonparametric bootstrap over test samples (10{,}000 resamples). Intervals reflect variability over SBP and material parameters for a single geometry per pathology.}
\label{tab:clinical_errors}
\end{table}

\subsection{Geometric and Constitutive Consistency}

The FE solutions carry three properties that PCNO is never given. The first is kinematic, in that the current configuration in the converged FE data is free of self-intersection. The second is volumetric, in that the term $U(J)$ of Eq.~\ref{eq:sed_vol} governs the local volume response at a finite bulk modulus $\kappa$, so volume change is present in the FE solutions rather than suppressed. The third is constitutive, in that the isotropic form of Eq.~\ref{eq:sed_dev} fixes a point-wise relationship between the stress and strain fields. PCNO reproduces none of the three by construction, so any such structure in its output must be inherited from the training data. We therefore ask how far that inheritance extends and whether it survives in OOD settings. The checks below cover the in-distribution and the joint SBP and material parameter generalization settings on both datasets, with the exception of the self-intersection test, which is restricted to the diseased dataset because the pairwise intersection computation is expensive.

We first ask whether a predicted current configuration self-intersects at all, and if so, how severe the resulting penetration is. Each quadrilateral is split into two triangles, and an intersection test over non-adjacent triangle pairs flags two different situations. Small errors in the predicted displacements can tilt a triangle enough that it clips another one a short distance away across the surface, which reflects the flat approximation of a curved leaflet rather than a loss of embedding, whereas genuine self-penetration occurs when two parts of the leaflet that are far apart across the surface come to occupy the same space. Penetration depth does not separate these two situations, so we count a pair only when its two triangles are far apart along the leaflet surface. We report the fraction of predicted configurations containing at least one such pair, the worst-case penetration depth over the mesh, and that depth normalized by the mean edge length of the same mesh, which measures penetration relative to the representative element size rather than in absolute units. The same test returns no pairs on the ground-truth finite-element configurations. Table~\ref{tab:intersection} reports the mean of each quantity per pathology.

Self-intersection is rare in tethering and dilation, affecting at most $1.58\%$ of configurations in any split. It is far more common in P2 prolapse, where the incidence rises from 2.83\% and 3.05\% for the in-distribution setting to 22.40\% and 57.14\% under joint generalization, reflecting the geometry of the pathology. The prolapsing segment brings the two leaflets into near contact over an extended region, so the clearance a predicted configuration has to preserve is much smaller than in the other two pathologies. What the higher incidence does not bring with it is a higher severity. The worst-case penetration depth in that setting is $0.22$\,mm on both splits, no larger than the $0.24$\,mm and $0.27$\,mm measured in distribution, and across the whole table it never exceeds $0.41$\,mm and never increases appreciably from a training split to the corresponding test split. The average number of distinct self-intersecting regions per configuration likewise stays between 1.00 and 1.75, so intersections, where they occur, are limited to a few localized overlaps. Normalized against the discretization, penetration remains a fraction of one element edge throughout, between $0.30$ and $0.66$; this ratio is mesh dependent by construction, and the three diseased geometries are meshed at comparable resolution, so the values are comparable across the pathologies reported here but not transferable to a different discretization.

We next examine the local volume ratio, computed from the right Cauchy--Green deformation tensor $\boldsymbol{C} = \boldsymbol{I} + 2\boldsymbol{E}$ as
\begin{equation}
    J = \sqrt{\det(\boldsymbol{C})},
\end{equation}
where $J \in \mathbb{R}^{N\times1}$ is the dimensionless ratio of local volume in the current configuration to that in the reference configuration. Since $\kappa$ is finite, the quantity to be recovered is the departure of $J$ from unity rather than the incompressible limit itself, so we normalize by the volume change present in the reference solution,
\begin{equation}
  \text{Relative } L^2(J)
    = 100 \times \frac{\|J - \tilde{J}\|_2}{\|J-1\|_2}.
  \label{eq:rel_l2_J}
\end{equation}
Table~\ref{tab:coaxiality} reports this quantity for both datasets. In the in-distribution setting, the error is small and essentially identical between the training and test splits, $4.02\%$ and $4.06\%$ on the base dataset and $0.92\%$ and $0.95\%$ on the diseased dataset. Under joint generalization, the training splits stay in the same range, $4.69\%$ and $3.30\%$, while the test splits rise to $11.14\%$ and $11.50\%$, comparable in magnitude to the strain and stress errors of that setting.

\begin{figure}[H]
    \centering
    \includegraphics[width=1\linewidth]{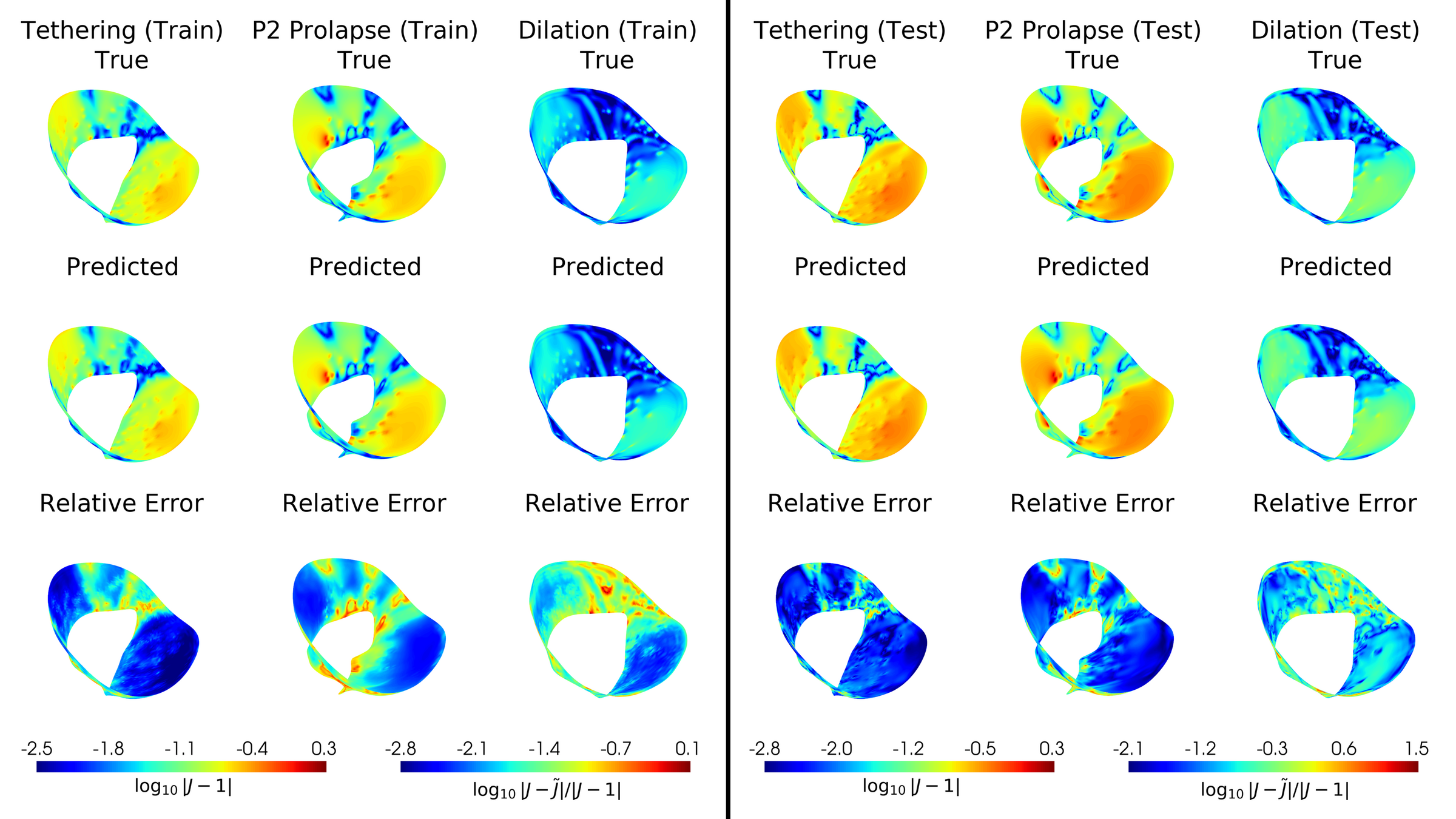}
    \caption{Local volume change under joint SBP and material parameter generalization on the diseased dataset. Columns are pathologies and rows give the per-node mean of $|J-1|$ for the reference solution, of $|\tilde{J}-1|$ for the prediction, and of the point-wise relative error $|J-\tilde{J}|/|J-1|$, plotted on the reference configuration on a $\log_{10}$ scale. Training split on the left, test split on the right, each with its own color scale.}
    \label{fig:J_dist}
\end{figure}

Figure~\ref{fig:J_dist} visualizes the same quantity over the mesh for the joint generalization setting on the diseased dataset, whose three geometries can each be shown separately. For every node, we take the mean over the samples of a split of $|J-1|$ for the reference solution, of $|\tilde{J}-1|$ for the prediction, and of the point-wise relative error $|J-\tilde{J}|/|J-1|$, plotted on the reference configuration on a logarithmic scale. Between the three pathologies, the volume changes are not identical. Tethering and P2 prolapse cases exhibit stronger volume changes than the dilation cases. The test configurations, which lie outside the training range in SBP and in all three material parameters, carry visibly larger volume changes than the training ones in each pathology. PCNO is therefore extrapolating beyond the volume changes it was shown, and the predicted maps nonetheless remain nearly indistinguishable from the reference maps.

Finally, we test the point-wise relationship that the constitutive law imposes between the stress and strain fields. For an isotropic hyperelastic material, the Cauchy stress is coaxial with the left Cauchy--Green deformation tensor $\boldsymbol{B}$, a consequence of the representation theorem for isotropic tensor functions \cite{rivlin1997stress}. For the uncoupled SED of Eq.~\ref{eq:sed_dev}, which depends on $I_1^*$ alone, this specializes to the stronger statement that the deviatoric stress is a strictly positive multiple of $\mathrm{dev}(\boldsymbol{B})$,

\begin{equation}\label{eq:dev_prop}
    \mathrm{dev}(\boldsymbol{\sigma}) = \alpha \, \mathrm{dev}(\boldsymbol{B}), \qquad \alpha = 2 J^{-5/3}\, \frac{\partial \Psi^*}{\partial I_1^*} > 0 ,
\end{equation}
positivity following from $I_1^* \geq 3$ and $c_0, c_1, c_2 > 0$ (see Supplementary Section~1 for the derivation). Here $\mathrm{dev}(\cdot)$ subtracts one third of the trace of a tensor from its diagonal, so the principal values of a deviator are those of the tensor with their mean removed. The volumetric pressure $dU/dJ$ contributes equally to all three principal stresses, and so it enters only their mean. Writing $\lambda_1^2 \geq \lambda_2^2 \geq \lambda_3^2$ for the eigenvalues of $\boldsymbol{C} = \boldsymbol{I} + 2\boldsymbol{E}$, which coincide with those of $\boldsymbol{B}$, and $\sigma_1 \geq \sigma_2 \geq \sigma_3$ for the eigenvalues of $\boldsymbol{\sigma}$, we collect the deviatoric principal values of each field into a vector in $\mathbb{R}^3$,

\begin{equation}\label{eq:coax_triples}
    a_i = \lambda_i^2 - \frac{1}{3}\sum_{j=1}^3 \lambda_j^2, \qquad b_i = \sigma_i - \frac{1}{3}\sum_{j=1}^3 \sigma_j ,
\end{equation}
so that $\boldsymbol{a}$ holds the principal values of $\mathrm{dev}(\boldsymbol{B})$ and $\boldsymbol{b}$ those of $\mathrm{dev}(\boldsymbol{\sigma})$. Eq.~\ref{eq:dev_prop} then requires $\boldsymbol{b} = \alpha \boldsymbol{a}$, and since $\alpha > 0$ preserves the ordering of the eigenvalues, the descending sort is consistent between the two. The two vectors must therefore be parallel, which we measure through the direction cosine
\begin{equation}\label{eq:coax_cosine}
    \rho = \cos\theta = \frac{\sum_{i=1}^3 a_i b_i}
                             {\|\boldsymbol{a}\|\,\|\boldsymbol{b}\|},
\end{equation}
with $\theta = \arccos\rho$ reported in degrees.

Eq.~\ref{eq:coax_cosine} tests the principal values of Eq.~\ref{eq:dev_prop} rather than the alignment of the principal directions. The latter is not testable in our setting, since $\boldsymbol{\sigma}$ is defined in the current configuration and $\boldsymbol{E}$ in the reference one, and comparing their principal directions would require the rotation relating the two frames, which PCNO does not predict. What $\rho = 1$ remains necessary for is the constitutive response of Eq.~\ref{eq:dev_prop}, which fails generically for any isotropic SED carrying an $I_2^*$ dependence.

Because the principal values of both fields are sorted, even unrelated stress and strain states give a large $\rho$, so $\rho$ must be read against an empirical baseline rather than against zero. We fix that baseline by pairing each strain vector with a stress vector drawn from a randomly chosen node of the same case, which preserves the marginal distribution of both fields and destroys only their point-wise correspondence. This gives $\theta_{\text{null}} \approx 14^\circ$--$16^\circ$ across all settings and supplies the zero of

\begin{equation}\label{eq:hat_rho}
    \hat{\rho} = \frac{\rho - \rho_\text{null}}{1 - \rho_\text{null}},
\end{equation}
on which $\hat\rho = 0$ means no better than mismatched stress and strain and $\hat\rho = 1$ is exact proportionality. The FE data does not reach $\rho = 1$, since stress and strain are recovered at the nodes by independent averaging and extrapolation from the element integration points, which breaks the point-wise constitutive link holding at those points. $\theta_{\text{FE}}$ is therefore a floor that PCNO inherits together with the training data, and the per-node difference $|\Delta\theta| = |\theta_{\text{PCNO}} - \theta_{\text{FE}}|$ isolates the model from it.

Table~\ref{tab:coaxiality} brackets every PCNO value between these two references. On the in-distribution settings, PCNO sits on the FE floor, with a per-node discrepancy of $0.46^\circ$ and $0.47^\circ$ on the base dataset and $0.22^\circ$ on both splits of the diseased dataset, and $\hat\rho$ within $0.003$ of the FE value throughout. Under joint generalization, the separation grows but stays small, the base dataset test split giving $2.94^\circ$ for PCNO against $2.19^\circ$ for FE and a per-node discrepancy of $1.76^\circ$. On the diseased dataset, where field errors are largest ($10.92\%$ for $\boldsymbol{E}$ and $12.06\%$ for $\boldsymbol{\sigma}$, Table~\ref{tab:results_summary}), the mean is $3.29^\circ$ for PCNO against $2.16^\circ$ for FE, or $\hat\rho = 0.8867$ against $0.9305$, with the per-node discrepancy averaging $1.98^\circ$.

Nothing in PCNO enforces the coupling between stress and strain. The two fields are decoded as separate channel blocks of a single conditioned output head and supervised by independent per-field terms, with no cross-field constraint and no physics residual, though both are decoded from a shared latent representation that the architecture is designed to exploit. The consistency reported here is therefore emergent rather than explicitly enforced. Its interest lies in the OOD setting, where agreement with the FE reference degrades to roughly $11\%$ and $12\%$ relative $L^2$ error in strain and stress while the predicted fields remain mutually consistent to within $1.98^\circ$ per node of that same reference.

\begin{table}[H]
\centering
\small
\begin{tabular}{lllcccc}
\toprule
\multirow{2}{*}{\textbf{Dataset}}
& \multirow{2}{*}{\textbf{Setting}}
& \multirow{2}{*}{\textbf{Split}}
& \multirow{2}{*}{\makecell[c]{$\theta$ ($^\circ$) \\ (FE / PCNO)}}
& \multirow{2}{*}{$|\Delta\theta|$ ($^\circ$)}
& \multirow{2}{*}{\makecell[c]{$\hat{\rho}$ \\ (FE / PCNO)}}
& \multirow{2}{*}{Rel. $L^2(J)$ (\%)} \\
& & & & & & \\
\midrule
\multirow{4}{*}{Base}
 & \multirow{2}{*}{In-distribution}
   & Train & 1.95 / 2.05 & 0.46 & 0.9578 / 0.9552 & 4.02 \\
 & & Test  & 1.95 / 2.05 & 0.47 & 0.9517 / 0.9499 & 4.06 \\
\cmidrule(lr){2-7}
 & \multirow{2}{*}{\makecell[l]{Joint SBP and Material \\ Parameter Generalization}}
   & Train & 1.91 / 1.99 & 0.44 & 0.9540 / 0.9527 & 4.69 \\
 & & Test  & 2.19 / 2.94 & 1.76 & 0.9344 / 0.9117 & 11.14 \\
\midrule
\multirow{4}{*}{Diseased}
 & \multirow{2}{*}{In-distribution}
   & Train & 2.10 / 2.13 & 0.22 & 0.9301 / 0.9294 & 0.92 \\
 & & Test  & 2.10 / 2.13 & 0.22 & 0.9303 / 0.9296 & 0.95 \\
\cmidrule(lr){2-7}
 & \multirow{2}{*}{\makecell[l]{Joint SBP and Material \\ Parameter Generalization}}
   & Train & 2.13 / 2.24 & 0.48 & 0.9259 / 0.9229 & 3.30 \\
 & & Test  & 2.16 / 3.29 & 1.98 & 0.9305 / 0.8867 & 11.50 \\
\bottomrule
\end{tabular}
\caption{Stress–strain deviatoric proportionality and volume-ratio consistency on both datasets, averaged over each split. $\theta$ is the angle between the deviatoric principal values of $\boldsymbol{\sigma}$ and $\boldsymbol{B}$ (Eq.~\ref{eq:coax_cosine}), given for the finite-element reference and for PCNO. $\Delta\theta$ is the mean per-node difference between them. $\hat{\rho}$ rescales $\cos\theta$ against a mismatched stress–strain baseline (Eq.~\ref{eq:hat_rho}), so $\hat{\rho} = 1$ is exact proportionality.}
\label{tab:coaxiality}
\end{table}

\begin{table}[H]
\centering
\begin{tabular}{@{}l l c c c c@{}}
\toprule
 & & \multicolumn{2}{c}{\textbf{In-dist.}} & \multicolumn{2}{c}{\textbf{Joint gen.}} \\
\cmidrule(lr){3-4} \cmidrule(lr){5-6}
\textbf{Pathology} & \textbf{Metric} & Train & Test & Train & Test \\
\midrule
\multirow[c]{4}{*}{Tethering}
  & Self-intersecting configurations (\%) & 0.04 & 0.07 & 1.58 & 0.00 \\
  & Self-intersecting regions             & 1.75 & 1.00 & 1.04 & -- \\
  & Penetration depth (mm)                & 0.41 & 0.27 & 0.19 & -- \\
  & Penetration depth / mean edge length  & 0.66 & 0.44 & 0.31 & -- \\
\midrule

\multirow[c]{4}{*}{P2 Prolapse}
  & Self-intersecting configurations (\%) & 2.83 & 3.05 & 22.40 & 57.14 \\
  & Self-intersecting regions             & 1.21 & 1.17 & 1.27 & 1.11 \\
  & Penetration depth (mm)                & 0.24 & 0.27 & 0.22 & 0.22 \\
  & Penetration depth / mean edge length  & 0.33 & 0.37 & 0.30 & 0.30 \\
\midrule

\multirow[c]{4}{*}{Dilation}
  & Self-intersecting configurations (\%) & 0.00 & 0.00 & 0.20 & 0.00 \\
  & Self-intersecting regions             & --   & --   & 1.29 & -- \\
  & Penetration depth (mm)                & --   & --   & 0.22 & -- \\
  & Penetration depth / mean edge length  & --   & --   & 0.33 & -- \\
\bottomrule
\end{tabular}
\caption{Self-intersection of the predicted current configurations on the diseased dataset, averaged over each split. Dashes mark splits with no self-intersecting configurations.}
\label{tab:intersection}
\end{table}

\newpage
\section{Discussion}

In this study, we developed PCNO, a transformer-based surrogate model for predicting the mechanical response (displacements, stresses, and strains) of atrioventricular valves under varying SBP and Lee--Sacks material parameters, with isotropic hyperelastic constitutive behavior. We evaluated PCNO on two datasets, a base dataset comprising both functional and regurgitant valves and a diseased dataset spanning three mitral valve pathologies (tethering, P2 prolapse, and dilation). Ground truth throughout is the FE solver used to generate the training data, so every reported error measures agreement with respect to that solver rather than with in vivo measurement. On both datasets, PCNO performed strongly under in-distribution testing and across several OOD regimes, maintaining accurate predictions despite substantial distributional shifts. Even in settings where the relative $L^2$ error was elevated, the predicted fields remained qualitatively faithful to the ground truth.

On the diseased dataset, PCNO accurately recovered the geometric measures of valve function (tenting height and volume, prolapse height and volume, and annular diameters). These quantities are never supervised during training and instead emerge from the predicted current configuration, and even under OOD conditions, the mean relative error on every measure remained below 3.5\%. Beyond field prediction and measure recovery, PCNO performs pathology classification from the same encoder, using only a lightweight task-specific head appended to the shared latent representation. The classification task is itself geometrically straightforward, since each pathology corresponds to a distinct geometry, so this result is best read as evidence that a single shared representation supports an additional task rather than as a difficult classification benchmark.

\subsection{Computational Cost}
For the base dataset, we compared PCNO inference time with the FE solver used to generate the ground-truth simulations. The base dataset contains three mesh sizes (1{,}029, 11{,}917, and 12{,}436 nodes), which we group into a coarse regime (1{,}029 nodes; 256{,}847 examples) and a fine regime (11{,}917--12{,}436 nodes; 74{,}930 examples). FE solves were run on a CPU cluster with 4 cores per simulation and 20--50 simulations executing in parallel; reported FE times are individual simulation run times and do not depend on the degree of parallelism. PCNO inference was timed on a single NVIDIA RTX A6000 GPU at batch size one after warm-up, averaged over 5,000 runs, and covers the forward pass over all mesh nodes. On the coarse mesh, FE required a mean of 13.59~s per simulation (median 9.00 s) against 6.01~ms for PCNO; on the fine mesh, 847.98~s (median 632.00~s) against 53.38--55.57~ms. Because the two pipelines run on different hardware, these figures describe the per-query cost of each as deployed rather than an algorithmic comparison. In a separate set of measurements on the same GPU, averaged over 100 runs per batch size, PCNO evaluates 183--238 coarse and 19--20 fine queries per second across batch sizes 1--64; coarse throughput reaches its maximum at batch size 32, while fine throughput is already saturated at batch size 1. Because the two sets of measurements were made separately, the throughput at batch size one differs from the rate implied by the per-query times above.

From the coarse to the fine meshes, node count increases roughly 12-fold, while mean FE run time rises 62-fold and PCNO inference time 9-fold. The two mesh groups also differ in geometry and boundary conditions, so we do not attribute the FE increase to mesh size alone or fit a scaling law to it. PCNO inference cost is driven by the number of nodes and grows approximately linearly with it, as expected from the Rank-Augmented Linear Attention mechanism \cite{fan2025breaking}, whose cost is linear in the number of nodes (see Methods; Supplementary Table~4, Supplementary Fig.~11).

These per-query figures exclude the one-time cost of obtaining the surrogate, summarized in Table~\ref{tab:datagen_cost}. Generating the full base and diseased datasets, including test examples, required 18{,}619 and 2{,}691 hours of FE run time, and training PCNO required 12.29 hours on a single GPU, a cost set by the fixed number of optimization steps and sampled nodes and independent of dataset size. Each model is trained only on its own training split; for the joint SBP and material parameter generalization setting, which gives the most demanding OOD results reported here, those splits required 5{,}028 hours on the base dataset and 817 hours on the diseased dataset, 27\% and 30\% of the respective totals. Because the training data are themselves FE solves, and the cost of a query is negligible by comparison, the one-time cost of a model is fully amortized after roughly as many queries as it has training examples.

\begin{table}[H]
\centering
\begin{tabular}{llccc}
\toprule
\textbf{Dataset} & \textbf{Split} & \textbf{Examples} & \textbf{FE Run Time} & \textbf{Wall-Clock} \\
\midrule
\multirow{2}{*}{Base}
 & Full (train + test) & 331{,}777 & 18{,}619 h & 15.5--38.8 days \\
 & Joint Gen. (train)  & 95{,}102  & 5{,}028 h  & 4.2--10.5 days \\
\midrule
\multirow{2}{*}{Diseased}
 & Full (train + test) & 41{,}052  & 2{,}691 h  & 2.2--5.6 days \\
 & Joint Gen. (train)  & 10{,}480  & 817 h    & 0.7--1.7 days \\
\bottomrule
\end{tabular}
\caption{Finite element data-generation cost on the base and diseased datasets. FE run time is the summed run time of individual simulations, and Joint Gen. (train) is the training split of the joint SBP and material parameter generalization setting. Wall-clock time divides FE run time by the 20--50 simulations run concurrently and is approximate.}
\label{tab:datagen_cost}
\end{table}

\subsection{Conditioning as Physics Integration} 

Physics integration in neural surrogates spans a spectrum, with soft penalties on PDE residuals at one end, hard architectural priors enforcing exact symmetries or conservation laws in the middle, and data-driven supervision augmented by parameter conditioning at the other. PCNO occupies the last of these by design, and we view this choice as the appropriate one for surrogate modeling at the scale required by multi-query parameter exploration. Residual-based formulations require automatic differentiation (auto-diff) of the network at every mesh node, which is inexpensive for the point-wise multilayer perceptrons commonly used in physics-informed neural networks \cite{raissi2019physics}, where each output depends only on its own input coordinate and all $N$ per-node derivatives follow from a single backward pass. Attention-based operator architectures do not admit this reduction, since token (mesh node) interactions couple every output to every input, leaving a dense per-node Jacobian whose diagonal entries must be recovered at a cost scaling as $\mathcal{O}(N^2)$ per derivative order. This cost counts backward passes rather than forward evaluations, so it is unchanged by linear-complexity attention, and it compounds further for the second-order spatial derivatives that arise in the momentum balance of finite-strain theory. Residual-based formulations also require the relative weighting of data, residual, and boundary-condition terms to be tuned per dataset, and they exhibit gradient pathologies, spectral bias, and causality violation that have been characterized extensively in the literature \cite{wang2022and,wang2021understanding,wang2024respecting,wang2026pinns}. PCNO instead attaches task-specific heads to a shared latent representation, with all objectives operating in compatible supervised regimes. Hard architectural priors such as exact equivariance or divergence-free constructions can in principle reduce the training data required to reach a given accuracy, but they restrict the family of admissible architectures and complicate extensions to heterogeneous geometries, multiple boundary-condition types, and multi-task pipelines, precisely the regime in which PCNO is designed to operate.

A further consequence of this design is that the architecture and training procedure are not tied to any single constitutive choice. Because the Lee--Sacks parameters enter PCNO only as conditioning inputs, the encoder learns a representation of valve mechanics that is shaped by, but not structurally tied to, the SED used to generate the training data. Extending PCNO to a different constitutive family would proceed by generating simulation data under the new model, supplying the corresponding material parameters as conditioning inputs, and fine-tuning the existing encoder, with the components handling the geometric and kinematic structure of the problem carrying over. Transfer of this form has been demonstrated for PDE foundation models, where a backbone pretrained on one family of governing equations is adapted to equations unseen during pretraining by reinitializing only the input and output embeddings and fine-tuning on a small number of task-specific samples \cite{herde2024poseidon}, and related work trains a single backbone across heterogeneous physical systems by projecting their fields into a shared embedding space \cite{mccabe2023multiple}. Constitutive models differ in how many parameters they carry, so the conditioning input is the one component a new family places a demand on. A fixed-width vector, as used here, accommodates this only through masking or through an integer index identifying the family, whose arbitrary spacing the network would read as meaningful. Forming each material parameter into a token from a learned parameter-specific embedding and its value avoids this. In this context, the conditioning input becomes a variable-length set to which attention applies unmodified, and a new family contributes additional embeddings while leaving those already learned intact. Such variable-specific encodings have been used to similar effect at the level of physical field variables, letting a single operator accept varying numbers of inputs and adapt to systems with additional interacting variables \cite{rahman2024pretraining}. A residual-based formulation admits no comparable path, since the residual is itself constitutive-model-specific, so any change to the SED requires redesigning the loss and re-training the model. We do not pursue cross-constitutive transfer in this work, and the evidence above concerns related operator-learning settings rather than valve mechanics specifically, but the design makes it a concrete direction for future work.

\subsection{Relation to Other Surrogate Architectures}

The comparisons above establish how closely PCNO reproduces the FE solver but not whether the attention-based design of PCNO is the appropriate one. We therefore compare PCNO with two representative graph-based surrogates, a standard message-passing graph neural network (GNN) \cite{gilmer2017neural} and a graph neural operator (GNO) \cite{li2020neural}, on the base dataset under the in-distribution and joint SBP and material parameter generalization settings. All models are trained on the same splits and receive the same nodal features and conditioning variables, with the graph baselines additionally operating on a graph constructed over the mesh nodes. We report PCNO (Large), comparable to the baselines in training time, and PCNO (Small), comparable in parameter count. Architectures and training configurations of all models are found in Supplementary Section~2.

PCNO (Large) achieves lower error than both baselines on every field in both settings (Table~\ref{tab:surrogate_baselines}), reducing the errors by 17.08\% to 39.61\% in-distribution. PCNO (Small), which trains in less than a fifth the time of either baseline, improves on the GNO by 7.41\% to 9.74\% in-distribution but has 6.11\% to 8.71\% higher error than the GNN, so at a comparable parameter count the in-distribution advantage holds only against the GNO. Under joint generalization, both PCNO configurations achieve lower error than both baselines on every field, with reductions of 2.77\% to 32.09\%. The reductions grow from displacement (2.77\% to 5.68\%) to strain (9.38\% to 11.86\%) to stress (14.12\% to 32.09\%), so the advantage outside the training distribution is concentrated in the derived fields. The two PCNO configurations perform nearly identically in this setting, with PCNO (Large) slightly more accurate in displacement and strain and PCNO (Small) marginally more accurate in stress, indicating that the additional capacity of PCNO (Large) yields little benefit outside the training region; this is consistent with the stronger weight decay adopted for this split to limit overfitting of the larger model (see Methods). Because PCNO (Small) outperforms both baselines under joint generalization at a comparable parameter count, the advantage in this setting is attributable to the architecture rather than to model size, whereas in-distribution the margin of PCNO (Large) over the GNN depends in part on its larger capacity.

We view the choice of attention over message passing as a deliberate trade-off. Message passing builds in the prior that a node is most strongly influenced by its neighbors, a locality assumption credited with the ability of graph networks to generalize from limited data \cite{battaglia2018relational}. Attention gives up this prior, the weak inductive bias noted in the Limitations, but in exchange aggregates over every node in every layer. The solution operator of the equilibrium problem is nonlocal even though the governing equations are local \cite{kovachki2023neural}, whereas after $L$ layers a message-passing model sees only nodes within $L$ hops. On a mesh-based graph this neighborhood shrinks physically as the mesh is refined, since each hop spans a shorter edge, while on a radius-based graph the reach is fixed but the per-layer cost grows. Two couplings central to valve closure are nonlocal in this sense. At coaptation, nodes on opposing free edges come into contact although they are separated in the reference configuration by the open orifice and, along the leaflet surface, by a path through the commissures. Because chordal connectivity is withheld from the inputs, whether a node is tethered also depends on its position relative to papillary muscle tips absent from the input, which must be inferred from global context that attention supplies at every layer and message passing gathers only with depth. The margins in Table~\ref{tab:surrogate_baselines} are consistent with these arguments but do not isolate them, and because the generalization setting varies pressure and material parameters rather than geometry, it does not test the regime in which the locality prior is expected to help most.

The same trade shapes how geometry enters each model. A message-passing model learns local geometry from the relative positions of neighboring nodes, whereas PCNO embeds each token point-wise and therefore receives local shape explicitly through the structure-tensor features, with global shape supplied by the integral curvature variables (see Methods). Because the graph baselines receive the same nodal features, the comparison in Table~\ref{tab:surrogate_baselines} contrasts the aggregation mechanism with the geometric encoding held fixed, and the explicit local descriptors do not allow the graph models to match PCNO (Large), or PCNO (Small) under joint generalization.

Finally, all models are evaluated on geometries seen during training. The comparison therefore does not test geometry generalization and should not be read as evidence that PCNO would retain its advantage on unseen geometries. It does establish that, for exploring pressure and tissue properties over a fixed set of geometries, the setting PCNO is designed for, the attention-based design is more accurate than representative graph-based surrogates at a comparable training time and at a comparable parameter count when generalizing jointly over pressure and material parameters.

\begin{table}[H]
\centering
\begin{tabular}{llccccc}
\toprule
\textbf{Model} & \textbf{Setting} & Rel. $L^2(\boldsymbol{u})$ & Rel. $L^2(\boldsymbol{E})$ & Rel. $L^2(\boldsymbol{\sigma})$ & \textbf{Params} & \textbf{Time} \\
\midrule
\multirow{2}{*}{\makecell[l]{PCNO\\(Large)}}
 & In-distribution & 0.93 & 3.48 & 3.69 & \multirow{2}{*}{93 M} & \multirow{2}{*}{12.29 h} \\
 & Joint Gen.      & 4.48 & 10.11 & 11.56 & & \\
\midrule
\multirow{2}{*}{\makecell[l]{PCNO\\(Small)}}
 & In-distribution & 1.39 & 4.62 & 4.81 & \multirow{2}{*}{12 M} & \multirow{2}{*}{2.06 h} \\
 & Joint Gen.      & 4.57 & 10.34 & 11.49 & & \\
\midrule
\multirow{4}{*}{GNN}
 & In-distribution & 1.31 & 4.25 & 4.45 & \multirow{4}{*}{14.5 M} & \multirow{4}{*}{12.63 h} \\
 & & \footnotesize{($-29.01$, $6.11$)}
   & \footnotesize{($-18.12$, $8.71$)}
   & \footnotesize{($-17.08$, $8.09$)} & & \\
 & Joint Gen.      & 4.70 & 11.47 & 16.92 & & \\
 & & \footnotesize{($-4.68$, $-2.77$)}
   & \footnotesize{($-11.86$, $-9.85$)}
   & \footnotesize{($-31.68$, $-32.09$)} & & \\
\midrule
\multirow{4}{*}{GNO}
 & In-distribution & 1.54 & 4.99 & 5.22 & \multirow{4}{*}{11.2 M} & \multirow{4}{*}{11.87 h} \\
 & & \footnotesize{($-39.61$, $-9.74$)}
   & \footnotesize{($-30.26$, $-7.41$)}
   & \footnotesize{($-29.31$, $-7.85$)} & & \\
 & Joint Gen.      & 4.75 & 11.41 & 13.46 & & \\
 & & \footnotesize{($-5.68$, $-3.79$)}
   & \footnotesize{($-11.39$, $-9.38$)}
   & \footnotesize{($-14.12$, $-14.64$)} & & \\
\bottomrule
\end{tabular}
\caption{Relative $L^2$ errors (\%) of PCNO and graph-based baselines on the base dataset. PCNO (Large) is comparable to the baselines in training time and PCNO (Small) in parameter count. Values in parentheses are $100 \times (e_{\mathrm{PCNO}} - e_{\mathrm{base}})/e_{\mathrm{base}}$ for PCNO (Large) and PCNO (Small), respectively, where $e$ is the relative $L^2$ error, so negative values favor PCNO. Time is the total training time in GPU-hours.}
\label{tab:surrogate_baselines}
\end{table}

\subsection{Limitations}
Several limitations remain. PCNO was trained on synthetic valves generated by FE simulations with noise-free input and output fields, and the chordae tendineae and papillary muscles, although present in every simulation, are not provided to the network as inputs. Robust deployment will require handling of input noise, validation on patient-derived geometries, and prospective clinical evaluation against measured outcomes. Training data uses the isotropic Lee--Sacks SED, whereas leaflet tissue is generally anisotropic. Because the constitutive model does not represent direction-dependent collagen recruitment during closure, the regional deformation and the location and magnitude of stress and strain concentrations may differ from those of native tissue, a distinction supported by prior comparisons of isotropic and anisotropic mitral leaflet models \cite{lee2014inverse}. This bears on the physiological interpretation of the predicted fields rather than on the agreement between PCNO and its FE ground truth: PCNO is supervised by solutions generated under the isotropic formulation and reproduces the nonlinear response within that framework, but it cannot recover fiber-dependent behavior absent from the training data. The architecture itself is not restricted to isotropic inputs, and the path to alternative constitutive families is discussed above; establishing accuracy under an anisotropic formulation would require separate training data and validation. PCNO predicts the mid-systolic configuration in a single forward pass rather than evolving the fields autoregressively over the cardiac cycle, which suits the peak-systolic geometric measures studied here but precludes time-resolved applications such as fluid-structure interaction or fatigue analysis. The model also outputs point estimates without associated uncertainty; calibrated uncertainty quantification, whether through ensembling, conformal prediction, or generative formulations, is important for clinical decision support \cite{psaros2023uncertainty,lopez2025uncertainty,ranftl2022stochastic}. Finally, geometry generalization remains the most challenging axis. While the transformer architecture flexibly handles diverse data modalities and unstructured inputs, it possesses comparatively weak inductive bias and therefore typically requires a large amount of training data to generalize well \cite{ebrahimi2026induction,dosovitskiy2020image,tay2023scaling}. Although our dataset is geometrically diverse, the variation in conditioning parameters substantially exceeds the variation across distinct geometries, making geometry generalization the harder regime for the surrogate. Future work will focus on expanding the geometric diversity of the training data, on architectural modifications that strengthen geometric inductive bias, and on extensions to patient-derived cases.

The scope of the dataset bounds these claims further. The geometries were produced by perturbing a single reference anatomy per valve type, not sampled to match measured population distributions, and every tricuspid geometry carries a three-leaflet topology, whereas two-, four-, and five-leaflet configurations are common clinically; the results therefore demonstrate multi-geometry learning within a controlled neighborhood of the reference anatomies rather than patient-level anatomical generalization. The disease-specific evaluation covers three mitral pathologies, so no performance claim is made for diseased tricuspid valves, and each pathology is represented by a single geometry. The prescribed annular condition is a uniform inward displacement, while patient-specific annular motion is spatially and temporally heterogeneous and can affect leaflet morphology and stress distributions \cite{rim2013effect}; other displacement magnitudes and image-derived annular trajectories were not evaluated and remain future work. The out-of-distribution tests extrapolate toward higher pressure and stiffer tissue. Extrapolation toward lower pressure and softer tissue, where the leaflets often deform beyond the range seen in training, is more challenging, and the accuracy reported here should not be assumed to extend to it.

Finally, the FE models treat the reference configuration as stress-free, whereas the mitral valve carries substantial prestrain in vivo; this limits the patient-specific interpretation of the predicted stress and strain fields, though it does not affect the demonstrated agreement with the FE solutions or the geometry-derived functional measures computed from the predicted configuration. Because the architecture accepts unstructured geometries and is not tied to a fixed mesh, image-derived geometries can be incorporated as such data become available without redesigning the operator.

\subsection{Toward Clinical Translation}

In the present study, pressure and material parameters are prescribed inputs. For patient-specific application, pressure loading could be estimated from clinically measured blood pressure together with valve-specific pressure gradients, or obtained from invasive measurement where available. Leaflet material parameters could be estimated through inverse calibration, adjusting the constitutive parameters until the predicted valve motion agrees with time-resolved images across multiple cardiac phases. The feasibility of noninvasive valve material estimation from 4D echocardiographic displacement data has recently been demonstrated using physics-informed neural networks \cite{wu2025noninvasive}. Because several parameter combinations may produce similar leaflet motion, such calibration should quantify uncertainty across plausible values. PCNO does not independently identify patient-specific material properties, but it can be embedded within image-based inverse calibration, where its low per-query cost allows many candidate parameter sets to be evaluated without repeating full FE simulations.


Patient-specific prestrain must likewise be estimated by calibrating a mechanical model against the valve configurations observed at end diastole and end systole. The reference configuration supplied to PCNO is the diastolic leaflet geometry used to initialize the simulation, so clinical application requires evaluating PCNO on image-derived diastolic geometries and testing its sensitivity to uncertainty in their reconstruction. This evaluation is separate from reconstructing the unobserved end-systolic coaptation region, which is an output of PCNO, not an input.

\subsection*{Conclusion}
Across in-distribution and out-of-distribution evaluations, PCNO reproduces the displacement, strain, and stress fields of the FE solutions, and the geometric measures derived from them, with a single model spanning mitral and tricuspid valves, mesh resolutions, and boundary conditions over a fixed set of geometries. The conditioning mechanism allows the model to generalize across pressure and tissue properties, including beyond the training range toward higher pressure and stiffer tissue, and the shared encoder supports auxiliary tasks through lightweight heads. Generalization to unseen geometries, uncertainty quantification, and validation against patient data remain open, as discussed above. Within these limits, PCNO provides a fast surrogate for exploring valve mechanics across loading and tissue properties, and a candidate forward model for the image-based calibration that patient-specific application would require.

\newpage
\section{Methods}

\subsection{Atrioventricular Valve Datasets}

Mitral and tricuspid valve closure was simulated as a dynamic problem in the open-source FE software FEBio \cite{maas2012febio,maas2017febio}. The base dataset spans two valve types, the mitral valve, which consists of two leaflets, and the tricuspid valve, which consists of three leaflets, and is composed of three sub-datasets: same geometry, diverse geometry, and moving boundary. The same geometry sub-dataset contains a single geometry per valve type with a fixed annulus, adopted from \cite{wu2022computational}. The diverse geometry sub-dataset extends this to ten distinct geometries per valve type, each also with a fixed annulus. The moving boundary sub-dataset mirrors the diverse geometry sub-dataset, covering the same ten mitral and ten tricuspid geometries, but replaces the fixed boundary condition with a prescribed annular displacement.

The prescribed displacement is spatially uniform, directed radially inward, and of fixed magnitude 2\,mm for every node on the annular curve. It is intended as a controlled and idealized moving-boundary case rather than a population-average measure of annular contraction, and its purpose is to extend the dataset beyond a fixed annulus so that a single operator is required to represent both fixed and prescribed annular boundary conditions across varying geometries. The perturbed geometries were generated by displacing the annular and free-edge curves of one normal reference geometry per valve type, as described below, so the resulting variability represents a controlled neighborhood of the two reference anatomies rather than a sample drawn to match measured population distributions of human valve morphology; in particular, all tricuspid geometries retain a three-leaflet topology.

In all simulations, the papillary muscles were assumed to be fixed in space, and the chordae tendineae were modeled as linear truss elements, while the valve leaflets were discretized using four-node linear quadrilateral shell elements, which can be straightforwardly triangulated when required for post-processing (e.g., volume computation). Chordal insertions were not distributed arbitrarily over the leaflet surface: for each papillary muscle tip, insertions were sampled within an anatomically constrained leaflet region associated with that tip, following the procedure detailed below. Papillary muscle tip positions were taken from the reference valve models of \cite{wu2022computational} and can in some cases be identified from clinical 3D imaging \cite{rim2013effect}. This construction follows \cite{khalighi2019development}, which showed that a simplified chordal topology with sufficient insertion density reproduces the closure, stress, and strain predicted by anatomically detailed chordal models.

\begin{figure}[H]
    \centering
    \includegraphics[width=1\linewidth]{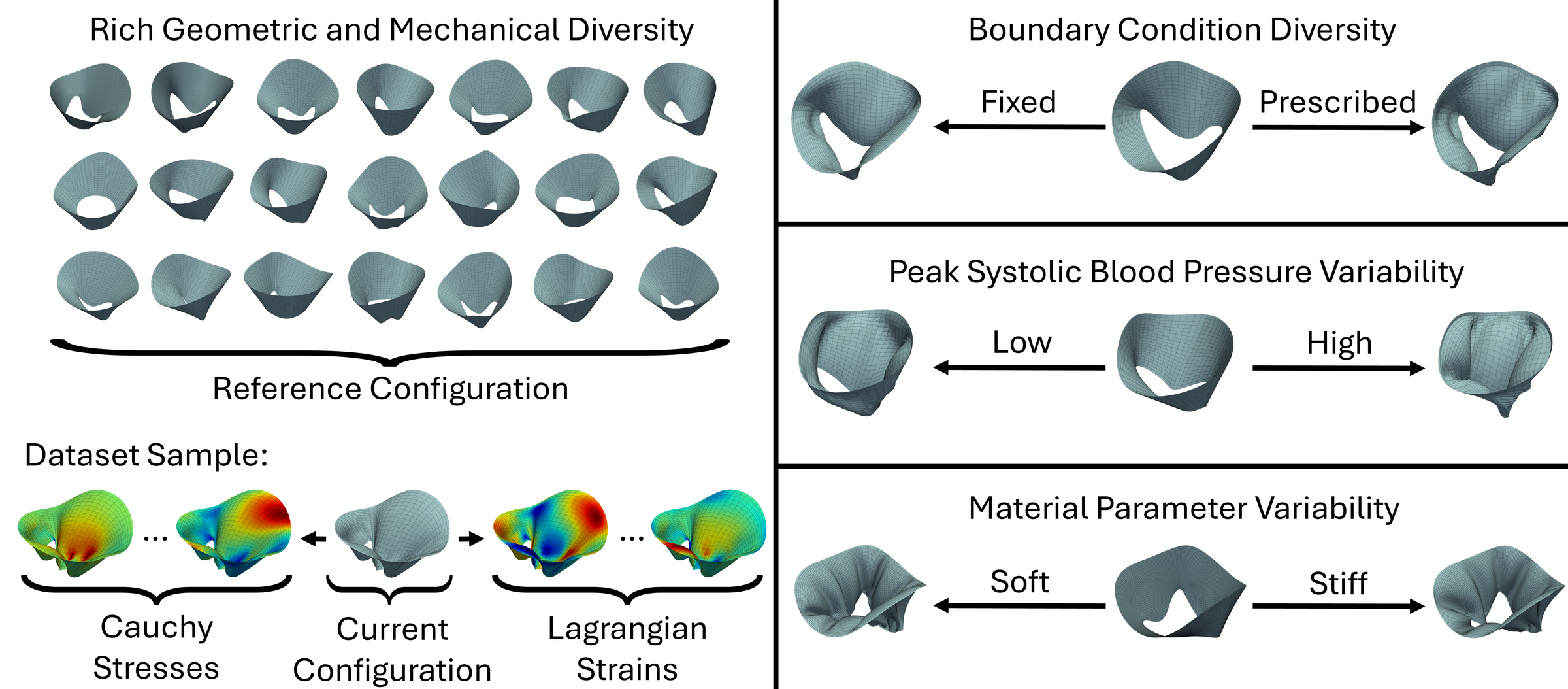}
    \caption{Visual overview of the base dataset. \textbf{Top left:} The dataset spans a wide variety of mitral and tricuspid valve geometries across the same geometry, diverse geometry, and moving boundary sub-datasets. \textbf{Right:} Sources of variability captured in the dataset, including fixed and prescribed annular boundary conditions, low-to-high peak systolic blood pressures, and soft-to-stiff material parameters. \textbf{Bottom left:} A representative dataset sample showing the ground truth current configuration alongside the corresponding stress and strain fields.}
    \label{fig:dataset_visual}
\end{figure}

A systolic transvalvular pressure was applied to the ventricular surface of the leaflet to simulate valve closure. Material parameters were randomly sampled from uniform distributions, and for each parameter set, simulations were performed across transvalvular pressures in increments of $0.25$ kPa. The range spans loading conditions relevant to both tricuspid and mitral valve closure: its lower portion includes the peak transvalvular pressure of $3.16$~kPa (23.7~mmHg) applied in a patient-specific tricuspid valve model \cite{kong2018finite}, and its upper end approaches the peak systolic pressure of $15.2$~kPa (114~mmHg) applied in a patient-specific mitral valve model \cite{pham2017finite}. Because the full range is applied to both valve types, each is also loaded outside its typical range, tricuspid valves up to 15~kPa and mitral valves down to 2~kPa, so that a single model covers the whole range for both. Pressure enters as a continuous conditioning variable, and the dataset is not partitioned into age-specific cohorts, so values in the upper portion of the range exceed the pressure a tricuspid valve normally experiences under typical physiological conditions. The framework is not inherently limited to this range and can be extended to higher-pressure regimes as additional training and validation data become available. Each sample consists of the reference (unloaded) and current (mid-systolic) nodal configurations, the corresponding displacement, Cauchy stress, and Lagrangian strain fields evaluated at the current configuration, and the corresponding physical conditioning variables $p$, $c_0$, $c_1$, $c_2$. An overview of the base dataset is provided in Table~\ref{tab:dataset_table} and visually in Fig.~\ref{fig:dataset_visual}.

\subsection*{Geometry Perturbation and Chordae Generation}
The annulus and free edge boundary curves of the reference leaflet surface were first extracted. Ten control points were sampled independently from each boundary. For each sampled point, a local neighborhood was defined using its $5$ nearest boundary points, and principal component analysis (PCA) was applied to estimate local geometric directions. Each sampled point was displaced along a randomly selected principal direction in the interval $[-1, 1]$~mm to produce smooth and physiologically realistic boundary curves. The perturbed annulus and free edge boundaries were then reconstructed using spline interpolation and uniformly resampled to $50$ points per boundary to enforce consistent discretization. A new quadrilateral leaflet surface was subsequently generated by lofting between the reconstructed boundaries.

For each generated valve geometry, PCA was computed to define a geometry-specific spatial basis and support robust correspondence to the reference configuration. Chordal insertion regions were identified on the new geometry by selecting the $200$ nearest surface points for each papillary muscle tip. From these points, $30$ insertions were randomly sampled to introduce physiologic variability in chordal distribution. Each papillary muscle tip was then connected to its sampled insertions to produce simulation-ready chordae structures for each perturbed geometry. 

\subsection*{Diseased Valve Geometries}

The tethering, P2 prolapse, and dilation geometries were adopted from our prior FE work \cite{wu2023effects}. All three were constructed from a single image-derived mitral valve model that was non-regurgitant and exhibited normal leaflet coaptation, and each introduces a distinct geometric mechanism of regurgitation: restricted leaflet motion, leaflet prolapse, and annular enlargement. Annular dilation was introduced by increasing the diameter of the annulus curve while leaving the leaflet lengths unchanged, so that the mismatch between the enlarged annulus and the available leaflet tissue reduces coaptation and produces a regurgitant gap. Leaflet tethering associated with ventricular remodeling was modeled by apically displacing the papillary muscle tips, which increased the distance to the leaflet insertion sites and restricted leaflet motion. The P2 prolapse model represented localized chordal dysfunction by reducing tension in the chordae supporting the middle scallop of the posterior leaflet. This idealized the diminished leaflet restraint associated with chordal elongation or rupture in mitral valve prolapse.

\begin{figure}[H]
    \centering
    \includegraphics[width=1\linewidth]{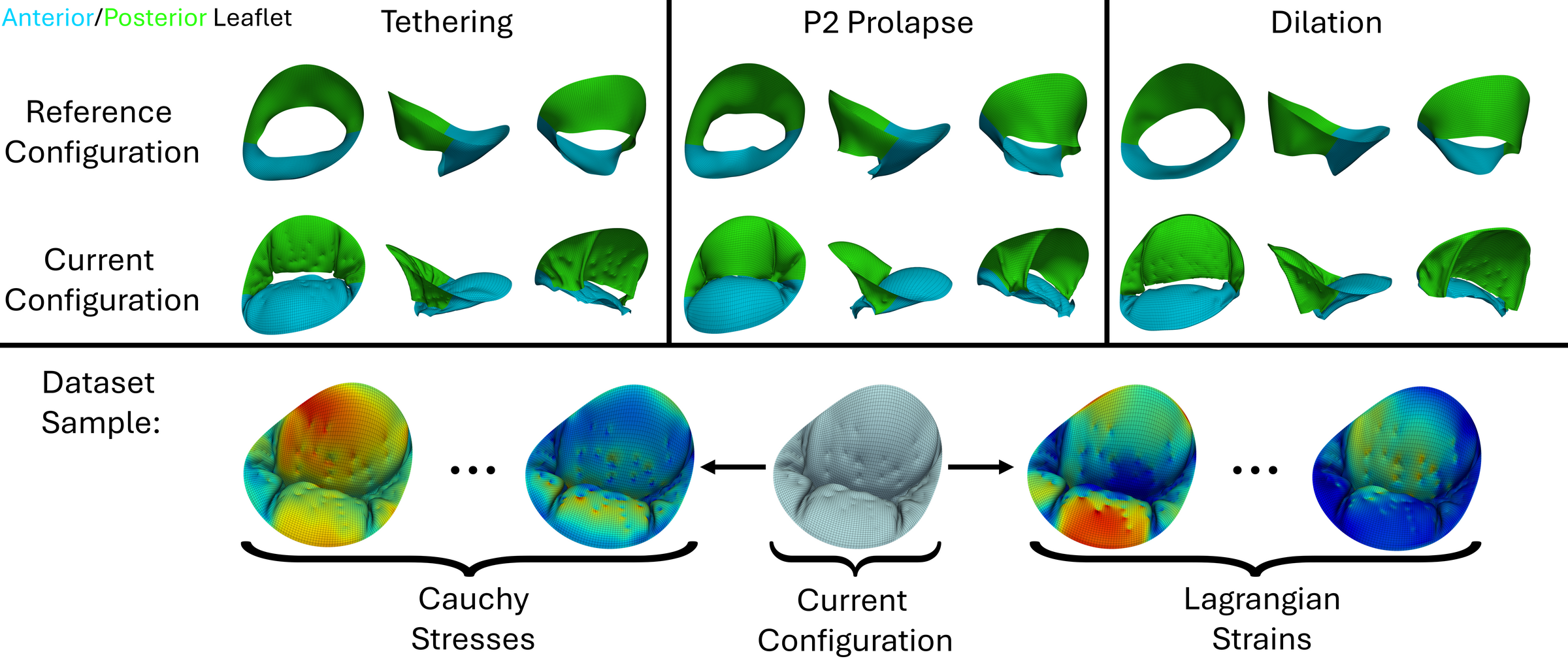}
    \caption{Visual overview of the diseased dataset. \textbf{Top:} Reference and current configurations for each pathology, shown from three viewpoints with anterior (green) and posterior (cyan) leaflets distinguished by color. Tethering exhibits leaflet displacement below the annular plane due to restricted leaflet motion, P2 prolapse shows the middle scallop of the posterior leaflet (P2 region) protruding above the annular plane, and annular dilation displays an increased area of the annular plane, which results in reduced leaflet coaptation. \textbf{Bottom:} A representative dataset sample showing the ground truth current configuration alongside the corresponding stress and strain fields.}
    \label{fig:disease_dataset_visual}
\end{figure}

All three pathologies are mitral, and no diseased tricuspid geometry is included, so the diseased dataset evaluates disease-specific performance for the mitral valve only. An overview of the diseased dataset is provided in Table~\ref{tab:disease_dataset_table} and visually in Fig.~\ref{fig:disease_dataset_visual}.

\begin{table}[H]
\centering
\begin{tabular}{lccccc}
\toprule
\multirow{2}{*}{} 
& \multicolumn{2}{c}{\textbf{Same Geometry}}
& \multicolumn{2}{c}{\textbf{Diverse Geometry / Moving Boundary}}
& \multirow{2}{*}{\textbf{Total}} \\
\cmidrule(lr){2-3}
\cmidrule(lr){4-5}
& \textbf{Mitral} & \textbf{Tricuspid}
& \textbf{Mitral} & \textbf{Tricuspid}
& \\
\midrule
Number of examples   & $31{,}613$ & $43{,}317$ & $69{,}557$ / $54{,}287$ & $72{,}545$ / $60{,}458$ & $331{,}777$ \\
Number of geometries & $1$ & $1$ & $10$ & $10$ & $22$\\
Number of nodes      & $11{,}917$ & $12{,}436$ & $1{,}029$ & $1{,}029$ & -- \\
Number of faces      & $11{,}689$ & $12{,}176$ & $980$ & $980$ & -- \\
Boundary condition   & Fixed & Fixed & Fixed / Prescribed & Fixed / Prescribed & -- \\
\bottomrule
\end{tabular}
\caption{Base dataset overview.}
\label{tab:dataset_table}
\end{table}

\begin{table}[H]
\centering
\begin{tabular}{lcccc}
\toprule
& \textbf{Tethering} & \textbf{P2 Prolapse} & \textbf{Dilation} & \textbf{Total} \\
\midrule
Number of examples   & $13{,}888$ & $13{,}273$ & $13{,}891$ & $41{,}052$ \\
Number of geometries & $1$ & $1$ & $1$ & $3$ \\
Number of nodes      & $8{,}651$ & $7{,}440$ & $7{,}471$ & --\\
Number of faces      & $8{,}440$ & $7{,}200$ & $7{,}230$ & -- \\
Boundary condition   & Fixed & Fixed & Prescribed & -- \\
\bottomrule
\end{tabular}
\caption{Diseased dataset overview.}
\label{tab:disease_dataset_table}
\end{table}

\subsection{Computation of Geometric Measures of Valve Function}

All geometric measures are computed from the current (mid-systolic) configuration, applying the same procedure to the ground-truth and predicted configurations. Many computations rely on a fitted annular surface, which serves as the reference datum from which heights, volumes, and distances are measured.

\paragraph{Annular surface fitting.}
Multiple conventions exist for defining the annular reference geometry, ranging from a least-squares plane fit \cite{blanke2014simplified} to smooth nonplanar surfaces spanning the annular contour, such as the minimal-area ``soap-film'' construction described by Salgo et al. \cite{salgo2002effect} and used in subsequent valve modeling work~ \cite{nguyen2019dynamic,nam2022dynamic,lasso2022slicerheart}. Because the mitral annulus is intrinsically saddle-shaped, a flat plane discards mechanically meaningful components of leaflet curvature \cite{salgo2002effect}. We adopt the same motivation but a different construction. The annular contour is expressed in a local frame given by the plane fit, and its out-of-plane coordinate is interpolated over the two in-plane coordinates using a thin-plate spline with no smoothing, so the fitted surface passes exactly through the annular nodes. The result is nonplanar and reproduces the saddle geometry, but it minimizes bending energy rather than area and is therefore not a minimal surface. Because the interpolant is fit on the annulus and evaluated at interior leaflet nodes, those evaluations constitute extrapolation of the fitted interpolant. The annular boundary condition is prescribed identically across cases within a pathology, so the ground-truth reference surface is fixed. The predicted annulus, by contrast, is an output of the model and varies slightly between cases, and the reference surface is therefore refit to the predicted annulus for each prediction, so that predicted and ground-truth measures are each evaluated against a surface derived from their own configuration.

The annular plane is first estimated by computing the centroid $\bar{\mathbf{x}}$ of the annulus node positions and performing a singular value decomposition of the mean-centered annulus coordinates. The plane normal $\mathbf{n}$ is taken as the right singular vector corresponding to the smallest singular value, with its orientation corrected so that $\mathbf{n}$ points toward the atrial side of the valve. Since the original mesh coordinates $(x, y, z)$ are defined in an arbitrary global frame with no special relationship to the annular geometry, we re-express all positions in a local coordinate system aligned with the annulus. An orthonormal basis $(\mathbf{u}, \mathbf{v})$ is constructed in the plane perpendicular to $\mathbf{n}$, so that $(\mathbf{u}, \mathbf{v}, \mathbf{n})$ forms a right-handed frame in which $\mathbf{u}$ and $\mathbf{v}$ parametrize position along the annular surface and $\mathbf{n}$ measures height above or below it. Each annulus node is projected onto this basis to obtain two-dimensional in-plane coordinates $(u_i, v_i)$ along with a signed out-of-plane height $w_i = (\mathbf{x}_i - \bar{\mathbf{x}}) \cdot \mathbf{n}$. A radial basis function interpolant $\hat{w}(u,v)$ is then fitted to the annulus data $\{(u_i, v_i, w_i)\}$, yielding a smooth nonplanar surface that interpolates the annular contour. For any query point $\mathbf{q}$, the projected height of each node above the annular surface, measured along the annular normal, is computed as

\begin{equation}
    d(\mathbf{q}) = (\mathbf{q} - \bar{\mathbf{x}}) \cdot \mathbf{n} - \hat{w}(u_{\mathbf{q}}, v_{\mathbf{q}}),
\end{equation}
where $(u_{\mathbf{q}}, v_{\mathbf{q}})$ are the in-plane coordinates of $\mathbf{q}$. Positive values of $d$ indicate displacement above the annular surface (toward the atrium), and negative values indicate displacement below it (toward the ventricle side of the valve). It is convenient to separate $d$ into its one-sided parts,
\begin{equation}
    d^{-}(\mathbf{q}) = \max\bigl(-d(\mathbf{q}),\, 0\bigr),
    \qquad
    d^{+}(\mathbf{q}) = \max\bigl(d(\mathbf{q}),\, 0\bigr),
\end{equation}
so that $d^{-}$ is the depth below the annular surface and $d^{+}$ the
height above it, each vanishing on the opposite side.

\paragraph{Tenting height.}
The tenting height quantifies the maximum depth to which the leaflets are displaced below the annular surface. Increased tenting height has been identified as a predictor of recurrent mitral regurgitation following annuloplasty in patients with heart failure \cite{ciarka2010predictors}. It is defined as
\begin{equation}
    \mathrm{TH} = \bigl| \min_{j}\, d(\mathbf{x}_j) \bigr|,
\end{equation}
where the minimum is taken over all mesh nodes, and the absolute value converts the negative projected height to a positive depth.

\begin{figure}[H]
    \centering
    \includegraphics[width=1\linewidth]{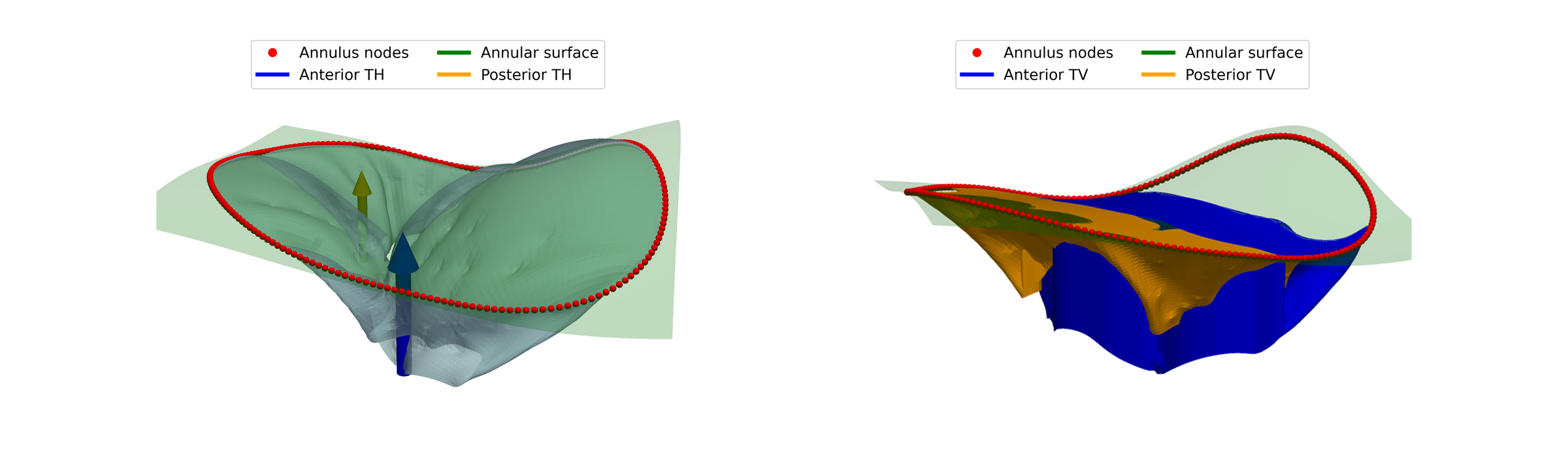}
    \caption{\textbf{Left:} Tenting height. \textbf{Right:} Tenting volume.}
    \label{fig:th_tv}
\end{figure}

\paragraph{Tenting volume.}
The tenting volume measures the total volume of the region enclosed between the leaflet surface and the annular surface on the ventricular side. Three-dimensional tenting volume has been identified as the only independent predictor of functional mitral regurgitation severity in multivariable analysis \cite{tibayan2007tenting}, and has been shown to correlate more strongly with coaptation area than either tenting height or tenting area \cite{mufarrih2023geometric}. The mesh is first triangulated, and for each triangular face with vertices $(\mathbf{p}_0, \mathbf{p}_1, \mathbf{p}_2)$, the projected area onto the annular plane is
\begin{equation}
    A_f = \frac{1}{2} \bigl| \bigl((\mathbf{p}_1 - \mathbf{p}_0) \times (\mathbf{p}_2 - \mathbf{p}_0)\bigr) \cdot \mathbf{n} \bigr|.
\end{equation}
Each face is weighted by the mean depth of its three vertices below the annular surface, and the tenting volume is approximated as
\begin{equation}
    \mathrm{TV} = \sum_{f=1}^{F} \frac{A_f}{3} \bigl( d^{-}(\mathbf{p}_0) + d^{-}(\mathbf{p}_1) + d^{-}(\mathbf{p}_2) \bigr),
\end{equation}
where the clipping in $d^{-}$ retains only the vertices lying on the ventricular side, so that leaflet regions above the annular surface contribute nothing.
The tenting height and tenting volume are visualized in Fig.~\ref{fig:th_tv}.

\paragraph{Prolapse height.}
The prolapse height quantifies the maximum extent to which leaflets are displaced above the annular surface. It is defined as
\begin{equation}
    \mathrm{PH} = \max_{j}\, d(\mathbf{x}_j),
\end{equation}
where the maximum is taken over all mesh nodes.

\begin{figure}[H]
    \centering
    \includegraphics[width=1\linewidth]{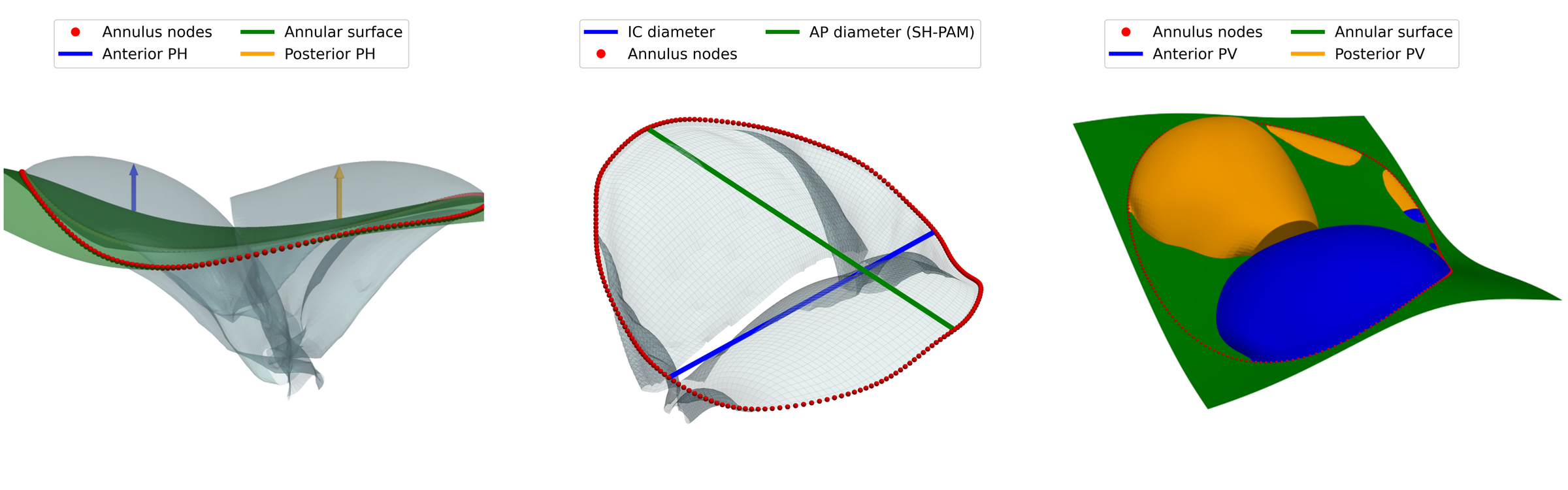}
    \caption{\textbf{Left:} Prolapse height. \textbf{Middle}: Anteroposterior and intercommissural diameters. \textbf{Right:} Prolapse volume.}
    \label{fig:ph_ap_ic_pv}
\end{figure}

\paragraph{Prolapse volume.}
The prolapse volume measures the total volume of the region enclosed between the leaflet surface and the annular surface on the atrial side. Using the same projected face area $A_f$ defined above, each face is weighted by the mean height of its three vertices above the annular surface, and the prolapse volume is approximated as
\begin{equation}
    \mathrm{PV} = \sum_{f=1}^{F} \frac{A_f}{3} \bigl( d^{+}(\mathbf{p}_0) + d^{+}(\mathbf{p}_1) + d^{+}(\mathbf{p}_2) \bigr).
\end{equation}
In both volumes, the clipping is applied at the vertices rather than by subdividing faces that straddle the annular surface, so such faces are treated approximately; the same functional is applied to the ground-truth and predicted configurations. The prolapse height and prolapse volume are visualized in Fig.~\ref{fig:ph_ap_ic_pv}.

\paragraph{Prolapse volume--prolapse height ratio.} 
The prolapse volume--prolapse height ratio provides a measure of the prolapse geometry by relating the total displaced volume above the annular surface to the maximum prolapse extent. It has been hypothesized \cite{kagiyama2017prolapse} that this ratio may be useful in differentiating degenerative mitral valve diseases such as Barlow's disease and fibroelastic deficiency. The ratio is defined as
\begin{equation}
    \text{PV--PH ratio} = \frac{\text{PV}}{\text{PH}}.
\end{equation}

\paragraph{Intercommissural diameter.}
The two commissure points are identified as the nodes that belong to both the annulus boundary and the shared boundary between the anterior and posterior leaflets. The intercommissural diameter is the Euclidean distance between these two points
\begin{equation}
    \mathrm{IC} = \|\mathbf{x}_{c_1} - \mathbf{x}_{c_2}\|_2,
\end{equation}
where $\mathbf{x}_{c_1}$ and $\mathbf{x}_{c_2}$ are the commissure point coordinates.


\paragraph{Anteroposterior diameter.}
The anteroposterior diameter is measured between two annulus landmarks: the saddle horn, the highest point of the annular contour relative to the annular plane, and the posterior annular midpoint, the point diametrically opposite it. Note that the reference here is the annular plane rather than the fitted annular surface, since the latter interpolates the annulus nodes exactly and assigns them zero height. The saddle horn is the node of greatest out-of-plane height,
\begin{equation}
    \mathbf{x}_{\mathrm{SH}} = \operatorname*{arg\,max}_{i} \; w_i,
    \qquad w_i = (\mathbf{x}_i - \bar{\mathbf{x}}) \cdot \mathbf{n},
\end{equation}
and, writing $\mathbf{p}_i = (\mathbf{x}_i - \bar{\mathbf{x}}) - w_i \mathbf{n}$ for the in-plane displacement, the posterior annular midpoint is the node whose in-plane direction is most nearly opposite that of the saddle horn,
\begin{equation}
    \mathbf{x}_{\mathrm{PAM}} = \operatorname*{arg\,min}_{i} \; \frac{\mathbf{p}_i \cdot \hat{\mathbf{p}}_{\mathrm{SH}}}{\|\mathbf{p}_i\|_2},
    \qquad \hat{\mathbf{p}}_{\mathrm{SH}} = \mathbf{p}_{\mathrm{SH}} / \|\mathbf{p}_{\mathrm{SH}}\|_2 .
\end{equation}
The anteroposterior diameter is the Euclidean distance between these two landmarks,
\begin{equation}
    \mathrm{AP} = \|\mathbf{x}_{\mathrm{SH}} - \mathbf{x}_{\mathrm{PAM}}\|_2 ,
\end{equation}
following the convention of \cite{munafo2024deep}. The anteroposterior and intercommissural diameters are visualized in Fig.~\ref{fig:ph_ap_ic_pv}.

\paragraph{Anteroposterior--intercommissural diameter ratio.}
The Anteroposterior--intercommissural diameter ratio characterizes the overall shape of the annulus by comparing its two principal dimensions. As degenerative mitral valve disease progresses, both the anteroposterior and intercommissural diameters increase, and the annulus transitions from a saddle-shaped to a flatter geometry \cite{viani2020mitral}. The ratio is defined as
\begin{equation}
    \text{AP--IC ratio} = \frac{\text{AP}}{\text{IC}}.
\end{equation}

\subsection{Evaluation Settings}

All experiments follow a common structure: a subset of the parameter space is designated for training, and the remainder is held out for testing. The base and diseased datasets span different parameter ranges, with the diseased ranges forming a strict subrange of the base ranges, so all settings below are defined relative to the bounds of the dataset in question. We write $p \in [p^{\ell}, p^{u}]$ for the SBP bounds and $c_i \in [c_i^{\ell}, c_i^{u}]$, $i = 0,1,2$, for the material parameter bounds of a given dataset. For the base dataset, $[p^{\ell}, p^{u}] = [2,15]$\,kPa, $[c_0^{\ell}, c_0^{u}] = [0,1000]$\,kPa, $[c_1^{\ell}, c_1^{u}] = [0,1000]$\,kPa, and $[c_2^{\ell}, c_2^{u}] = [0,10]$; for the diseased dataset, $[p^{\ell}, p^{u}] = [8,15]$\,kPa, $[c_0^{\ell}, c_0^{u}] = [200,800]$\,kPa, $[c_1^{\ell}, c_1^{u}] = [200,800]$\,kPa, and $[c_2^{\ell}, c_2^{u}] = [2,8]$.

\paragraph{In-distribution (base dataset).} The base dataset is randomly divided into 80\% training and 20\% testing subsets, with both covering the full ranges of SBP and material parameters of that dataset.

\paragraph{SBP generalization (base dataset).} The dataset is split by a pressure cutoff $p_{\max}$ with
\begin{equation}
    p^{\mathrm{Train}} = [p^{\ell}, p_{\max}], \quad p^{\mathrm{Test}} = (p_{\max}, p^{u}],
\end{equation}
and for the base dataset we set $p_{\max} = 10$\,kPa, so that $p^{\mathrm{Train}} = [2,10]$\,kPa and $p^{\mathrm{Test}} = (10,15]$\,kPa.

\paragraph{Material parameter generalization (base dataset).} 
We define the full material parameter domain
\begin{equation}
    \mathcal{X} = \Bigl\{(c_0,c_1,c_2)\in\mathbb{R}^3 \,:\, c_0^{\ell}\le c_0\le c_0^{u},\; c_1^{\ell}\le c_1\le c_1^{u},\; c_2^{\ell}\le c_2\le c_2^{u} \Bigr\},
\end{equation}
with training ranges $R_0 = [c_0^{\ell},c_{0_{\max}}]$, $R_1 = [c_1^{\ell},c_{1_{\max}}]$, $R_2 = [c_2^{\ell},c_{2_{\max}}]$, and the corresponding training subset $\mathcal{S}^{\mathrm{Train}} = \mathcal{X}\cap (R_0\times R_1\times R_2)$. All material parameter generalization experiments share this training subset; only the test partition changes. To define OOD regions within $\mathcal{X}$, we introduce the complements $\overline{R}_0 = (c_{0_{\max}},c_0^{u}]$, $\overline{R}_1 = (c_{1_{\max}},c_1^{u}]$, $\overline{R}_2 = (c_{2_{\max}},c_2^{u}]$. Three test subsets of increasing difficulty are then defined: (i) a univariate split in which exactly one parameter lies outside its training range,
\begin{equation}
    \mathcal{S}^{\mathrm{Test}}_{\mathrm{Uni}} = \mathcal{X}\cap \Bigl[(\overline{R}_0\times R_1\times R_2) \;\cup\; (R_0\times \overline{R}_1\times R_2) \;\cup\; (R_0\times R_1\times \overline{R}_2)\Bigr],
\end{equation}
(ii) a bivariate split in which exactly two are outside,
\begin{equation}
    \mathcal{S}^{\mathrm{Test}}_{\mathrm{Bi}} = \mathcal{X}\cap \Bigl[(\overline{R}_0\times \overline{R}_1\times R_2) \;\cup\; (\overline{R}_0\times R_1\times \overline{R}_2) \;\cup\; (R_0\times \overline{R}_1\times \overline{R}_2)\Bigr],
\end{equation}
and (iii) a trivariate split in which all three are simultaneously OOD,
\begin{equation}
    \mathcal{S}^{\mathrm{Test}}_{\mathrm{Tri}} = \mathcal{X}\cap (\overline{R}_0\times \overline{R}_1\times \overline{R}_2).
\end{equation}
For each split we let $c_{0_{\max}} = 750$\,kPa, $c_{1_{\max}} = 750$\,kPa, and $c_{2_{\max}} = 7.5$.

\paragraph{Joint SBP and material parameter generalization (base dataset).} 
Both the SBP and all three material parameters are simultaneously withheld from the training distribution,
\begin{equation}
 \mathcal{S}^{\mathrm{Test}}_{\mathrm{Joint}} = \mathcal{X} \cap (\overline{R}_0 \times \overline{R}_1 \times \overline{R}_2) \;\;\text{with}\;\; p \in p^{\mathrm{Test}}.
\end{equation}

\paragraph{In-distribution (diseased dataset).} The diseased dataset is randomly divided into 80\% training and 20\% testing subsets, with both covering the full ranges of SBP and material parameters of that dataset.

\paragraph{Joint SBP and material parameter generalization (diseased dataset).} The same joint generalization structure is applied to the diseased dataset, with $\mathcal{X}$, $R_i$, $\overline{R}_i$, and $p^{\mathrm{Train}}$, $p^{\mathrm{Test}}$ instantiated using the diseased dataset bounds given above. We set $p_{\max} = 13$\,kPa and material parameter training ranges $c_{0_{\max}} = 620$\,kPa, $c_{1_{\max}} = 640$\,kPa, and $c_{2_{\max}} = 6.2$, so that $p^{\mathrm{Train}} = [8,13]$\,kPa and $p^{\mathrm{Test}} = (13,15]$\,kPa, and the held-out material parameter regions are $\overline{R}_0 = (620,800]$\,kPa, $\overline{R}_1 = (640,800]$\,kPa, and $\overline{R}_2 = (6.2,8]$.

\subsection{Architecture Description}

Our proposed PCNO architecture predicts displacement, strain, and stress fields from geometric inputs and auxiliary conditioning variables. The model uses an encoder-style backbone in which both attention and feed-forward sublayers are modulated by global context through shift--scale--gate conditioning, conceptually related to adaptive layer norm style conditioning \cite{peebles2023scalable} and Feature-wise Linear Modulation \cite{perez2018film} (FiLM). The model takes as input a sequence of point-wise features
\begin{equation}
    \mathbf{x}_0\in\mathbb{R}^{N\times d_x},
\end{equation}
representing point-cloud coordinates and local geometric features, together with scalar conditioning variables
\begin{equation}
    \mathbf{c}_0\in\mathbb{R}^{1\times d_c},
\end{equation}
encoding physical parameters and global valve-related metadata. Both inputs are first projected into a latent space of dimension $d$
\begin{equation}
    \mathbf{x}=\mathrm{Dense}(\mathbf{x}_0),\qquad
    \mathbf{c}=\mathrm{Dense}(\mathbf{c}_0).
\end{equation}
For each encoder layer, a modulation network maps $\mathbf{c}$ to six vectors each of dimension $d$,
\begin{equation}
    (\beta_1,\gamma_1,\alpha_1,\beta_2,\gamma_2,\alpha_2)=\mathrm{ModNet}(\mathbf{c}),
\end{equation}
corresponding to shift ($\beta$), scale ($\gamma$), and residual gate ($\alpha$) parameters for the attention and feed-forward branches. Following the convention used in \cite{peebles2023scalable}, the $\mathrm{ModNet}$ and modulation operator are given by
\begin{equation}
    \mathrm{ModNet}(\mathbf{z}) = \mathrm{SiLU}(\mathrm{Dense}(\mathbf{z})), \quad \mathrm{modulate}(\mathbf{z},\beta,\gamma)=(1+\gamma)\odot\mathbf{z}+\beta,
\end{equation} 
where $\odot$ represents element-wise multiplication. Each encoder layer consists of an attention branch ($\mathrm{Attn}$), a point-wise feed-forward network ($\mathrm{FFN}$), and a layer norm ($\mathrm{LN}$). In the attention branch, we employ Rank-Augmented Linear Attention (RALA) \cite{fan2025breaking}. The resulting block update is
\begin{align}
    \mathbf{x} &\leftarrow \mathbf{x} + \alpha_1\odot
    \mathrm{Attn}\left(\mathrm{modulate}\left(\mathrm{LN}(\mathbf{x}),\beta_1,\gamma_1\right)\right),\\
    \mathbf{x} &\leftarrow \mathbf{x}+ \alpha_2\odot\mathrm{FFN}\!\left(\mathrm{modulate}\left(\mathrm{LN}(\mathbf{x}),\beta_2,\gamma_2\right)\right).
\end{align}
This design allows the conditioning variables to control both (i) feature statistics through the shift and scale modulation and (ii) update magnitude through residual gating. In particular, the gating coefficients $\alpha_1$ and $\alpha_2$ modulate the contribution of the attention and feed-forward branches to the residual update, enabling the network to smoothly vary how aggressively each layer transforms the latent representation as a function of the conditioning variables. The feed-forward network is a two-layer multi-layer perceptron (MLP) with Gaussian error linear unit (GELU) activation with expansion ratio $r$
\begin{equation}
\mathrm{FFN}(\mathbf{z})=W_2\mathrm{GELU}(W_1\mathbf{z}),
\qquad
W_1:d\rightarrow r\times d,\quad W_2:r\times d\rightarrow d.
\end{equation}
After the encoder stack, the latent features are processed by a conditioned MLP output head. Conditioning is applied at the head to allow the final mapping from latent features to physical quantities to adapt directly to the conditioning variables. A second modulation network produces $2L_o+2$ vectors for $L_o$ residual MLP layers, together with a final modulation pair
\begin{equation}
(\beta^{(1)},\gamma^{(1)},\ldots,\beta^{(L_o)},\gamma^{(L_o)},\beta^{(o)},\gamma^{(o)})=\mathrm{ModNet}(\mathbf{c}).
\end{equation}
Each residual head layer applies a modulated transformation followed by a residual update,
\begin{equation}\label{eq:mlp}
\mathbf{x}\leftarrow\mathbf{x}+\mathrm{GELU}\left(\mathrm{Dense}\left(\mathrm{modulate}\left(\mathrm{LN}(\mathbf{x}),\beta^{(\ell)},\gamma^{(\ell)}\right)\right)\right).
\end{equation}
The final prediction is obtained through a modulated linear projection
\begin{equation}
\mathbf{y}=\mathrm{Dense}\left(\mathrm{modulate}\left(\mathrm{LN}(\mathbf{x}),\beta^{(o)},\gamma^{(o)}\right)\right), \qquad \mathbf{y}\in\mathbb{R}^{N\times (3+6+6)}.
\end{equation}
These output channels are then partitioned into physically meaningful components
\begin{equation}
    \tilde{\mathbf{u}}=\mathbf{y}_{[:,0:3]},\qquad
    \tilde{\boldsymbol{E}}=\mathbf{y}_{[:,3:9]},\qquad
    \tilde{\boldsymbol{\sigma}}=\mathbf{y}_{[:,9:15]},
\end{equation}
where $\tilde{\mathbf{u}}\in\mathbb{R}^{3}$ denotes displacement, $\tilde{\boldsymbol{E}}\in\mathbb{R}^{6}$ denotes Lagrangian strain in Voigt form, and $\tilde{\boldsymbol{\sigma}}\in\mathbb{R}^{6}$ denotes Cauchy stress in Voigt form. This structured parameterization enables the simultaneous prediction of multiple coupled mechanical fields in a single forward pass. By sharing a common latent representation and a conditioned output head, the model can exploit correlations between displacement, strain, and stress while remaining flexible to variations in material parameters, loading conditions, and geometry. This design allows a single network to represent a broad and diverse family of mechanical responses without retraining.

\begin{figure}[H]
    \centering
    \includegraphics[width=1\linewidth]{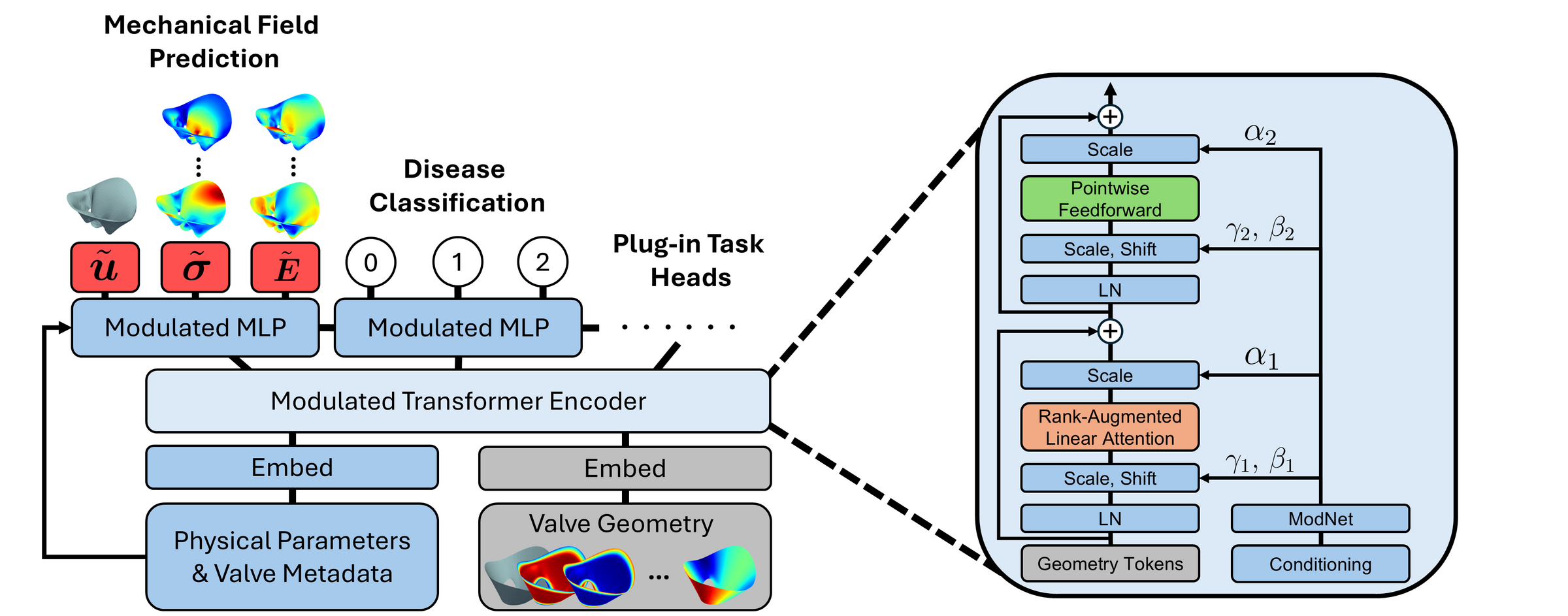}
    \caption{\textbf{PCNO architecture.} \textbf{Left:} The model takes as input point-cloud geometric features and scalar conditioning variables (physical parameters and valve metadata), which are independently embedded and passed through a modulated transformer encoder. A modulated MLP head decodes the latent representation into displacement $\tilde{\boldsymbol{u}}$, Cauchy stress $\tilde{\boldsymbol{\sigma}}$, and Lagrangian strain $\tilde{\boldsymbol{E}}$ fields simultaneously. Additional plug-in task heads can be attached to the same encoder representation to handle auxiliary tasks trained jointly (e.g., disease classification) or new downstream tasks without retraining the encoder. \textbf{Right:} Details of a single encoder layer. The conditioning vector is mapped through a ModNet to produce shift ($\beta$), scale ($\gamma$), and gate ($\alpha$) parameters that modulate both the attention and feedforward branches via shift--scale--gate conditioning applied after layer normalization. Residual connections are scaled by the gate parameters $\alpha_1$ and $\alpha_2$.}
    \label{fig:arch}
\end{figure}

In addition to predicting the mechanical fields on the diseased dataset, we equip the model with a classification head that identifies the disease pathology from the learned representation. The classification head takes as input a mean-pooled aggregation of the encoder output tokens and passes it through a conditioned MLP, identical in structure to the output head described in Eq.~\ref{eq:mlp}, with an output dimension equal to the number of classes. The model is trained with a cross-entropy loss over the class labels jointly with the field prediction losses, so that in a single forward pass, the network simultaneously predicts the displacement, strain, and stress fields alongside the disease classification. It is worth noting that the disease classification considered here is determined almost entirely by the input geometry, since each pathology corresponds to a distinct valve shape. From a purely geometric standpoint, the classification task may appear straightforward. However, the model does not have a dedicated geometric encoder for classification; rather, it must produce a single shared latent representation from which both the spatially resolved mechanical fields and the disease labels are simultaneously decoded. The encoder must therefore be expressive enough to support accurate point-wise field prediction while also encoding sufficient global geometric information in its pooled representation to distinguish between pathologies. The PCNO architecture is visualized in Fig.~\ref{fig:arch}.

\subsection{Local Geometric Features}

Using Cartesian coordinates $(x, y, z)$ alone is generally insufficient to encode meaningful local geometric structure, as they provide only absolute positional information and do not explicitly capture neighborhood-level shape or orientation. In particular, Cartesian coordinates do not uniquely determine local geometry: points with similar spatial locations may belong to neighborhoods with fundamentally different geometric configurations.

To enrich the geometric representation, we augment the raw coordinates with point-wise local features computed from a $k$-nearest-neighbor covariance matrix, commonly referred to as a structure tensor. For each point, a weighted covariance matrix is constructed over its local neighborhood, from which a set of eigenvalue- and eigenvector-based features is derived. These features encode intrinsic geometric properties such as linearity, planarity, surface variation, anisotropy, and dominant orientation, providing a compact and translation- and rotation-invariant\footnote{Note that verticality is not rotation-invariant; however, all other local features considered are.} description of local geometric structure. Such geometric features have been extensively explored in the context of 3D point-cloud object detection, classification, and reconstruction \cite{hackel2016contour, zheng2019adaptive, jager2025featuregs, slimani2024logdesc, demantke2011dimensionality, weinmann2015contextual, west2004context}, but have received comparatively little attention in the scientific machine learning literature. The resulting augmented input combines global positional information with locally invariant
geometric cues, enabling the model to better distinguish between different local structures and improving its ability to
learn geometry-dependent mappings.

\begin{figure}[H]
    \centering
    \includegraphics[width=1\linewidth]{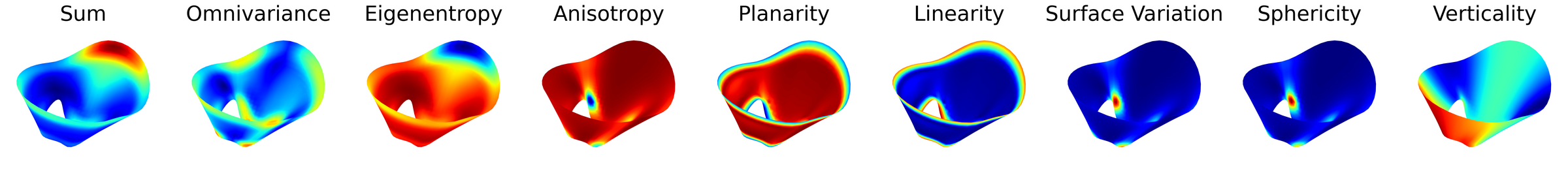}
    \caption{Visualization of the geometric eigen-features from the structure tensor $\Sigma$.}
    \label{fig:pca}
\end{figure}

These local geometric features are computed by first identifying the set of $k$-nearest neighbors of each mesh node $i$, denoted by $n \in \mathcal{N}_i^k$. Within each neighborhood, we compute a weighted mean position
\begin{equation}
    \mu_i = \frac{1}{S_i}\sum_{n\in\mathcal{N}_i^k} w_{in} X_n, \qquad S_i = \sum_{n\in\mathcal{N}_i^k} w_{in},
\end{equation}
where $X_n$ denotes the coordinates of neighbor node $n$. The weights $w_{in}$ are defined using a Gaussian kernel based on the Euclidean distance
\begin{equation}
    w_{in} = \exp\!\left( -\left( \frac{d_{in}}{\tau_i} \right)^2 \right), \qquad d_{in} = \|X_i - X_n\|_2, \qquad \tau_i = \mathrm{std}\!\left( \{ d_{in} \}_{n\in\mathcal{N}_i^k} \right).
\end{equation}
Using these weights, the structure tensor at node $i$ is constructed as
\begin{equation}
    \Sigma_i = \frac{1}{S_i} \sum_{n\in\mathcal{N}_i^k} w_{in} (X_n - \mu_i)(X_n - \mu_i)^T.
\end{equation}
The weights decay with increasing Euclidean distance, ensuring that nearby neighbors contribute more strongly to the local covariance. This reduces sensitivity to the choice of neighborhood size and prevents distant points from biasing the local geometric features. Given the structure tensor $\Sigma_i$, we compute its eigenvalues $(\eta_1, \eta_2, \eta_3)$ and corresponding eigenvectors $(\mathbf{e}_1, \mathbf{e}_2, \mathbf{e}_3)$, ordered such that $\eta_1 \ge \eta_2 \ge \eta_3$. These quantities are then used to construct the local geometric features, which are summarized in Table~\ref{tab:pca} and visualized in Fig.~\ref{fig:pca}. In all experiments, we set $k=96$.

\begin{table}[H]
\centering
\begin{tabular}{lc}
\toprule
\textbf{Features} & \textbf{Formula} \\
\midrule
Sum & $\eta_1 + \eta_2 + \eta_3$ \\
Omnivariance & $(\eta_1 \cdot \eta_2 \cdot \eta_3)^{1/3}$ \\
Eigenentropy & $-\sum_{i=1}^3\eta_i\cdot\ln{(\eta_i)}$ \\
Anisotropy & $(\eta_1-\eta_3)/\eta_1$ \\
Planarity & $(\eta_2-\eta_3)/\eta_1$ \\
Linearity & $(\eta_1-\eta_2)/\eta_1$ \\
Surface Variation & $\eta_3/(\eta_1+\eta_2+\eta_3)$ \\
Sphericity & $\eta_3/\eta_1$ \\
Verticality & $1 - |\langle [0\,0\,1], \mathbf{e}_3\rangle|$ \\
\bottomrule
\end{tabular}
\caption{Definitions of the geometric eigen-features from the structure tensor $\Sigma$, where $\langle\cdot,\cdot\rangle$ represents a dot-product.}
\label{tab:pca}
\end{table}

\subsection{Global Conditioning Variables}

In addition to local geometric features, we condition the model on a set of global auxiliary variables that encode physical parameters and coarse geometric properties of the domain. These variables are provided as scalar inputs and are used to modulate the model response across different loading conditions, material parameters, and geometric configurations. The physical conditioning variables include the SBP $p$ and the set of material parameters $(c_0, c_1, c_2)$. While these quantities strongly influence the mechanical response, they do not capture the geometric variability of the underlying surface. To incorporate global shape information in a form compatible with scalar conditioning, we augment the physical parameters with a set of global geometric features derived from integral curvature measures. Specifically, given a triangulated surface $S$ with faces indexed by $i = 1, \dots, F$ and face areas $A_i$, we compute the total surface area together with three integral curvature functionals
\begin{align}
    A &= \sum_{i=1}^F A_i 
    \quad \text{(Surface Area)}, \\
    \mathcal{W} &= \int_S H^2 \, dA 
    \;\approx\; \sum_{i=1}^F H_i^2 \, A_i 
    \quad \text{(Willmore Energy)}, \\
    M &= \int_S |H| \, dA 
    \;\approx\; \sum_{i=1}^F |H_i| \, A_i 
    \quad \text{(Total Mean Curvature)}, \\
    G &= \int_S |K| \, dA 
    \;\approx\; \sum_{i=1}^F |K_i| \, A_i 
    \quad \text{(Total Gaussian Curvature)},
\end{align}
where $H_i$ and $K_i$ denote the mean and Gaussian curvature associated with face $i$, respectively. These features provide low-dimensional summaries of global shape complexity, bending, and curvature distribution. Importantly, they allow the model to condition on geometric variation at the object level without introducing additional spatially varying inputs, complementing the local structure tensor features described in the previous section. These features are easily computable as a preprocessing step using \cite{sullivan2019pyvista}.

In addition to physical and geometric parameters, we include discrete categorical (binary) variables identifying the valve type $\mathcal{V}$ (mitral or tricuspid) and boundary condition class $\mathcal{B}$ (fixed or prescribed displacement), which capture known structural and kinematic differences across configurations. An overview of all conditioning variables is provided in
Table~\ref{tab:conditioning}. The full conditioning vector $\mathbf{c}$ is formed by concatenation

\begin{equation}
    \mathbf{c}_0
    = \mathrm{concat}(p, c_0, c_1, c_2, A, \mathcal{W}, M, G, \mathcal{V}, \mathcal{B})
    \in \mathbb{R}^{1\times10},
\end{equation}
and is provided to the model alongside the spatially varying inputs\footnote{Note that the diseased dataset consists of only mitral valves and so the valve type $\mathcal{V}$ was not included there.}. Supplementary Fig.~16 visualizes the distributions of $A, \mathcal{W}, M, G$ on the base dataset.

Compressing global geometric information into a small number of scalar conditioning variables has precedent across several domains. In aerodynamic surrogate modeling, Sung et al.~\cite{sung2025blendednet} condition a FiLM network on explicit geometric design parameters of the airframe together with flight conditions, modulating each layer by scale and shift. Lutheran et al.~\cite{lutheran2026diffusion} adopt a similar split for structural optimization, supplying spatial fields by concatenation while injecting global scalar descriptors such as the prescribed volume fraction through adaptive layer normalization. In computer vision, Saxena et al.~\cite{saxena2023zero} resolve depth-scale ambiguity by conditioning a diffusion model on a single geometric scalar, the vertical field of view, embedded sinusoidally and applied via FiLM layers. These cases motivate our use of integral curvature functionals as a compact global description of shape.

\begin{table}[H]
\centering
\begin{tabular}{lcc}
\toprule
\textbf{Variable} & \textbf{Type} & \textbf{Description} \\
\midrule
$p$   & Physical  & Peak Systolic Blood Pressure \\
$c_0$ & Physical  & Material parameter \\
$c_1$ & Physical  & Material parameter \\
$c_2$ & Physical  & Material parameter \\
$A$   & Geometric & Total surface area \\
$\mathcal{W}$ & Geometric & Willmore energy \\
$M$   & Geometric & Total mean curvature \\
$G$   & Geometric & Total Gaussian curvature \\
$\mathcal{V}$   & Categorical & Valve type (Mitral / Tricuspid) \\
$\mathcal{B}$   & Categorical & Boundary condition (Fixed / Prescribed) \\
\bottomrule
\end{tabular}
\caption{Auxiliary conditioning variables used to modulate the model.}
\label{tab:conditioning}
\end{table}

\subsection{Rank-Augmented Linear Attention}

The quadratic cost of softmax attention \cite{vaswani2017attention} with respect to sequence length has motivated a range of modifications aimed at reducing $\mathcal{O}(N^2)$ complexity to $\mathcal{O}(N)$, both in the computer vision and natural language processing communities \cite{katharopoulos2020transformers, han2024bridging, han2023flatten, wang2020linformer, choromanski2020rethinking} and in scientific machine learning \cite{hao2023gnot, zhdanov2025erwin, alkin2024universal, wu2024transolver, colagrande2025linear}. However, many linear attention variants have been found to exhibit low-rank attention maps \cite{han2023flatten, fan2025breaking} and output feature representations, limiting feature diversity and expressive spatial modeling compared to softmax attention. RALA \cite{fan2025breaking} addresses this limitation by building on the linear attention formulation of \cite{katharopoulos2020transformers} and introducing rank augmentation at two stages: a weighted construction of the key--value (KV) buffer and a token-dependent modulation of the output features. Softmax attention is given by 
\begin{equation}
    \mathrm{Attn}(Q, K, V) = \mathrm{Softmax}\left(\frac{QK^T}{\sqrt{d}}\right)V,
\end{equation}
where $Q,\,K,\,V\in\mathbb{R}^{N\times d}$ are linear embeddings of a latent input. For simplicity, we omit multi-head notation. Computing the attention scores $\mathrm{Softmax}(QK^T/\sqrt{d})$ requires $\mathcal{O}(N^2)$ operations, which becomes prohibitive for large sequence lengths (e.g., finer meshes). The generalized attention mechanism is given as
\begin{equation}\label{eq:linear_attention}
    Y_i = \sum_{j=1}^N\frac{\text{sim}(Q_i,K_j)}{\sum_{j=1}^N\text{sim}(Q_i,K_j)}V_j,
\end{equation}
where softmax attention can be recovered by letting the kernel $\text{sim}(Q,K)=\exp\left(QK^T/\sqrt{d}\right)$. The only constraint on $\text{sim}(\cdot)$ that needs to be imposed in Eq.~\ref{eq:linear_attention} to define an attention mechanism is non-negativity. This kernel can be written as
\begin{equation}
    \text{sim}(Q,K) = \phi(Q)\phi(K)^T,
\end{equation}
and we obtain
\begin{equation}\label{eq:full_linear_attention}
    Y_i = \sum_{j=1}^N\frac{\text{sim}(Q_i,K_j)}{\sum_{j=1}^N\text{sim}(Q_i,K_j)}V_j = \sum_{j=1}^N\frac{\phi(Q_i)\phi(K_j)^T}{\sum_{j=1}^N\phi(Q_i)\phi(K_j)^T}V_j = \frac{\phi(Q_i)\left(\sum_{j=1}^N \phi(K_j)^T V_j\right)}{\phi(Q_i)\left(\sum_{j=1}^N\phi(K_j)^T\right)}.
\end{equation}
Note that the computation order has changed from $(\phi(Q)\phi(K)^T)V$ to $\phi(Q)(\phi(K)^TV)$, allowing for computation in linear complexity. Common choices for $\phi(\cdot)$ are $\mathrm{ReLU}(\cdot)$ and $\mathrm{elu}(\cdot)+1$; we use the latter in all experiments. Each $\phi(Q_i)$ is multiplied by the KV buffer $\left(\sum_{j=1}^N \phi(K_j)^T V_j\right)$, and so the diversity of the output features $Y_i$ are reflected by the rank of the KV buffer. Note that the denominator $\phi(Q_i)\left(\sum_{j=1}^N\phi(K_j)^T\right)\in\mathbb{R}^{1\times 1}$ and does not affect the rank of the output features, hence, its analysis is omitted.

In Eq.~\ref{eq:full_linear_attention}, a uniform weight of $1$ is assigned to each token in the buffer. RALA instead introduces token-dependent weights by assigning higher weights to tokens that contain more information
\begin{equation}
    B_{KV} = \sum_{j=1}^N\omega_j \phi(K_j)^TV_j, \quad \sum_{j=1}^N\omega_j=1,
\end{equation}
where $B_{KV}$ represents the KV buffer and $\omega_j$ is the weight coefficient of the $j$-th token. The design of the weights is inspired by softmax attention: instead of computing a score between every query and every key, a global (mean-pooled) query is computed, and all keys are subsequently compared to it

\begin{equation}
    Q_g = \frac{1}{N}\sum_{j=1}^N\phi(Q_j), \quad \omega_j = N\times \frac{\exp\left(Q_g\phi(K_j)^T\right)}{\sum_{m=1}^N\exp\left(Q_g\phi(K_m)^T\right)} = N\times\mathrm{Softmax}\left(Q_g \phi(K)^T\right),
\end{equation}
which can still be computed in linear complexity. The introduction of $\omega_j$ increases the rank of the KV buffer by breaking the uniform aggregation of token features, allowing different tokens to contribute uniquely and reducing linear dependencies in the resulting representation. Lastly, the output features are further augmented by a post-processing step on each token $Y_i$ such that

\begin{equation}
    Y_i = \mathrm{Dense}(X_i)\odot\frac{\phi(Q_i)\left(\sum_{j=1}^N\omega_j\phi(K_j)^TV_j\right)}{\phi(Q_i)\left(\sum_{j=1}^N\phi(K_j)^T\right)},
\end{equation}
further transitioning a possible low-rank state to a full-rank state due to the fact that for two similarly shaped matrices $A,\,B$, $\mathrm{rank}(A\odot B)\leq \mathrm{rank}(A)\times\mathrm{rank}(B)$. That is, the Hadamard product can raise the upper bound of the matrix rank.

\subsection{Training Details}

PCNO was implemented in JAX \cite{jax2018github}, Flax \cite{flax2020github}, and Optax \cite{deepmind2020jax}. In all experiments, we employ the Muon optimizer \cite{jordan2024muon} for two-dimensional learnable parameters and Adam \cite{loshchilov2017decoupled,kingma2014adam} for one-dimensional learnable parameters. For Muon, we set $\beta=0.95$, and use 5 Newton-Schulz iterations with coefficients $(3.4445,\,-4.775,\,2.0315)$. For Adam, we set $\beta_1=0.9$ and $\beta_2=0.999$. Training is performed for 200{,}000 optimization steps. We use a learning rate scheduler consisting of a linear warm-up from $10^{-8}$ to a peak of $10^{-3}$ over 2{,}000 steps, followed by an exponential decay with a rate of $0.99$ over $7.5 \times 10^{2}$ transition steps, with a minimum learning rate of $10^{-8}$. To stabilize training \cite{pascanu2013difficulty}, all gradients are clipped to a maximum global $L^2$-norm of $1$. Following the initialization strategy of \cite{peebles2023scalable}, all weights in the modulation networks, as well as the final layer of the output MLP head(s), are initialized to zero. All remaining weights are initialized using a Xavier/Glorot uniform initializer \cite{glorot2010understanding}, and all biases are initialized to zero. In all experiments, we use a batch size of 4 and weight decay of $10^{-4}$, with two exceptions: the Joint SBP and Material Parameter Generalization setting on the base dataset uses a weight decay of $10^{-1}$, and on the diseased dataset a batch size of 2 and weight decay of $5 \times 10^{-1}$ is used. These settings were chosen to mitigate overfitting, given the severity of the OOD shift combined with the limited training data available in these settings. For all evaluations, we follow \cite{morales2024exponential} and use an exponential moving average (EMA) of the model parameters. Let $\theta_s$ denote the parameters at training step $s$; the EMA parameters $\theta_s^{\mathrm{EMA}}$ are computed as
\begin{equation}
    \theta^{\mathrm{EMA}}_0 = \theta_0, \quad
    \theta^{\mathrm{EMA}}_{s+1} = \nu \theta^{\mathrm{EMA}}_s + (1-\nu)\theta_{s+1}, \quad s = 0,\dots,S-1,
\end{equation}
where $S$ denotes the final training step and $\nu$ is the momentum parameter, set to $0.999$ in all experiments.

Due to varying mesh sizes across datasets, we apply zero padding so that all inputs share a common shape. Alongside the padded inputs, the model is provided with a one-dimensional binary mask of shape $(N{,})$ that indicates whether a token corresponds to a valid mesh node or a padded (ghost) node. The use of masking is particularly important for the attention mechanism. Unlike other components of the model, which operate independently on each token, the attention mechanism explicitly aggregates information across all input tokens. Without proper masking, padded nodes---despite being initialized as zeros---can acquire non-zero values after passing through linear layers and nonlinearities. These ghost nodes would then contribute spurious terms to the attention scores, distorting the attention weights and degrading the learned representations of valid nodes.

To prevent this, the binary mask is applied directly to the query and key tensors within the attention mechanism, ensuring that padded nodes do not contribute feature content to the attention computation. However, masking the queries and keys alone is not sufficient when attention weights are computed using a softmax normalization. Even zero-valued logits corresponding to padded nodes can receive non-zero probability mass due to the global normalization performed by the softmax. To fully exclude padded nodes from the attention mechanism, we additionally mask the attention logits prior to the softmax operation by adding a large negative constant to the logits associated with padded positions, effectively setting them to $-\infty$. This guarantees that padded nodes receive zero attention weight after normalization and therefore do not participate in the attention aggregation. As a result, attention is computed exclusively over valid mesh nodes, preserving the fidelity and interpretability of the learned node representations.

To improve computational efficiency during training, we randomly sample $3{,}000$ nodes from each mesh for every element of the batch. In the same geometry sub-dataset, the total number of mesh nodes always exceeds $3{,}000$, and therefore no zero padding is required. In contrast, the diverse geometry and moving boundary sub-datasets contain fewer than $3{,}000$ nodes per mesh, so zero padding (and corresponding masking) is applied only to samples drawn from these datasets. For the diseased dataset, the total number of mesh nodes always exceeds $3{,}000$, and therefore no zero padding is required.

Normalization statistics are computed only from the training split and then reused unchanged for the test split. For nodal (local) features, the mean and standard deviation are computed feature-wise by pooling values over all nodes across all training meshes/geometries. For conditioning (global) features, the mean and standard deviation are computed feature-wise over the training set of conditioning vectors. Any categorical variables (e.g., valve type or boundary condition IDs) are left unnormalized.

We found that training using a mean absolute error (MAE) over a mean squared error loss function achieved noticeably improved performance. This is likely attributed to MAE being more robust towards outliers, such as strong stress or strain concentrations. Our loss function between the model predictions and the corresponding targets is described below
\begin{equation}
    \mathcal{L} = \frac{1}{N_b}\frac{1}{N}\frac{1}{3}\sum_{i=1}^{N_b}\sum_{j=1}^{N}\sum_{k=1}^{3}|\tilde{\boldsymbol{u}}_{ijk}-\boldsymbol{u}_{ijk}| + \frac{1}{N_b}\frac{1}{N}\frac{1}{6}\sum_{i=1}^{N_b}\sum_{j=1}^{N}\sum_{k=1}^{6}|\tilde{\boldsymbol{E}}_{ijk}-\boldsymbol{E}_{ijk}| + \frac{1}{N_b}\frac{1}{N}\frac{1}{6}\sum_{i=1}^{N_b}\sum_{j=1}^{N}\sum_{k=1}^{6}|\tilde{\boldsymbol{\sigma}}_{ijk}-\boldsymbol{\sigma}_{ijk}|,
\end{equation}
where where $N_b$ is the batch size and $\Box_{ijk}$ denotes the $i$-th sample in the batch, $j$-th node and $k$-th feature, for $\Box \in\{\boldsymbol{u},\boldsymbol{E},\,\boldsymbol{\sigma}\}$. When training on the diseased dataset, this loss function is augmented with a multi-class cross-entropy term over the disease labels
\begin{equation}
    \mathcal{L}_{\mathrm{cls}} = -\frac{1}{N_b}\sum_{i=1}^{N_b}\sum_{c=1}^{C} y_{ic}\log \hat{y}_{ic},
\end{equation}
where $C$ is the number of classes, $y_{ic}$ is the one-hot encoded ground-truth label, and $\hat{y}_{ic} = \mathrm{softmax}(\mathbf{z}_i)_c$ denotes the predicted probability for class $c$ obtained from the classification head logits $\mathbf{z}_i$. The total loss for the diseased dataset is then $\mathcal{L} + \mathcal{L}_{\mathrm{cls}}$. Supplementary Table~3 provides an ablation study over model sizes (as described in Supplementary Table~2), the EMA parameters, loss functions, and attention mechanisms, and Supplementary Figs.~12--15 show the corresponding relative $L^2$ error curves over training steps for each ablation. 

\newpage
\section{Declaration Statements}
\subsection{Data Availability}
The authors are committed to open-sourcing the data to support reproducibility. Data associated with this work is available from the authors upon reasonable request and will be made publicly available in an open-access repository upon acceptance of the manuscript.

\subsection{Code Availability}
The authors are committed to open-sourcing the code to support reproducibility. The code associated with this work will be made publicly available in an open-source repository upon acceptance of the manuscript.

\subsection{Acknowledgments}
We gratefully acknowledge the following funding sources: the U.S. Department of Energy, Office of Science, Advanced Scientific Computing Research program, under Award No. DE-SC0024563; Two Additional Ventures' Single Ventricle Research Fund; National Institutes of Health (NHLBI K25 HL168235 and R01 HL153166); the Topolewski Endowed Chair in Pediatric Cardiology; and the Topolewski Pediatric Valve Center. The authors thank the Children's Hospital of Philadelphia Research Institute for generously providing the CPU resources necessary to support this project. We also thank the developers of the software that enabled our research, including FEBio \cite{maas2012febio,maas2017febio}, JAX \cite{jax2018github}, Flax \cite{flax2020github}, Optax \cite{deepmind2020jax}, NumPy \cite{harris2020array}, SciPy \cite{2020SciPy-NMeth}, Einops \cite{rogozhnikov2022einops}, Weights \& Biases \cite{wandb}, ML Collections \cite{ml_collections}, Matplotlib \cite{Hunter:2007}, PyVista \cite{sullivan2019pyvista}, and the Visualization Toolkit (VTK) \cite{vtkBook}.  

\subsection{Author Contributions}
P.P. and W.W. conceived and supervised the study and provided conceptual guidance. S.K. developed the methodology, implemented the model, performed all experiments, and wrote the manuscript. W.W. generated the datasets. M.A.J. supplied clinical, physiologic, and valve pathology input, as well as the model framework for training data. All authors provided critical revisions of the manuscript.

\subsection{Competing Interests}
All authors declare no financial or non-financial competing interests. 

\newpage
\addcontentsline{toc}{section}{References}
\bibliographystyle{plain}
\bibliography{arxiv_v2/references}

@article{vaswani2017attention,
  title={Attention is all you need},
  author={Vaswani, Ashish and Shazeer, Noam and Parmar, Niki and Uszkoreit, Jakob and Jones, Llion and Gomez, Aidan N and Kaiser, {\L}ukasz and Polosukhin, Illia},
  journal={Advances in neural information processing systems},
  volume={30},
  year={2017}
}

@article{choromanski2020rethinking,
  title={Rethinking attention with performers},
  author={Choromanski, Krzysztof and Likhosherstov, Valerii and Dohan, David and Song, Xingyou and Gane, Andreea and Sarlos, Tamas and Hawkins, Peter and Davis, Jared and Mohiuddin, Afroz and Kaiser, Lukasz and others},
  journal={arXiv preprint arXiv:2009.14794},
  year={2020}
}

@inproceedings{fan2025breaking,
  title={Breaking the low-rank dilemma of linear attention},
  author={Fan, Qihang and Huang, Huaibo and He, Ran},
  booktitle={Proceedings of the Computer Vision and Pattern Recognition Conference},
  pages={25271--25280},
  year={2025}
}

@inproceedings{katharopoulos2020transformers,
  title={Transformers are rnns: Fast autoregressive transformers with linear attention},
  author={Katharopoulos, Angelos and Vyas, Apoorv and Pappas, Nikolaos and Fleuret, Fran{\c{c}}ois},
  booktitle={International conference on machine learning},
  pages={5156--5165},
  year={2020},
  organization={PMLR}
}

@article{han2024bridging,
  title={Bridging the divide: Reconsidering softmax and linear attention},
  author={Han, Dongchen and Pu, Yifan and Xia, Zhuofan and Han, Yizeng and Pan, Xuran and Li, Xiu and Lu, Jiwen and Song, Shiji and Huang, Gao},
  journal={Advances in Neural Information Processing Systems},
  volume={37},
  pages={79221--79245},
  year={2024}
}

@inproceedings{han2023flatten,
  title={Flatten transformer: Vision transformer using focused linear attention},
  author={Han, Dongchen and Pan, Xuran and Han, Yizeng and Song, Shiji and Huang, Gao},
  booktitle={Proceedings of the IEEE/CVF international conference on computer vision},
  pages={5961--5971},
  year={2023}
}

@article{wang2020linformer,
  title={Linformer: Self-attention with linear complexity},
  author={Wang, Sinong and Li, Belinda Z and Khabsa, Madian and Fang, Han and Ma, Hao},
  journal={arXiv preprint arXiv:2006.04768},
  year={2020}
}

@inproceedings{perez2018film,
  title={Film: Visual reasoning with a general conditioning layer},
  author={Perez, Ethan and Strub, Florian and De Vries, Harm and Dumoulin, Vincent and Courville, Aaron},
  booktitle={Proceedings of the AAAI conference on artificial intelligence},
  volume={32},
  year={2018}
}

@inproceedings{peebles2023scalable,
  title={Scalable diffusion models with transformers},
  author={Peebles, William and Xie, Saining},
  booktitle={Proceedings of the IEEE/CVF international conference on computer vision},
  pages={4195--4205},
  year={2023}
}

@inproceedings{hackel2016contour,
  title={Contour detection in unstructured 3D point clouds},
  author={Hackel, Timo and Wegner, Jan D and Schindler, Konrad},
  booktitle={Proceedings of the IEEE conference on computer vision and pattern recognition},
  pages={1610--1618},
  year={2016}
}

@article{zheng2019adaptive,
  title={An adaptive end-to-end classification approach for mobile laser scanning point clouds based on knowledge in urban scenes},
  author={Zheng, Mingxue and Wu, Huayi and Li, Yong},
  journal={Remote Sensing},
  volume={11},
  number={2},
  pages={186},
  year={2019},
  publisher={MDPI}
}

@article{jager2025featuregs,
  title={FeatureGS: Eigenvalue-Feature Optimization in 3D Gaussian Splatting for Geometrically Accurate and Artifact-Reduced Reconstruction},
  author={J{\"a}ger, Miriam and Hillemann, Markus and Jutzi, Boris},
  journal={arXiv preprint arXiv:2501.17655},
  year={2025}
}

@inproceedings{slimani2024logdesc,
  title={LoGDesc: Local geometric features aggregation for robust point cloud registration},
  author={Slimani, Karim and Tamadazte, Brahim and Achard, Catherine},
  booktitle={Proceedings of the Asian Conference on Computer Vision},
  pages={1952--1968},
  year={2024}
}

@inproceedings{demantke2011dimensionality,
  title={Dimensionality based scale selection in 3D lidar point clouds},
  author={Demantk{\'e}, J{\'e}r{\^o}me and Mallet, Cl{\'e}ment and David, Nicolas and Vallet, Bruno},
  booktitle={Laserscanning},
  year={2011}
}

@article{weinmann2015contextual,
  title={Contextual classification of point cloud data by exploiting individual 3D neigbourhoods},
  author={Weinmann, Martin and Schmidt, Alena and Mallet, Cl{\'e}ment and Hinz, Stefan and Rottensteiner, Franz and Jutzi, Boris},
  journal={ISPRS Annals of the Photogrammetry, Remote Sensing and Spatial Information Sciences; II-3/W4},
  volume={2},
  pages={271--278},
  year={2015},
  publisher={G{\"o}ttingen: Copernicus GmbH}
}

@inproceedings{west2004context,
  title={Context-driven automated target detection in 3D data},
  author={West, Karen F and Webb, Brian N and Lersch, James R and Pothier, Steven and Triscari, Joseph M and Iverson, A Evan},
  booktitle={Automatic Target Recognition XIV},
  volume={5426},
  pages={133--143},
  year={2004},
  organization={SPIE}
}

@misc{jordan2024muon,
  author       = {Keller Jordan and Yuchen Jin and Vlado Boza and You Jiacheng and
                  Franz Cesista and Laker Newhouse and Jeremy Bernstein},
  title        = {Muon: An optimizer for hidden layers in neural networks},
  year         = {2024},
  url          = {https://kellerjordan.github.io/posts/muon/}
}

@inproceedings{hao2023gnot,
  title={Gnot: A general neural operator transformer for operator learning},
  author={Hao, Zhongkai and Wang, Zhengyi and Su, Hang and Ying, Chengyang and Dong, Yinpeng and Liu, Songming and Cheng, Ze and Song, Jian and Zhu, Jun},
  booktitle={International Conference on Machine Learning},
  pages={12556--12569},
  year={2023},
  organization={PMLR}
}

@article{zhdanov2025erwin,
  title={Erwin: A tree-based hierarchical transformer for large-scale physical systems},
  author={Zhdanov, Maksim and Welling, Max and van de Meent, Jan-Willem},
  journal={arXiv preprint arXiv:2502.17019},
  year={2025}
}

@article{alkin2024universal,
  title={Universal physics transformers: A framework for efficiently scaling neural operators},
  author={Alkin, Benedikt and F{\"u}rst, Andreas and Schmid, Simon and Gruber, Lukas and Holzleitner, Markus and Brandstetter, Johannes},
  journal={Advances in Neural Information Processing Systems},
  volume={37},
  pages={25152--25194},
  year={2024}
}

@article{wu2024transolver,
  title={Transolver: A fast transformer solver for pdes on general geometries},
  author={Wu, Haixu and Luo, Huakun and Wang, Haowen and Wang, Jianmin and Long, Mingsheng},
  journal={arXiv preprint arXiv:2402.02366},
  year={2024}
}

@article{colagrande2025linear,
  title={Linear Attention with Global Context: A Multipole Attention Mechanism for Vision and Physics},
  author={Colagrande, Alex and Caillon, Paul and Feillet, Eva and Allauzen, Alexandre},
  journal={arXiv preprint arXiv:2507.02748},
  year={2025}
}

@article{kingma2014adam,
  title={Adam: A method for stochastic optimization},
  author={Kingma, Diederik P},
  journal={arXiv preprint arXiv:1412.6980},
  year={2014}
}

@article{loshchilov2017decoupled,
  title={Decoupled weight decay regularization},
  author={Loshchilov, Ilya and Hutter, Frank},
  journal={arXiv preprint arXiv:1711.05101},
  year={2017}
}

@inproceedings{glorot2010understanding,
  title={Understanding the difficulty of training deep feedforward neural networks},
  author={Glorot, Xavier and Bengio, Yoshua},
  booktitle={Proceedings of the thirteenth international conference on artificial intelligence and statistics},
  pages={249--256},
  year={2010},
  organization={JMLR Workshop and Conference Proceedings}
}

@article{morales2024exponential,
  title={Exponential moving average of weights in deep learning: Dynamics and benefits},
  author={Morales-Brotons, Daniel and Vogels, Thijs and Hendrikx, Hadrien},
  journal={arXiv preprint arXiv:2411.18704},
  year={2024}
}

@inproceedings{pascanu2013difficulty,
  title={On the difficulty of training recurrent neural networks},
  author={Pascanu, Razvan and Mikolov, Tomas and Bengio, Yoshua},
  booktitle={International conference on machine learning},
  pages={1310--1318},
  year={2013},
  organization={Pmlr}
}

@article{maas2012febio,
    author = {Maas, Steve A. and Ellis, Benjamin J. and Ateshian, Gerard A. and Weiss, Jeffrey A.},
    title = {FEBio: Finite Elements for Biomechanics},
    journal = {Journal of Biomechanical Engineering},
    volume = {134},
    number = {1},
    pages = {011005},
    year = {2012},
    month = {02},
    issn = {0148-0731},
    doi = {10.1115/1.4005694},
    url = {https://doi.org/10.1115/1.4005694},
    eprint = {https://asmedigitalcollection.asme.org/biomechanical/article-pdf/134/1/011005/5665064/011005_1.pdf},
}

@article{sullivan2019pyvista,
  doi = {10.21105/joss.01450},
  url = {https://doi.org/10.21105/joss.01450},
  year = {2019},
  month = {May},
  publisher = {The Open Journal},
  volume = {4},
  number = {37},
  pages = {1450},
  author = {Bane Sullivan and Alexander Kaszynski},
  title = {{PyVista}: {3D} plotting and mesh analysis through a streamlined interface for the {Visualization Toolkit} ({VTK})},
  journal = {Journal of Open Source Software}
}

@article{kagiyama2017prolapse,
  title={Prolapse volume to prolapse height ratio for differentiating Barlow’s disease from fibroelastic deficiency},
  author={Kagiyama, Nobuyuki and Toki, Misako and Hayashida, Akihiro and Ohara, Minako and Hirohata, Atsushi and Yamamoto, Keizo and Totsugawa, Toshinori and Sakaguchi, Taichi and Yoshida, Kiyoshi and Isobe, Mitsuaki},
  journal={Circulation Journal},
  volume={81},
  number={11},
  pages={1730--1735},
  year={2017},
  publisher={The Japanese Circulation Society}
}

@article{tibayan2007tenting,
  title={Tenting volume: three-dimensional assessment of geometric perturbations in functional mitral regurgitation and implications for surgical repair.},
  author={Tibayan, Frederick A and Wilson, Ariane and Lai, DT and Timek, Tomasz A and Dagum, Paul and Rodriguez, Filiberto and Zasio, Mary K and Liang, David and Daughters, George T and Ingels Jr, Neil B and others},
  journal={The Journal of heart valve disease},
  volume={16},
  number={1},
  pages={1--7},
  year={2007}
}

@article{mufarrih2023geometric,
  title={Geometric indices for predicting ischemic mitral regurgitation: Correlation of mitral valve coaptation area with tenting height, tenting area and tenting volume},
  author={Mufarrih, Syed Hamza and Sharkey, Aidan and Mahmood, Feroze and Yunus, Rayaan Ahmed and Qureshi, Nada Qaisar and Senthilnathan, Venkatachalam and Chu, Louis and Liu, David and Khabbaz, Kamal},
  journal={Journal of Cardiothoracic and Vascular Anesthesia},
  volume={37},
  number={1},
  pages={8--15},
  year={2023},
  publisher={Elsevier}
}

@article{ciarka2010predictors,
  title={Predictors of mitral regurgitation recurrence in patients with heart failure undergoing mitral valve annuloplasty},
  author={Ciarka, Agnieszka and Braun, Jerry and Delgado, Victoria and Versteegh, Michel and Boersma, Eric and Klautz, Robert and Dion, Robert and Bax, Jeroen J and Van de Veire, Nico},
  journal={The American journal of cardiology},
  volume={106},
  number={3},
  pages={395--401},
  year={2010},
  publisher={Elsevier}
}

@article{viani2020mitral,
  title={Mitral annulus morphometry in degenerative mitral regurgitation phenotypes},
  author={Viani, Giacomo Maria and Leo, Laura Anna and Borruso, Maria Giuliana and Klersy, Catherine and Paiocchi, Vera Lucia and Schlossbauer, Susanne Anna and Caretta, Alessandro and Demertzis, Stefanos and Faletra, Francesco Fulvio},
  journal={Echocardiography},
  volume={37},
  number={4},
  pages={612--619},
  year={2020},
  publisher={Wiley Online Library}
}

@article{figlioli2025global,
  title={Global prevalence of mitral regurgitation: a systematic review and meta-analysis of population-based studies},
  author={Figlioli, Gisella and Sticchi, Alessandro and Christodoulou, Maria Nefeli and Hadjidemetriou, Andreas and Amorim Moreira Alves, Gabriel and De Carlo, Marco and Praz, Fabien and Caterina, Raffaele De and Nikolopoulos, Georgios K and Bonovas, Stefanos and others},
  journal={Journal of Clinical Medicine},
  volume={14},
  number={8},
  pages={2749},
  year={2025},
  publisher={MDPI}
}

@article{pierard2010ischaemic,
  title={Ischaemic mitral regurgitation: pathophysiology, outcomes and the conundrum of treatment},
  author={Pierard, Luc A and Carabello, Blase A},
  journal={European heart journal},
  volume={31},
  number={24},
  pages={2996--3005},
  year={2010},
  publisher={Oxford University Press}
}

@article{topilsky2019burden,
  title={Burden of tricuspid regurgitation in patients diagnosed in the community setting},
  author={Topilsky, Yan and Maltais, Simon and Medina Inojosa, Jose and Oguz, Didem and Michelena, Hector and Maalouf, Joseph and Mahoney, Douglas W and Enriquez-Sarano, Maurice},
  journal={JACC: Cardiovascular Imaging},
  volume={12},
  number={3},
  pages={433--442},
  year={2019},
  publisher={American College of Cardiology Foundation Washington, DC}
}

@article{zack2017national,
  title={National trends and outcomes in isolated tricuspid valve surgery},
  author={Zack, Chad J and Fender, Erin A and Chandrashekar, Pranav and Reddy, Yogesh NV and Bennett, Courtney E and Stulak, John M and Miller, Virginia M and Nishimura, Rick A},
  journal={Journal of the American College of Cardiology},
  volume={70},
  number={24},
  pages={2953--2960},
  year={2017},
  publisher={American College of Cardiology Foundation Washington, DC}
}

@article{kawsara2021determinants,
  title={Determinants of morbidity and mortality associated with isolated tricuspid valve surgery},
  author={Kawsara, Akram and Alqahtani, Fahad and Nkomo, Vuyisile T and Eleid, Mackram F and Pislaru, Sorin V and Rihal, Charanjit S and Nishimura, Rick A and Schaff, Hartzell V and Crestanello, Juan A and Alkhouli, Mohamad},
  journal={Journal of the American Heart Association},
  volume={10},
  number={2},
  pages={e018417},
  year={2021}
}

@article{habehh2021machine,
  title={Machine learning in healthcare},
  author={Habehh, Hafsa and Gohel, Suril},
  journal={Current genomics},
  volume={22},
  number={4},
  pages={291--300},
  year={2021},
  publisher={Bentham Science Publishers direct}
}

@inproceedings{ahsan2022machine,
  title={Machine-learning-based disease diagnosis: A comprehensive review},
  author={Ahsan, Md Manjurul and Luna, Shahana Akter and Siddique, Zahed},
  booktitle={Healthcare},
  volume={10},
  pages={541},
  year={2022},
  organization={MDPI}
}

@article{marchal2025applications,
  title={Applications of digital twins in medicine},
  author={Marchal, Iris},
  journal={nature biotechnology},
  volume={43},
  number={10},
  pages={1606--1612},
  year={2025},
  publisher={Nature Publishing Group US New York}
}

@article{sel2024building,
  title={Building digital twins for cardiovascular health: from principles to clinical impact},
  author={Sel, Kaan and Osman, Deen and Zare, Fatemeh and Masoumi Shahrbabak, Sina and Brattain, Laura and Hahn, Jin-Oh and Inan, Omer T and Mukkamala, Ramakrishna and Palmer, Jeffrey and Paydarfar, David and others},
  journal={Journal of the American Heart Association},
  volume={13},
  number={19},
  pages={e031981},
  year={2024}
}

@article{sun2020surrogate,
  title={Surrogate modeling for fluid flows based on physics-constrained deep learning without simulation data},
  author={Sun, Luning and Gao, Han and Pan, Shaowu and Wang, Jian-Xun},
  journal={Computer Methods in Applied Mechanics and Engineering},
  volume={361},
  pages={112732},
  year={2020},
  publisher={Elsevier}
}

@article{du2022deep,
  title={Deep learning-based surrogate model for three-dimensional patient-specific computational fluid dynamics},
  author={Du, Pan and Zhu, Xiaozhi and Wang, Jian-Xun},
  journal={Physics of Fluids},
  volume={34},
  number={8},
  year={2022},
  publisher={AIP Publishing}
}

@article{motiwale2024neural,
  title={A neural network finite element approach for high speed cardiac mechanics simulations},
  author={Motiwale, Shruti and Zhang, Wenbo and Feldmeier, Reese and Sacks, Michael S},
  journal={Computer Methods in Applied Mechanics and Engineering},
  volume={427},
  pages={117060},
  year={2024},
  publisher={Elsevier}
}

@article{liang2018deep,
  title={A deep learning approach to estimate stress distribution: a fast and accurate surrogate of finite-element analysis},
  author={Liang, Liang and Liu, Minliang and Martin, Caitlin and Sun, Wei},
  journal={Journal of The Royal Society Interface},
  volume={15},
  number={138},
  year={2018},
  publisher={The Royal Society}
}

@article{balu2019deep,
  title={A deep learning framework for design and analysis of surgical bioprosthetic heart valves},
  author={Balu, Aditya and Nallagonda, Sahiti and Xu, Fei and Krishnamurthy, Adarsh and Hsu, Ming-Chen and Sarkar, Soumik},
  journal={Scientific reports},
  volume={9},
  number={1},
  pages={18560},
  year={2019},
  publisher={Nature Publishing Group UK London}
}

@article{dosovitskiy2020image,
  title={An image is worth 16x16 words: Transformers for image recognition at scale},
  author={Dosovitskiy, Alexey and Beyer, Lucas and Kolesnikov, Alexander and Weissenborn, Dirk and Zhai, Xiaohua and Unterthiner, Thomas and Dehghani, Mostafa and Minderer, Matthias and Heigold, Georg and Gelly, Sylvain and others},
  journal={arXiv preprint arXiv:2010.11929},
  year={2020}
}

@inproceedings{liu2021swin,
  title={Swin transformer: Hierarchical vision transformer using shifted windows},
  author={Liu, Ze and Lin, Yutong and Cao, Yue and Hu, Han and Wei, Yixuan and Zhang, Zheng and Lin, Stephen and Guo, Baining},
  booktitle={Proceedings of the IEEE/CVF international conference on computer vision},
  pages={10012--10022},
  year={2021}
}

@inproceedings{arnab2021vivit,
  title={Vivit: A video vision transformer},
  author={Arnab, Anurag and Dehghani, Mostafa and Heigold, Georg and Sun, Chen and Lu{\v{c}}i{\'c}, Mario and Schmid, Cordelia},
  booktitle={Proceedings of the IEEE/CVF international conference on computer vision},
  pages={6836--6846},
  year={2021}
}

@article{baevski2020wav2vec,
  title={wav2vec 2.0: A framework for self-supervised learning of speech representations},
  author={Baevski, Alexei and Zhou, Yuhao and Mohamed, Abdelrahman and Auli, Michael},
  journal={Advances in neural information processing systems},
  volume={33},
  pages={12449--12460},
  year={2020}
}

@article{bodnar2024aurora,
  title={Aurora: A foundation model of the atmosphere},
  author={Bodnar, Cristian and Bruinsma, Wessel P and Lucic, Ana and Stanley, Megan and Brandstetter, Johannes and Garvan, Patrick and Riechert, Maik and Weyn, Jonathan and Dong, Haiyu and Vaughan, Anna and others},
  journal={arXiv preprint arXiv:2405.13063},
  volume={1},
  number={8},
  year={2024}
}

@article{kovachki2023neural,
  title={Neural operator: Learning maps between function spaces with applications to pdes},
  author={Kovachki, Nikola and Li, Zongyi and Liu, Burigede and Azizzadenesheli, Kamyar and Bhattacharya, Kaushik and Stuart, Andrew and Anandkumar, Anima},
  journal={Journal of Machine Learning Research},
  volume={24},
  number={89},
  pages={1--97},
  year={2023}
}

@article{cao2021choose,
  title={Choose a transformer: Fourier or galerkin},
  author={Cao, Shuhao},
  journal={Advances in neural information processing systems},
  volume={34},
  pages={24924--24940},
  year={2021}
}

@article{wang2024cvit,
  title={Cvit: Continuous vision transformer for operator learning},
  author={Wang, Sifan and Seidman, Jacob H and Sankaran, Shyam and Wang, Hanwen and Pappas, George J and Perdikaris, Paris},
  journal={arXiv preprint arXiv:2405.13998},
  year={2024}
}

@inproceedings{jaegle2021perceiver,
  title={Perceiver: General perception with iterative attention},
  author={Jaegle, Andrew and Gimeno, Felix and Brock, Andy and Vinyals, Oriol and Zisserman, Andrew and Carreira, Joao},
  booktitle={International conference on machine learning},
  pages={4651--4664},
  year={2021},
  organization={PMLR}
}

@article{jaegle2021perceiver2,
  title={Perceiver io: A general architecture for structured inputs \& outputs},
  author={Jaegle, Andrew and Borgeaud, Sebastian and Alayrac, Jean-Baptiste and Doersch, Carl and Ionescu, Catalin and Ding, David and Koppula, Skanda and Zoran, Daniel and Brock, Andrew and Shelhamer, Evan and others},
  journal={arXiv preprint arXiv:2107.14795},
  year={2021}
}

@article{wagner2014subvalvular,
  title={Subvalvular techniques to optimize surgical repair of ischemic mitral regurgitation},
  author={Wagner, Cynthia E and Kron, Irving L},
  journal={Current opinion in cardiology},
  volume={29},
  number={2},
  pages={140--144},
  year={2014},
  publisher={LWW}
}

@article{delling2014epidemiology,
  title={Epidemiology and pathophysiology of mitral valve prolapse: new insights into disease progression, genetics, and molecular basis},
  author={Delling, Francesca N and Vasan, Ramachandran S},
  journal={Circulation},
  volume={129},
  number={21},
  pages={2158--2170},
  year={2014},
  publisher={Lippincott Williams \& Wilkins Hagerstown, MD}
}

@misc{jax2018github,
  author = {James Bradbury and Roy Frostig and Peter Hawkins and Matthew James Johnson and Yash Katariya and Chris Leary and Dougal Maclaurin and George Necula and Adam Paszke and Jake Vander{P}las and Skye Wanderman-{M}ilne and Qiao Zhang},
  title = {{JAX}: composable transformations of {P}ython+{N}um{P}y programs},
  url = {http://github.com/jax-ml/jax},
  version = {0.3.13},
  year = {2018},
}

@misc{deepmind2020jax,
  title = {The {D}eep{M}ind {JAX} {E}cosystem},
  author = {DeepMind and Babuschkin, Igor and Baumli, Kate and Bell, Alison and Bhupatiraju, Surya and Bruce, Jake and Buchlovsky, Peter and Budden, David and Cai, Trevor and Clark, Aidan and Danihelka, Ivo and Dedieu, Antoine and Fantacci, Claudio and Godwin, Jonathan and Jones, Chris and Hemsley, Ross and Hennigan, Tom and Hessel, Matteo and Hou, Shaobo and Kapturowski, Steven and Keck, Thomas and Kemaev, Iurii and King, Michael and Kunesch, Markus and Martens, Lena and Merzic, Hamza and Mikulik, Vladimir and Norman, Tamara and Papamakarios, George and Quan, John and Ring, Roman and Ruiz, Francisco and Sanchez, Alvaro and Sartran, Laurent and Schneider, Rosalia and Sezener, Eren and Spencer, Stephen and Srinivasan, Srivatsan and Stanojevi\'{c}, Milo\v{s} and Stokowiec, Wojciech and Wang, Luyu and Zhou, Guangyao and Viola, Fabio},
  url = {http://github.com/google-deepmind},
  year = {2020},
}

@misc{flax2020github,
  author = {Jonathan Heek and Anselm Levskaya and Avital Oliver and Marvin Ritter and Bertrand Rondepierre and Andreas Steiner and Marc van {Z}ee},
  title = {{F}lax: A neural network library and ecosystem for {JAX}},
  url = {http://github.com/google/flax},
  version = {0.12.7},
  year = {2024},
}

@article{lee2014inverse,
  title={An inverse modeling approach for stress estimation in mitral valve anterior leaflet valvuloplasty for in-vivo valvular biomaterial assessment},
  author={Lee, Chung-Hao and Amini, Rouzbeh and Gorman, Robert C and Gorman III, Joseph H and Sacks, Michael S},
  journal={Journal of biomechanics},
  volume={47},
  number={9},
  pages={2055--2063},
  year={2014},
  publisher={Elsevier}
}

@article{wu2022computational,
  title={A computational framework for atrioventricular valve modeling using open-source software},
  author={Wu, Wensi and Ching, Stephen and Maas, Steve A and Lasso, Andras and Sabin, Patricia and Weiss, Jeffrey A and Jolley, Matthew A},
  journal={Journal of Biomechanical Engineering},
  volume={144},
  number={10},
  pages={101012},
  year={2022},
  publisher={American Society of Mechanical Engineers}
}

@article{wu2023effects,
  title={The effects of leaflet material properties on the simulated function of regurgitant mitral valves},
  author={Wu, Wensi and Ching, Stephen and Sabin, Patricia and Laurence, Devin W and Maas, Steve A and Lasso, Andras and Weiss, Jeffrey A and Jolley, Matthew A},
  journal={Journal of the mechanical behavior of biomedical materials},
  volume={142},
  pages={105858},
  year={2023},
  publisher={Elsevier}
}

@article{wu2018anisotropic,
  title={An anisotropic constitutive model for immersogeometric fluid--structure interaction analysis of bioprosthetic heart valves},
  author={Wu, Michael CH and Zakerzadeh, Rana and Kamensky, David and Kiendl, Josef and Sacks, Michael S and Hsu, Ming-Chen},
  journal={Journal of biomechanics},
  volume={74},
  pages={23--31},
  year={2018},
  publisher={Elsevier}
}

@inproceedings{tay2023scaling,
  title={Scaling laws vs model architectures: How does inductive bias influence scaling?},
  author={Tay, Yi and Dehghani, Mostafa and Abnar, Samira and Chung, Hyung and Fedus, William and Rao, Jinfeng and Narang, Sharan and Tran, Vinh and Yogatama, Dani and Metzler, Donald},
  booktitle={Findings of the Association for Computational Linguistics: EMNLP 2023},
  pages={12342--12364},
  year={2023}
}

@article{ebrahimi2026induction,
  title={On the" Induction Bias" in Sequence Models},
  author={Ebrahimi, M Reza and Defferrard, Micha{\"e}l and Panchal, Sunny and Memisevic, Roland},
  journal={arXiv preprint arXiv:2602.18333},
  year={2026}
}

@article{wang2022and,
  title={When and why PINNs fail to train: A neural tangent kernel perspective},
  author={Wang, Sifan and Yu, Xinling and Perdikaris, Paris},
  journal={Journal of Computational Physics},
  volume={449},
  pages={110768},
  year={2022},
  publisher={Elsevier}
}

@article{wang2021understanding,
  title={Understanding and mitigating gradient flow pathologies in physics-informed neural networks},
  author={Wang, Sifan and Teng, Yujun and Perdikaris, Paris},
  journal={SIAM Journal on Scientific Computing},
  volume={43},
  number={5},
  pages={A3055--A3081},
  year={2021},
  publisher={SIAM}
}

@article{wang2026pinns,
  title={When PINNs Go Wrong: Pseudo-Time Stepping Against Spurious Solutions},
  author={Wang, Sifan and Koohy, Shawn and Lu, Yiping and Perdikaris, Paris},
  journal={arXiv preprint arXiv:2604.23528},
  year={2026}
}

@article{wang2024respecting,
  title={Respecting causality for training physics-informed neural networks},
  author={Wang, Sifan and Sankaran, Shyam and Perdikaris, Paris},
  journal={Computer Methods in Applied Mechanics and Engineering},
  volume={421},
  pages={116813},
  year={2024},
  publisher={Elsevier}
}

@article{mccabe2023multiple,
  title={Multiple physics pretraining for physical surrogate models},
  author={McCabe, Michael and Blancard, Bruno R{\'e}galdo-Saint and Parker, Liam Holden and Ohana, Ruben and Cranmer, Miles and Bietti, Alberto and Eickenberg, Michael and Golkar, Siavash and Krawezik, Geraud and Lanusse, Francois and others},
  journal={arXiv preprint arXiv:2310.02994},
  year={2023}
}

@article{raissi2019physics,
  title={Physics-informed neural networks: A deep learning framework for solving forward and inverse problems involving nonlinear partial differential equations},
  author={Raissi, Maziar and Perdikaris, Paris and Karniadakis, George E},
  journal={Journal of Computational physics},
  volume={378},
  pages={686--707},
  year={2019},
  publisher={Elsevier}
}

@article{psaros2023uncertainty,
  title={Uncertainty quantification in scientific machine learning: Methods, metrics, and comparisons},
  author={Psaros, Apostolos F and Meng, Xuhui and Zou, Zongren and Guo, Ling and Karniadakis, George Em},
  journal={Journal of Computational Physics},
  volume={477},
  pages={111902},
  year={2023},
  publisher={Elsevier}
}

@article{lopez2025uncertainty,
  title={Uncertainty quantification for machine learning in healthcare: a survey},
  author={L{\'o}pez, L and Elsharief, Shaza and Jorf, Dhiyaa Al and Darwish, Firas and Ma, Congbo and Shamout, Farah E},
  journal={arXiv preprint arXiv:2505.02874},
  year={2025}
}

@article{ranftl2022stochastic,
  title={Stochastic modeling of inhomogeneities in the aortic wall and uncertainty quantification using a Bayesian encoder--decoder surrogate},
  author={Ranftl, Sascha and Rolf-Pissarczyk, Malte and Wolkerstorfer, Gloria and Pepe, Antonio and Egger, Jan and von der Linden, Wolfgang and Holzapfel, Gerhard A},
  journal={Computer Methods in Applied Mechanics and Engineering},
  volume={401},
  pages={115594},
  year={2022},
  publisher={Elsevier}
}

@article{blanke2014simplified,
  title={A simplified D-shaped model of the mitral annulus to facilitate CT-based sizing before transcatheter mitral valve implantation},
  author={Blanke, Philipp and Dvir, Danny and Cheung, Anson and Ye, Jian and Levine, Robert A and Precious, Bruce and Berger, Adam and Stub, Dion and Hague, Cameron and Murphy, Darra and others},
  journal={Journal of cardiovascular computed tomography},
  volume={8},
  number={6},
  pages={459--467},
  year={2014},
  publisher={Elsevier}
}

@article{salgo2002effect,
  title={Effect of annular shape on leaflet curvature in reducing mitral leaflet stress},
  author={Salgo, Ivan S and Gorman III, Joseph H and Gorman, Robert C and Jackson, Benjamin M and Bowen, Frank W and Plappert, Theodore and St John Sutton, Martin G and Edmunds Jr, L Henry},
  journal={Circulation},
  volume={106},
  number={6},
  pages={711--717},
  year={2002},
  publisher={Lippincott Williams \& Wilkins}
}

@article{lasso2022slicerheart,
  title={SlicerHeart: An open-source computing platform for cardiac image analysis and modeling},
  author={Lasso, Andras and Herz, Christian and Nam, Hannah and Cianciulli, Alana and Pieper, Steve and Drouin, Simon and Pinter, Csaba and St-Onge, Samuelle and Vigil, Chad and Ching, Stephen and others},
  journal={Frontiers in cardiovascular medicine},
  volume={9},
  pages={886549},
  year={2022},
  publisher={Frontiers Media SA}
}

@article{nguyen2019dynamic,
  title={Dynamic three-dimensional geometry of the tricuspid valve annulus in hypoplastic left heart syndrome with a Fontan circulation},
  author={Nguyen, Alex V and Lasso, Andras and Nam, Hannah H and Faerber, Jennifer and Aly, Ahmed H and Pouch, Alison M and Scanlan, Adam B and McGowan, Francis X and Mercer-Rosa, Laura and Cohen, Meryl S and others},
  journal={Journal of the American Society of Echocardiography},
  volume={32},
  number={5},
  pages={655--666},
  year={2019},
  publisher={Elsevier}
}

@article{nam2022dynamic,
  title={Dynamic annular modeling of the unrepaired complete atrioventricular canal annulus},
  author={Nam, Hannah H and Dinh, Patrick V and Lasso, Andras and Herz, Christian and Huang, Jing and Posada, Adriana and Aly, Ahmed H and Pouch, Alison M and Kabir, Saleha and Simpson, John and others},
  journal={The Annals of thoracic surgery},
  volume={113},
  number={2},
  pages={654--662},
  year={2022},
  publisher={Elsevier}
}

@article{hibino2024valvular,
  title={Valvular heart disease-related mortality between middle-and high-income countries during 2000 to 2019},
  author={Hibino, Makoto and Ueyama, Hiroki A and Halkos, Michael E and Grubb, Kendra J and Verma, Raj and Majeed, Azeem and Nienaber, Christoph A and Yanagawa, Bobby and Bhatt, Deepak L and Verma, Subodh},
  journal={JACC: Advances},
  volume={3},
  number={12\_Part\_2},
  pages={101133},
  year={2024},
  publisher={American College of Cardiology Foundation Washington DC}
}

@article{roth2020global,
  title={Global burden of cardiovascular diseases and risk factors, 1990--2019: update from the GBD 2019 study},
  author={Roth, Gregory A and Mensah, George A and Johnson, Catherine O and Addolorato, Giovanni and Ammirati, Enrico and Baddour, Larry M and Barengo, No{\"e}l C and Beaton, Andrea Z and Benjamin, Emelia J and Benziger, Catherine P and others},
  journal={Journal of the American college of cardiology},
  volume={76},
  number={25},
  pages={2982--3021},
  year={2020},
  publisher={American College of Cardiology Foundation Washington DC}
}

@article{dziadzko2018outcome,
  title={Outcome and undertreatment of mitral regurgitation: a community cohort study},
  author={Dziadzko, Volha and Clavel, Marie-Annick and Dziadzko, Mikhail and Medina-Inojosa, Jose R and Michelena, Hector and Maalouf, Joseph and Nkomo, Vuyisile and Thapa, Prabin and Enriquez-Sarano, Maurice},
  journal={The Lancet},
  volume={391},
  number={10124},
  pages={960--969},
  year={2018},
  publisher={Elsevier}
}

@article{pingitore20223d,
  title={3D mitral annulus echocardiography assessment in patients affected by degenerative mitral regurgitation who underwent mitral valve repair with flexible band},
  author={Pingitore, Annachiara and Polizzi, Vincenzo and Cardillo, Ilaria and Lio, Antonio and Ranocchi, Federico and Pergolini, Amedeo and Musumeci, Francesco},
  journal={Journal of Cardiac Surgery},
  volume={37},
  number={12},
  pages={4269--4277},
  year={2022},
  publisher={Wiley Online Library}
}

@article{muraru2021right,
  title={Right atrial volume is a major determinant of tricuspid annulus area in functional tricuspid regurgitation: a three-dimensional echocardiographic study},
  author={Muraru, Denisa and Addetia, Karima and Guta, Andrada C and Ochoa-Jimenez, Roberto C and Genovese, Davide and Veronesi, Federico and Basso, Cristina and Iliceto, Sabino and Badano, Luigi P and Lang, Roberto M},
  journal={European Heart Journal-Cardiovascular Imaging},
  volume={22},
  number={6},
  pages={660--669},
  year={2021},
  publisher={Oxford University Press}
}

@article{oguz2019quantitative,
  title={Quantitative three-dimensional echocardiographic correlates of optimal mitral regurgitation reduction during transcatheter mitral valve repair},
  author={Oguz, Didem and Eleid, Mackram F and Dhesi, Sumandeep and Pislaru, Sorin V and Mankad, Sunil V and Malouf, Joseph F and Nkomo, Vuyisile T and Oh, Jae K and Holmes, David R and Reeder, Guy S and others},
  journal={Journal of the American Society of Echocardiography},
  volume={32},
  number={11},
  pages={1426--1435},
  year={2019},
  publisher={Elsevier}
}

@article{sacks2019simulation,
  title={On the simulation of mitral valve function in health, disease, and treatment},
  author={Sacks, Michael S and Drach, Andrew and Lee, Chung-Hao and Khalighi, Amir H and Rego, Bruno V and Zhang, Will and Ayoub, Salma and Yoganathan, Ajit P and Gorman, Robert C and Gorman III, Joseph H},
  journal={Journal of biomechanical engineering},
  volume={141},
  number={7},
  pages={070804},
  year={2019},
  publisher={American Society of Mechanical Engineers}
}

@article{kong2020finite,
  title={Finite element analysis of MitraClip procedure on a patient-specific model with functional mitral regurgitation},
  author={Kong, Fanwei and Caballero, Andr{\'e}s and McKay, Raymond and Sun, Wei},
  journal={Journal of biomechanics},
  volume={104},
  pages={109730},
  year={2020},
  publisher={Elsevier}
}

@article{biffi2019workflow,
  title={A workflow for patient-specific fluid--structure interaction analysis of the mitral valve: A proof of concept on a mitral regurgitation case},
  author={Biffi, Benedetta and Gritti, Maurizio and Grasso, Agata and Milano, Elena G and Fontana, Marianna and Alkareef, Hamad and Davar, Joseph and Jeetley, Paramijit and Whelan, Carol and Anderson, Sarah and others},
  journal={Medical Engineering \& Physics},
  volume={74},
  number={1},
  pages={153--161},
  year={2019},
  publisher={IOP Publishing}
}

@article{choi2014novel,
  title={A novel finite element-based patient-specific mitral valve repair: virtual ring annuloplasty},
  author={Choi, Ahnryul and Rim, Yonghoon and Mun, Jeffrey S and Kim, Hyunggun},
  journal={Bio-medical materials and engineering},
  volume={24},
  number={1},
  pages={341--347},
  year={2014},
  publisher={SAGE Publications Sage UK: London, England}
}

@article{choi2017neochordoplasty,
  title={Neochordoplasty versus leaflet resection for ruptured mitral chordae treatment: Virtual mitral valve repair},
  author={Choi, Ahnryul and McPherson, David D and Kim, Hyunggun},
  journal={Computers in Biology and Medicine},
  volume={90},
  pages={50--58},
  year={2017},
  publisher={Elsevier}
}

@article{nkomo2006burden,
  title={Burden of valvular heart diseases: a population-based study},
  author={Nkomo, Vuyisile T and Gardin, Julius M and Skelton, Thomas N and Gottdiener, John S and Scott, Christopher G and Enriquez-Sarano, Maurice},
  journal={The lancet},
  volume={368},
  number={9540},
  pages={1005--1011},
  year={2006},
  publisher={Elsevier}
}

@article{aslan2025simulation,
  title={Simulation of Transcatheter Therapies for Atrioventricular Valve Regurgitation in an Open-Source Finite Element Simulation Framework},
  author={Aslan, Seda and Mangine, Nicolas R and Laurence, Devin W and Sabin, Patricia M and Wu, Wensi and Herz, Christian and Unger, Justin S and Maas, Steve A and Gillespie, Matthew J and Weiss, Jeffrey A and others},
  journal={arXiv preprint arXiv:2509.22865},
  year={2025}
}

@article{vergnat2012influence,
  title={The influence of saddle-shaped annuloplasty on leaflet curvature in patients with ischaemic mitral regurgitation},
  author={Vergnat, Mathieu and Levack, Melissa M and Jassar, Arminder S and Jackson, Benjamin M and Acker, Michael A and Woo, Y Joseph and Gorman, Robert C and Gorman III, Joseph H},
  journal={European journal of cardio-thoracic surgery},
  volume={42},
  number={3},
  pages={493--499},
  year={2012},
  publisher={Oxford University Press}
}

@article{herz2021segmentation,
  title={Segmentation of tricuspid valve leaflets from transthoracic 3D echocardiograms of children with hypoplastic left heart syndrome using deep learning},
  author={Herz, Christian and Pace, Danielle F and Nam, Hannah H and Lasso, Andras and Dinh, Patrick and Flynn, Maura and Cianciulli, Alana and Golland, Polina and Jolley, Matthew A},
  journal={Frontiers in Cardiovascular Medicine},
  volume={8},
  pages={735587},
  year={2021},
  publisher={Frontiers Media SA}
}

@article{jassar2014regional,
  title={Regional annular geometry in patients with mitral regurgitation: implications for annuloplasty ring selection},
  author={Jassar, Arminder S and Vergnat, Mathieu and Jackson, Benjamin M and McGarvey, Jeremy R and Cheung, Albert T and Ferrari, Giovanni and Woo, Y Joseph and Acker, Michael A and Gorman, Robert C and Gorman III, Joseph H},
  journal={The Annals of thoracic surgery},
  volume={97},
  number={1},
  pages={64--70},
  year={2014},
  publisher={Elsevier}
}

@article{laurence2024febio,
  title={FEBio FINESSE: An open-source finite element simulation approach to estimate in vivo heart valve strains using shape enforcement},
  author={Laurence, Devin W and Sabin, Patricia M and Sulentic, Analise M and Daemer, Matthew and Maas, Steve A and Weiss, Jeffrey A and Jolley, Matthew A},
  journal={Annals of biomedical engineering},
  volume={53},
  number={1},
  pages={241},
  year={2024}
}

@article{el2021valve,
  title={Valve strain quantitation in normal mitral valves and mitral prolapse with variable degrees of regurgitation},
  author={El-Tallawi, K Carlos and Zhang, Peng and Azencott, Robert and He, Jiwen and Herrera, Elizabeth L and Xu, Jiaqiong and Chamsi-Pasha, Mohammed and Jacob, Jessen and Lawrie, Gerald M and Zoghbi, William A},
  journal={Cardiovascular Imaging},
  volume={14},
  number={6},
  pages={1099--1109},
  year={2021},
  publisher={American College of Cardiology Foundation Washington DC}
}

@article{el2021mitral,
  title={Mitral valve remodeling and strain in secondary mitral regurgitation: comparison with primary regurgitation and normal valves},
  author={El-Tallawi, K Carlos and Zhang, Peng and Azencott, Robert and He, Jiwen and Xu, Jiaqiong and Herrera, Elizabeth L and Jacob, Jessen and Chamsi-Pasha, Mohammed and Lawrie, Gerald M and Zoghbi, William A},
  journal={Cardiovascular Imaging},
  volume={14},
  number={4},
  pages={782--793},
  year={2021},
  publisher={American College of Cardiology Foundation Washington DC}
}

@article{maas2017febio,
  title={FEBio: history and advances},
  author={Maas, Steve A and Ateshian, Gerard A and Weiss, Jeffrey A},
  journal={Annual review of biomedical engineering},
  volume={19},
  pages={279--299},
  year={2017},
  publisher={Annual Reviews}
}

@Article{         harris2020array,
 title         = {Array programming with {NumPy}},
 author        = {Charles R. Harris and K. Jarrod Millman and St{\'{e}}fan J.
                 van der Walt and Ralf Gommers and Pauli Virtanen and David
                 Cournapeau and Eric Wieser and Julian Taylor and Sebastian
                 Berg and Nathaniel J. Smith and Robert Kern and Matti Picus
                 and Stephan Hoyer and Marten H. van Kerkwijk and Matthew
                 Brett and Allan Haldane and Jaime Fern{\'{a}}ndez del
                 R{\'{i}}o and Mark Wiebe and Pearu Peterson and Pierre
                 G{\'{e}}rard-Marchant and Kevin Sheppard and Tyler Reddy and
                 Warren Weckesser and Hameer Abbasi and Christoph Gohlke and
                 Travis E. Oliphant},
 year          = {2020},
 month         = sep,
 journal       = {Nature},
 volume        = {585},
 number        = {7825},
 pages         = {357--362},
 doi           = {10.1038/s41586-020-2649-2},
 publisher     = {Springer Science and Business Media {LLC}},
 url           = {https://doi.org/10.1038/s41586-020-2649-2}
}

@ARTICLE{2020SciPy-NMeth,
  author  = {Virtanen, Pauli and Gommers, Ralf and Oliphant, Travis E. and
            Haberland, Matt and Reddy, Tyler and Cournapeau, David and
            Burovski, Evgeni and Peterson, Pearu and Weckesser, Warren and
            Bright, Jonathan and {van der Walt}, St{\'e}fan J. and
            Brett, Matthew and Wilson, Joshua and Millman, K. Jarrod and
            Mayorov, Nikolay and Nelson, Andrew R. J. and Jones, Eric and
            Kern, Robert and Larson, Eric and Carey, C J and
            Polat, {\.I}lhan and Feng, Yu and Moore, Eric W. and
            {VanderPlas}, Jake and Laxalde, Denis and Perktold, Josef and
            Cimrman, Robert and Henriksen, Ian and Quintero, E. A. and
            Harris, Charles R. and Archibald, Anne M. and
            Ribeiro, Ant{\^o}nio H. and Pedregosa, Fabian and
            {van Mulbregt}, Paul and {SciPy 1.0 Contributors}},
  title   = {{{SciPy} 1.0: Fundamental Algorithms for Scientific
            Computing in Python}},
  journal = {Nature Methods},
  year    = {2020},
  volume  = {17},
  pages   = {261--272},
  adsurl  = {https://rdcu.be/b08Wh},
  doi     = {10.1038/s41592-019-0686-2},
}

@Article{Hunter:2007,
  Author    = {Hunter, J. D.},
  Title     = {Matplotlib: A 2D graphics environment},
  Journal   = {Computing in Science \& Engineering},
  Volume    = {9},
  Number    = {3},
  Pages     = {90--95},
  publisher = {IEEE COMPUTER SOC},
  doi       = {10.1109/MCSE.2007.55},
  year      = 2007
}

@Book{vtkBook,
  author    = "Will Schroeder and Ken Martin and Bill Lorensen",
  title     = "The Visualization Toolkit (4th ed.)",
  publisher = "Kitware",
  year      = "2006",
  isbn      = "978-1-930934-19-1",
}

@inproceedings{
    rogozhnikov2022einops,
    title={Einops: Clear and Reliable Tensor Manipulations with Einstein-like Notation},
    author={Alex Rogozhnikov},
    booktitle={International Conference on Learning Representations},
    year={2022},
    url={https://openreview.net/forum?id=oapKSVM2bcj}
}

@misc{wandb,
title = {Experiment Tracking with Weights and Biases},
year = {2020},
note = {Software available from wandb.com},
url={https://www.wandb.com/},
author = {Biewald, Lukas},
}

@misc{ml_collections,
  author  = {G{\'o}mez Colmenarejo, Sergio and Czarnecki, Wojciech Marian and Watters, Nicholas and Reddy, Mohit},
  title   = {{ML Collections}},
  url     = {https://github.com/google/ml_collections},
  version = {1.1.0},
  year    = {2025},
  organization = {Google}
}

@article{katsi2019role,
  title={The role of arterial hypertension in mitral valve regurgitation},
  author={Katsi, Vasiliki and Georgiopoulos, Georgios and Magkas, Nikolaos and Oikonomou, Dimitrios and Virdis, Agostino and Nihoyannopoulos, Petros and Toutouzas, Konstantinos and Tousoulis, Dimitrios},
  journal={Current Hypertension Reports},
  volume={21},
  number={3},
  pages={20},
  year={2019},
  publisher={Springer}
}

@article{rahimi2017elevated,
  title={Elevated blood pressure and risk of mitral regurgitation: A longitudinal cohort study of 5.5 million United Kingdom adults},
  author={Rahimi, Kazem and Mohseni, Hamid and Otto, Catherine M and Conrad, Nathalie and Tran, Jenny and Nazarzadeh, Milad and Woodward, Mark and Dwyer, Terence and MacMahon, Stephen},
  journal={PLoS medicine},
  volume={14},
  number={10},
  pages={e1002404},
  year={2017},
  publisher={Public Library of Science San Francisco, CA USA}
}

@article{li2024physics,
  title={Physics-informed neural operator for learning partial differential equations},
  author={Li, Zongyi and Zheng, Hongkai and Kovachki, Nikola and Jin, David and Chen, Haoxuan and Liu, Burigede and Azizzadenesheli, Kamyar and Anandkumar, Anima},
  journal={ACM/IMS Journal of Data Science},
  volume={1},
  number={3},
  pages={1--27},
  year={2024},
  publisher={ACM New York, NY}
}

@article{ross2024bayesian,
  title={Bayesian Optimization-Based Inverse Finite Element Analysis for Atrioventricular Heart Valves},
  author={Ross, Colton J and Laurence, Devin W and Aggarwal, Ankush and Hsu, Ming-Chen and Mir, Arshid and Burkhart, Harold M and Lee, Chung-Hao},
  journal={Annals of biomedical engineering},
  volume={52},
  number={3},
  pages={611--626},
  year={2024},
  publisher={Springer}
}

@article{herde2024poseidon,
  title={Poseidon: Efficient foundation models for pdes},
  author={Herde, Maximilian and Raoni{\'c}, Bogdan and Rohner, Tobias and K{\"a}ppeli, Roger and Molinaro, Roberto and De Bezenac, Emmanuel and Mishra, Siddhartha},
  journal={Advances in Neural Information Processing Systems},
  volume={37},
  pages={72525--72624},
  year={2024}
}

@article{rahman2024pretraining,
  title={Pretraining codomain attention neural operators for solving multiphysics pdes},
  author={Rahman, Ashiqur and George, Robert J and Elleithy, Mogab and Leibovici, Daniel and Li, Zongyi and Bonev, Boris and White, Colin and Berner, Julius and Yeh, Raymond A and Kossaifi, Jean and others},
  journal={Advances in Neural Information Processing Systems},
  volume={37},
  pages={104035--104064},
  year={2024}
}

@article{rim2013effect,
  title={The effect of patient-specific annular motion on dynamic simulation of mitral valve function},
  author={Rim, Yonghoon and McPherson, David D and Chandran, Krishnan B and Kim, Hyunggun},
  journal={Journal of biomechanics},
  volume={46},
  number={6},
  pages={1104--1112},
  year={2013},
  publisher={Elsevier}
}

@article{kong2018finite,
  title={Finite element analysis of tricuspid valve deformation from multi-slice computed tomography images},
  author={Kong, Fanwei and Pham, Thuy and Martin, Caitlin and McKay, Raymond and Primiano, Charles and Hashim, Sabet and Kodali, Susheel and Sun, Wei},
  journal={Annals of biomedical engineering},
  volume={46},
  number={8},
  pages={1112--1127},
  year={2018},
  publisher={Springer}
}

@article{khalighi2019development,
  title={Development of a Functionally Equivalent Model of the Mitral Valve Chordae Tendineae Through Topology Optimization},
  author={Khalighi, Amir H and Rego, Bruno V and Drach, Andrew and Gorman, Robert C and Gorman III, Joseph H and Sacks, Michael S},
  journal={Annals of biomedical engineering},
  volume={47},
  number={1},
  pages={60--74},
  year={2019},
  publisher={Springer}
}

@article{wu2025noninvasive,
  title={A noninvasive method for determining elastic parameters of valve tissue using physics-informed neural networks},
  author={Wu, Wensi and Daneker, Mitchell and Herz, Christian and Dewey, Hannah and Weiss, Jeffrey A and Pouch, Alison M and Lu, Lu and Jolley, Matthew A},
  journal={Acta biomaterialia},
  volume={200},
  pages={283--298},
  year={2025},
  publisher={Elsevier}
}

@article{pham2017finite,
  title={Finite element analysis of patient-specific mitral valve with mitral regurgitation},
  author={Pham, Thuy and Kong, Fanwei and Martin, Caitlin and Wang, Qian and Primiano, Charles and McKay, Raymond and Elefteriades, John and Sun, Wei},
  journal={Cardiovascular engineering and technology},
  volume={8},
  number={1},
  pages={3--16},
  year={2017},
  publisher={Springer}
}

@article{munafo2024deep,
  title={A deep learning-based fully automated pipeline for regurgitant mitral valve anatomy analysis from 3D echocardiography},
  author={Munafo, Riccardo and Saitta, Simone and Ingallina, Giacomo and Denti, Paolo and Maisano, Francesco and Agricola, Eustachio and Redaelli, Alberto and Votta, Emiliano},
  journal={IEEE Access},
  volume={12},
  pages={5295--5308},
  year={2024},
  publisher={IEEE}
}

@inproceedings{sung2025blendednet,
  title={Blendednet: A blended wing body aircraft dataset and surrogate model for aerodynamic predictions},
  author={Sung, Nicholas and Spreizer, Steven and Elrefaie, Mohamed and Samuel, Kaira and Jones, Matthew C and Ahmed, Faez},
  booktitle={International Design Engineering Technical Conferences and Computers and Information in Engineering Conference},
  volume={89237},
  pages={V03BT03A049},
  year={2025},
  organization={American Society of Mechanical Engineers}
}

@article{lutheran2026diffusion,
  title={Diffusion Transformers with Hybrid Conditioning for Structural Optimization},
  author={Lutheran, Aaron and Das, Srijan and Tabarraei, Alireza},
  journal={arXiv preprint arXiv:2605.02158},
  year={2026}
}

@article{saxena2023zero,
  title={Zero-shot metric depth with a field-of-view conditioned diffusion model},
  author={Saxena, Saurabh and Hur, Junhwa and Herrmann, Charles and Sun, Deqing and Fleet, David J},
  journal={arXiv preprint arXiv:2312.13252},
  year={2023}
}

@article{rivlin1997stress,
  title={Stress-deformation relations for isotropic materials},
  author={Rivlin, Ronald Samuel and Ericksen, Jerald LaVerne},
  journal={Collected Papers of RS Rivlin: Volume I and II},
  pages={911--1013},
  year={1997},
  publisher={Springer}
}

@inproceedings{gilmer2017neural,
  title={Neural message passing for quantum chemistry},
  author={Gilmer, Justin and Schoenholz, Samuel S and Riley, Patrick F and Vinyals, Oriol and Dahl, George E},
  booktitle={International conference on machine learning},
  pages={1263--1272},
  year={2017},
  organization={Pmlr}
}

@article{battaglia2018relational,
  title={Relational inductive biases, deep learning, and graph networks},
  author={Battaglia, Peter W and Hamrick, Jessica B and Bapst, Victor and Sanchez-Gonzalez, Alvaro and Zambaldi, Vinicius and Malinowski, Mateusz and Tacchetti, Andrea and Raposo, David and Santoro, Adam and Faulkner, Ryan and others},
  journal={arXiv preprint arXiv:1806.01261},
  volume={2},
  number={3},
  pages={5},
  year={2018}
}

@article{li2020neural,
  title={Neural operator: Graph kernel network for partial differential equations},
  author={Li, Zongyi and Kovachki, Nikola and Azizzadenesheli, Kamyar and Liu, Burigede and Bhattacharya, Kaushik and Stuart, Andrew and Anandkumar, Anima},
  journal={arXiv preprint arXiv:2003.03485},
  year={2020}
}

\newpage

\setcounter{figure}{0}
\setcounter{table}{0}
\renewcommand{\thefigure}{S\arabic{figure}}
\renewcommand{\thetable}{S\arabic{table}}

\section{Supplementary Section 1: Deviatoric Proportionality of \texorpdfstring{$\bs{\sigma}$}{sigma} and \texorpdfstring{$\bs{B}$}{B} for the Uncoupled Lee--Sacks Model}

\subsection{Kinematic preliminaries}

Let $\bs{F}$ denote the deformation gradient and $J = \det(\bs{F}) > 0$ the
local volume ratio. The right and left Cauchy--Green tensors are
\begin{equation}
    \bs{C} = \bs{F}^T\bs{F}, \qquad \bs{B} = \bs{F}\bs{F}^T ,
\end{equation}
both of which are symmetric positive definite and share the same invariants and in particular $\operatorname{tr}(\bs{C}) = \operatorname{tr}(\bs{B}) = I_1$. To separate volumetric from shape-changing deformation we introduce the isochoric part of the deformation gradient and the corresponding isochoric deformation tensors,
\begin{equation}
    \bs{F}^* = J^{-1/3}\bs{F}, \qquad
    \bs{C}^* = \bs{F}^{*T}\bs{F}^* = J^{-2/3}\bs{C}, \qquad
    \bs{B}^* = \bs{F}^*\bs{F}^{*T} = J^{-2/3}\bs{B},
\end{equation}
which satisfy $\det(\bs{C}^*) = \det(\bs{B}^*) = 1$ and therefore measure distortion at fixed volume. The first invariant of the isochoric deformation is
\begin{equation}
    I_1^* = \operatorname{tr}(\bs{C}^*) = J^{-2/3}I_1 .
\end{equation}
Finally, the deviatoric operator is defined by
\begin{equation}
    \operatorname{dev}(\cdot) = (\cdot) - \tfrac{1}{3}\operatorname{tr}(\cdot)\,\bs{I}.
\end{equation}

\subsection{Constitutive model}

The uncoupled Lee--Sacks model assumes an additive split of the strain energy density into an isochoric contribution, depending on $\bs{C}$ only through $I_1^*$, and a volumetric contribution depending only on $J$:
\begin{equation}
    \Psi(\bs{C}) = \Psi^*(I_1^*) + U(J),
\end{equation}
with
\begin{equation}
    \Psi^*(I_1^*) = \frac{c_0}{2}\left(I_1^* - 3\right)
    + \frac{c_1}{2}\left(e^{c_2\left(I_1^*-3\right)^2} - 1\right),
    \qquad
    U(J) = \frac{\kappa}{2}\left(\ln J\right)^2 ,
\end{equation}
where $c_0, c_1, c_2, \kappa > 0$.

\subsection{Stress response}

The second Piola--Kirchhoff and Cauchy stresses follow from the strain energy as,
\begin{equation}
    \bs{S} = 2\pd{\Psi}{\bs{C}}, \qquad
    \bs{\sigma} = J^{-1}\bs{F}\bs{S}\bs{F}^T .
\end{equation}
Evaluating the derivative requires the two elementary identities
\begin{equation}
    \pd{I_1}{\bs{C}} = \bs{I}, \qquad
    \pd{J}{\bs{C}} = \frac{1}{2}J\bs{C}^{-1},
\end{equation}
from which the derivative of the volumetric scaling factor is obtained by the chain rule,
\begin{equation}
    \pd{J^{-2/3}}{\bs{C}}
    = -\frac{2}{3}J^{-5/3}\cdot\frac{1}{2}J\bs{C}^{-1}
    = -\frac{1}{3}J^{-2/3}\bs{C}^{-1}.
\end{equation}
Applying the product rule to $I_1^* = J^{-2/3}I_1$ then gives
\begin{equation}
    \pd{I_1^*}{\bs{C}}
    = J^{-2/3}\bs{I} + I_1\left(-\frac{1}{3}J^{-2/3}\bs{C}^{-1}\right)
    = J^{-2/3}\left(\bs{I} - \frac{1}{3}I_1\bs{C}^{-1}\right).
\end{equation}
Because the strain energy is additively split, so is the stress. Writing $\Psi^{*\prime} \equiv \pd{\Psi^*}{I_1^*}$ and $U^\prime(J) \equiv \pd{U}{J}$, the chain rule yields
\begin{equation}
    \bs{S}
    = \underbrace{2\Psi^{*\prime}\pd{I_1^*}{\bs{C}}}_{\bs{S}_\text{iso}}
    + \underbrace{2U^\prime(J)\pd{J}{\bs{C}}}_{\bs{S}_\text{vol}},
\end{equation}
with the two contributions
\begin{align}
    \bs{S}_\text{vol} &= 2U^\prime(J)\pd{J}{\bs{C}} = J\,U^\prime(J)\,\bs{C}^{-1}, \\
    \bs{S}_\text{iso} &= 2\Psi^{*\prime}\pd{I_1^*}{\bs{C}}
    = 2\Psi^{*\prime}J^{-2/3}\left(\bs{I} - \frac{1}{3}I_1\bs{C}^{-1}\right).
\end{align}
Since the push-forward $\bs{S}\mapsto J^{-1}\bs{F}\bs{S}\bs{F}^T$ is linear in $\bs{S}$, the split carries over to the Cauchy stress:
\begin{equation}
    \bs{\sigma}
    = J^{-1}\bs{F}\left(\bs{S}_\text{iso} + \bs{S}_\text{vol}\right)\bs{F}^T
    = \underbrace{J^{-1}\bs{F}\bs{S}_\text{iso}\bs{F}^T}_{\bs{\sigma}_\text{iso}}
    + \underbrace{J^{-1}\bs{F}\bs{S}_\text{vol}\bs{F}^T}_{\bs{\sigma}_\text{vol}} .
\end{equation}
The volumetric part reduces to a spherical tensor, since $\bs{F}\bs{C}^{-1}\bs{F}^T = \bs{F}\bs{F}^{-1}\bs{F}^{-T}\bs{F}^T = \bs{I}$:
\begin{equation}
    \bs{\sigma}_\text{vol}
    = J^{-1}\bs{F}\left(J\,U^\prime(J)\bs{C}^{-1}\right)\bs{F}^T
    = U^\prime(J)\,\bs{I} .
\end{equation}
Thus the volumetric term contributes a pure hydrostatic pressure and carries no deviatoric content. For the isochoric part, use $\bs{F}\bs{I}\bs{F}^T = \bs{B}$ together with $\bs{F}\bs{C}^{-1}\bs{F}^T = \bs{I}$ and $I_1 = \operatorname{tr}(\bs{C}) = \operatorname{tr}(\bs{B})$:
\begin{align}
    \bs{\sigma}_\text{iso}
    &= J^{-1}\left[2\Psi^{*\prime}J^{-2/3}\bs{F}
       \left(\bs{I} - \frac{1}{3}I_1\bs{C}^{-1}\right)\bs{F}^T\right]
     = \frac{2\Psi^{*\prime}}{J}J^{-2/3}
       \left[\bs{B} - \frac{1}{3}\operatorname{tr}(\bs{B})\bs{I}\right] \\
    &= \frac{2\Psi^{*\prime}}{J}
       \left[\bs{B}^* - \frac{1}{3}\operatorname{tr}(\bs{B}^*)\bs{I}\right]
     = \frac{2\Psi^{*\prime}}{J}\operatorname{dev}(\bs{B}^*) ,
\end{align}
so that the total Cauchy stress reads
\begin{equation}
    \bs{\sigma}
    = \frac{2\Psi^{*\prime}}{J}\operatorname{dev}(\bs{B}^*)
    + U^\prime(J)\,\bs{I} .
\end{equation}

\subsection{Deviatoric proportionality}

Taking the deviator of the preceding expression, the second term drops out because it is spherical, while the first is already traceless:
\begin{equation}
    \operatorname{dev}(\bs{\sigma}) = \frac{2\Psi^{*\prime}}{J}\operatorname{dev}(\bs{B}^*).
\end{equation}
The deviatoric operator is linear, so the constant factor $J^{-2/3}$ relating $\bs{B}^*$ and $\bs{B}$ passes through it,
\begin{equation}
    \operatorname{dev}(\bs{B}^*) = \operatorname{dev}\!\left(J^{-2/3}\bs{B}\right)
    = J^{-2/3}\operatorname{dev}(\bs{B}),
\end{equation}
and we obtain the desired proportionality
\begin{equation}\label{eq:proportionality}
    \boxed{\;\operatorname{dev}(\bs{\sigma}) = \alpha\,\operatorname{dev}(\bs{B}),
    \qquad \alpha = 2J^{-5/3}\Psi^{*\prime}. \;}
\end{equation}
It remains to establish the sign of $\alpha$. Differentiating the isochoric energy gives
\begin{equation}
    \Psi^{*\prime} = \pd{\Psi^*}{I_1^*}
    = \frac{c_0}{2} + c_1c_2\left(I_1^* - 3\right)e^{c_2\left(I_1^*-3\right)^2}.
\end{equation}
Because $\bs{C}^*$ is symmetric positive definite with unit determinant, the AM--GM inequality applied to its eigenvalues bounds its trace from below,
\begin{equation}
    I_1^* = \operatorname{tr}(\bs{C}^*) \geq 3\left(\det(\bs{C}^*)\right)^{1/3} = 3 .
\end{equation}
Hence $I_1^* - 3 \geq 0$, the exponential factor is strictly positive, and the second term is nonnegative for $c_1, c_2 > 0$. Therefore
\begin{equation}
    \Psi^{*\prime} \geq \frac{c_0}{2} > 0 ,
\end{equation}
and since $J > 0$ we conclude
\begin{equation}
    \operatorname{dev}(\bs{\sigma}) = \alpha\,\operatorname{dev}(\bs{B}),
    \qquad \alpha > 0 ,
\end{equation}
for every admissible deformation. The Cauchy stress deviator is thus a strictly positive multiple of the deviator of the left Cauchy--Green tensor, with the scalar $\alpha$ depending on the deformation only through $J$ and $I_1^*$.

\subsection{Coaxiality of \texorpdfstring{$\bs{\sigma}$}{sigma} and \texorpdfstring{$\bs{B}$}{B}}

Two symmetric tensors are said to be \emph{coaxial} if they admit a common orthonormal eigenbasis, that is, if their principal directions coincide. We now show that the proportionality \eqref{eq:proportionality} implies this property for $\bs{\sigma}$ and $\bs{B}$. Reconstructing the Cauchy stress from its deviatoric and spherical parts and substituting \eqref{eq:proportionality} gives
\begin{align}
    \bs{\sigma}
    &= \operatorname{dev}(\bs{\sigma}) + \frac{1}{3}\operatorname{tr}(\bs{\sigma})\bs{I}
     = \alpha\operatorname{dev}(\bs{B}) + \frac{1}{3}\operatorname{tr}(\bs{\sigma})\bs{I} \\
    &= \alpha\left[\bs{B} - \frac{1}{3}\operatorname{tr}(\bs{B})\bs{I}\right]
       + \frac{1}{3}\operatorname{tr}(\bs{\sigma})\bs{I}
     = \alpha\bs{B} + \beta\bs{I},
\end{align}
where the scalar $\beta$ collects the two spherical contributions,
\begin{equation}
    \beta = \frac{1}{3}\left[\operatorname{tr}(\bs{\sigma}) - \alpha\operatorname{tr}(\bs{B})\right].
\end{equation}
The Cauchy stress is therefore an isotropic tensor function of $\bs{B}$ alone, consisting of a multiple of $\bs{B}$ plus a spherical term. Since $\bs{B}$ is symmetric positive definite, the spectral theorem guarantees an orthonormal basis $\{\bs{n}_i\}_{i=1}^{3}$ of eigenvectors of $\bs{B}$ with strictly positive eigenvalues, written in terms of the principal stretches
$\lambda_i > 0$ as
\begin{equation}
    \bs{B}\bs{n}_i = \lambda_i^2\,\bs{n}_i, \qquad i = 1,2,3 .
\end{equation}
Applying $\bs{\sigma} = \alpha\bs{B} + \beta\bs{I}$ to each such eigenvector, and noting that every vector is an eigenvector of $\bs{I}$, yields
\begin{equation}
    \bs{\sigma}\bs{n}_i
    = \alpha\bs{B}\bs{n}_i + \beta\bs{n}_i
    = \alpha\lambda_i^2\,\bs{n}_i + \beta\bs{n}_i
    = \left(\alpha\lambda_i^2 + \beta\right)\bs{n}_i .
\end{equation}
Thus each $\bs{n}_i$ is also an eigenvector of $\bs{\sigma}$, with
corresponding principal stress
\begin{equation}
    \sigma_i = \alpha\lambda_i^2 + \beta .
\end{equation}
The basis $\{\bs{n}_i\}$ is therefore principal for both tensors. This means $\bs{\sigma}$ and $\bs{B}$ are coaxial, and the principal stresses are obtained from the squared principal stretches by the same affine map for all three directions. In particular, the ordering of the principal stresses follows that of the principal stretches, since $\alpha > 0$.

\section{Supplementary Section 2: Details of the GNN and GNO baselines}

The GNN and GNO baselines are given the same inputs and exactly the same outputs as PCNO, so that the comparison isolates the architecture rather than the input representation. Using the notation from the Architecture Description section, both take the point-wise features $\mathbf{x}_0 \in \mathbb{R}^{N \times d_x}$ and the scalar conditioning variables $\mathbf{c}_0 \in \mathbb{R}^{1 \times d_c}$ constructed in the Local Geometric Features and Global Conditioning Variables sections, and both predict $\mathbf{y}\in\mathbb{R}^{N\times(3+6+6)}$, partitioned into displacement, Voigt-form Lagrangian strain, and Voigt-form Cauchy stress as for PCNO. The essential architectural difference is that neither baseline uses the shift--scale--gate conditioning mechanism of PCNO: the conditioning variables are not embedded into their own latent space and never generate modulation parameters, but instead enter as uniform input features that are broadcast across all nodes and all edges.

Unlike PCNO, which operates directly on the point cloud through attention, both baselines require an explicit graph. For each sample we therefore construct a $k$-nearest-neighbor graph on the reference coordinates $\mathbf{X}$, with the neighbor set of node $i$ denoted $\mathcal{N}_i^k$, where $i \notin \mathcal{N}_i^k$ and padded (ghost) nodes are excluded from the candidate set using the same binary mask described in the Training Details section. This $k$ controls graph connectivity only and is distinct from the neighborhood size used to build the structure tensor features. Because nodes are randomly sampled at every optimization step, the graph is not a fixed mesh connectivity but is rebuilt for each sample at each step, so the baselines see the same stochastic point clouds as PCNO.

The point-wise features are lifted into the latent space of dimension $d$ by a single dense layer that also carries the conditioning variables, which are broadcast across all $N$ nodes and concatenated channel-wise before the projection,
\begin{equation}
    \mathbf{v}^{(0)} = \mathrm{Dense}\left(\mathrm{concat}\left(\mathbf{x}_0,\mathrm{broadcast}(\mathbf{c}_0)\right)\right) \in \mathbb{R}^{N\times d}, \quad \mathrm{broadcast}(\mathbf{c}_0) \in \mathbb{R}^{N\times d_c},
\end{equation}
so that the local geometric features and the global conditioning variables are mixed once, additively, at the input rather than at every layer. 

The conditioning variables enter the message passing itself in the same uniform fashion, as a constant block appended to every edge feature vector. For each directed edge from $n \in \mathcal{N}_i^k$ into node $i$ we form
\begin{equation}
    \mathbf{e}_{in} = \mathrm{concat}\left( \mathbf{X}_i,~\mathbf{X}_n,~\mathbf{X}_i - \mathbf{X}_n,~\lVert \mathbf{X}_i - \mathbf{X}_n \rVert_2,~\mathbf{c}_0 \right) \in \mathbb{R}^{10 + d_c},
\end{equation}
in which the first ten channels carry the absolute, relative, and metric geometry of the edge and the remaining $d_c$ channels are the conditioning vector broadcast unchanged over every edge of every graph. This is the standard way of conditioning a message-passing network, and it differs from PCNO in that $\mathbf{c}_0$ acts only as a constant additive input to the kernel and message networks, whereas in PCNO the conditioning variables multiplicatively modulate the normalized latent features and gate the residual updates at every layer. Throughout, $\mathcal{N}_i^k$ contains only valid mesh nodes, and all neighborhood averages below are mask-weighted so that padded nodes contribute neither feature content nor normalization weight to a node update.

The GNO baseline is a stack of $L$ kernel integral layers, each combining a point-wise linear map with a learned kernel acting over the neighborhood,
\begin{equation}
    \mathbf{v}_i^{(\ell+1)} = \mathrm{GELU}\left( \mathrm{Dense}\left( \mathbf{v}_i^{(\ell)} \right) + \frac{1}{\lvert \mathcal{N}_i^k \rvert} \sum_{n \in \mathcal{N}_i^k} \kappa^{(\ell)}\left( \mathbf{e}_{in} \right) \odot \mathbf{v}_n^{(\ell)} \right),
\end{equation}
where $\kappa^{(\ell)} : \mathbb{R}^{10 + d_c} \to \mathbb{R}^{d}$ is an MLP with GELU activations and $\odot$ again denotes element-wise multiplication. We use a diagonal kernel, meaning that $\kappa^{(\ell)}$ emits $d$ channel-wise gains on each edge, rather than a full matrix-valued kernel that would emit a dense $d \times d$ operator. The full kernel requires $d^2$ outputs on each of the $N k$ edges of every sample, which at the latent width used here amounts to more than $4 \times 10^5$ values per edge and is intractable in both memory and compute at these node counts; 
the diagonal restriction is what permits the GNO baseline to be run at the same latent width as PCNO rather than at a width smaller by more than an order of magnitude.

The GNN baseline replaces the kernel integral with a generic learned message that additionally sees the latent features of both endpoints,
\begin{equation}
    \mathbf{v}_i^{(\ell+1)} = \mathrm{GELU}\left( \mathrm{Dense}\left( \mathbf{v}_i^{(\ell)} \right) + \frac{1}{\lvert \mathcal{N}_i^k \rvert} \sum_{n \in \mathcal{N}_i^k} \phi^{(\ell)}\left( \mathrm{concat}\left( \mathbf{v}_i^{(\ell)},~ \mathbf{v}_n^{(\ell)}, ~\mathbf{e}_{in} \right) \right) \right),
\end{equation}
with $\phi^{(\ell)} : \mathbb{R}^{2d + 10 + d_c} \to \mathbb{R}^{d}$ again an MLP with GELU activations. The two baselines are thus identical apart from the form of the neighborhood term, and the GNN may be read as the GNO with the multiplicative channel-wise kernel replaced by an unrestricted nonlinear function of the endpoint pair.

Both baselines are decoded by the same unconditioned output head, a stack of $L_o$ dense layers with GELU activations followed by a final linear projection to the $3 + 6 + 6$ output channels,
\begin{equation}
    \mathbf{y} = \mathrm{Dense}\left( \mathrm{MLP}\left( \mathbf{v}^{(L)} \right) \right) \in \mathbb{R}^{N \times (3+6+6)},
\end{equation}
which are partitioned into $\tilde{\mathbf{u}}$, $\tilde{\mathbf{E}}$, and $\tilde{\boldsymbol{\sigma}}$ exactly as for PCNO. As in PCNO, the final layer of the head is initialized to zero; unlike PCNO, the head receives no modulation from the conditioning variables, consistent with the design choice that $\mathbf{c}_0$ enters only as an input feature.

For both baselines we set the latent dimension to $d = 640$, matching the Large PCNO variant of Supplementary Table 2, with $L = 8$ layers, $L_o = 3$ output head layers of width $640$, and $k = 8$ graph neighbors. The GNO kernel network $\kappa^{(\ell)}$ has three layers of width $640$, and the GNN message network $\phi^{(\ell)}$ has two layers of width $640$, the additional depth on the GNO side compensating for the restricted multiplicative form of its kernel. All remaining weights are initialized with a Xavier/Glorot uniform initializer and all biases to zero, as for PCNO.

All training settings are identical to those used for PCNO and are not repeated here: the optimizer and its hyperparameters, the learning rate schedule, gradient clipping, batch size, weight decay, node subsampling, zero padding and masking, normalization statistics, EMA of the model parameters, and the MAE loss of Eq.~(58) are exactly as described in the Training Details section. The baselines differ from PCNO only in the architecture described above.

\section{Supplementary Figures and Tables}

\begin{figure}[H]
    \centering
    \includegraphics[width=1\linewidth]{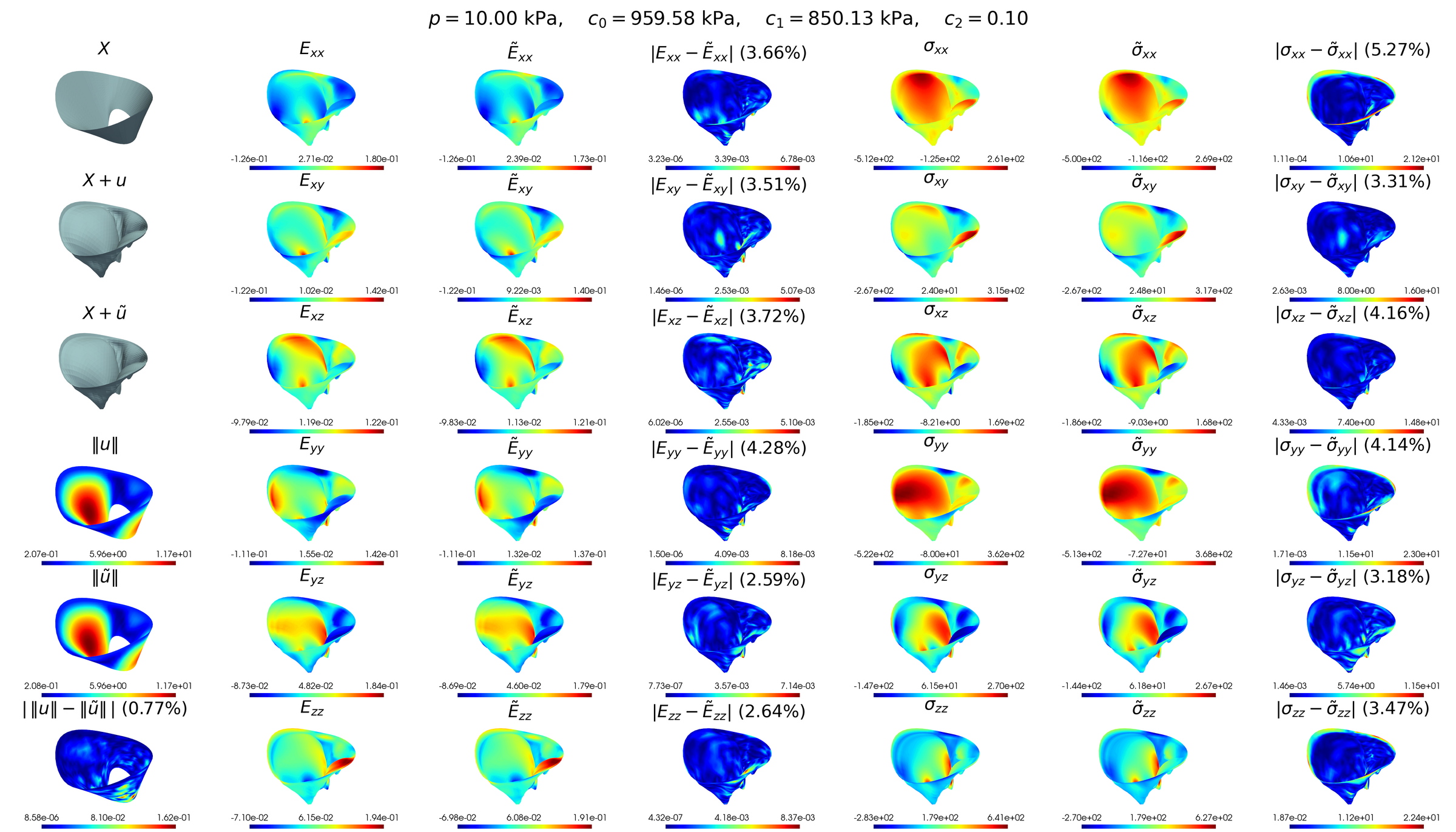}
    \caption{Test example from the in-distribution setting corresponding to a mitral valve, showing the individual stress and strain components along with their point-wise absolute errors and relative $L^2$ errors in parentheses.}
    \label{fig:id_example2}
\end{figure}

\begin{figure}[H]
    \centering
    \includegraphics[width=1\linewidth]{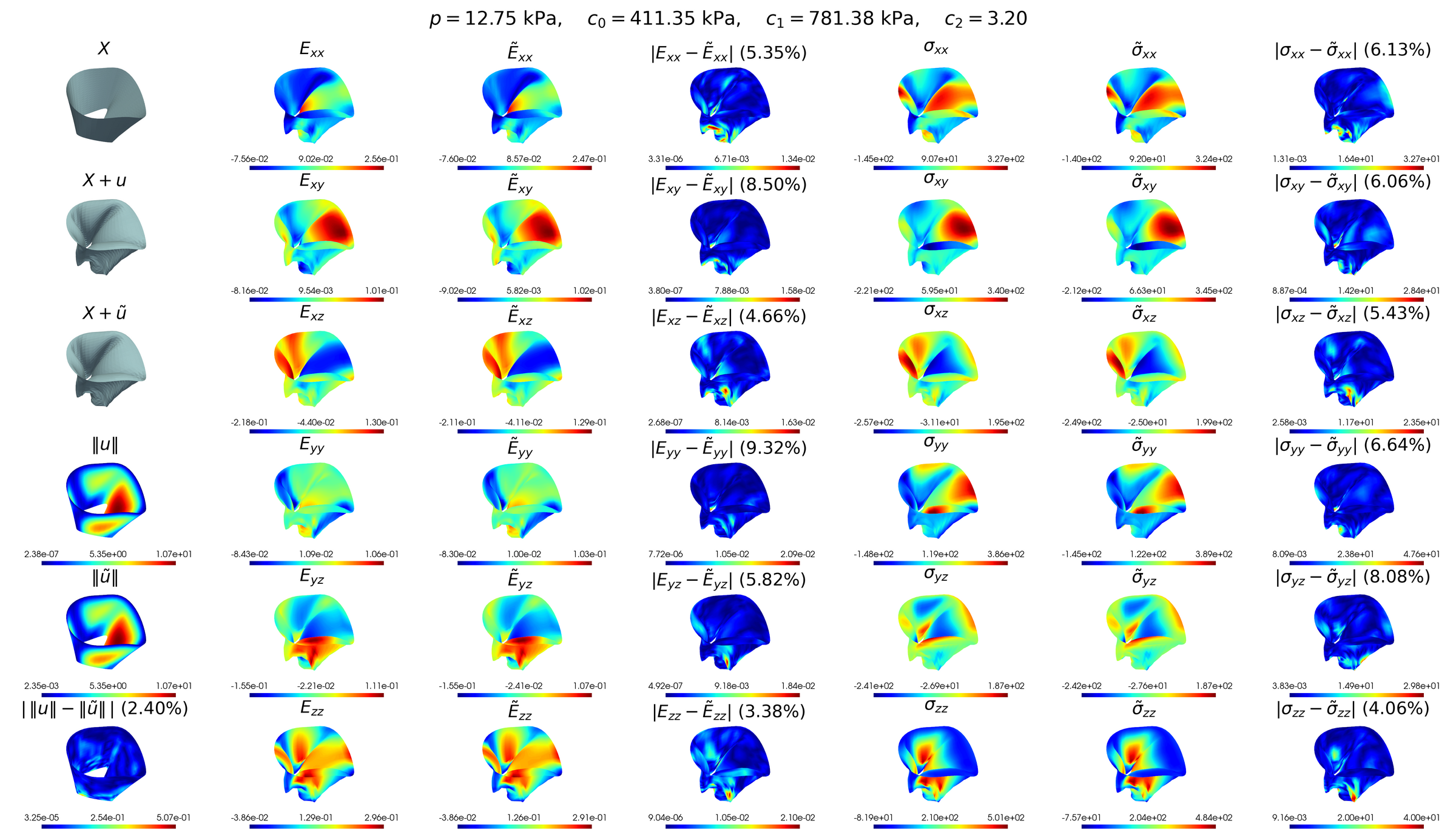}
    \caption{Test example from the SBP generalization setting corresponding to a tricuspid valve, showing the individual stress and strain components along with their point-wise absolute errors and relative $L^2$ errors in parentheses.}
    \label{fig:sbp_example2}
\end{figure}

\begin{figure}[H]
    \centering
    \includegraphics[width=1\linewidth]{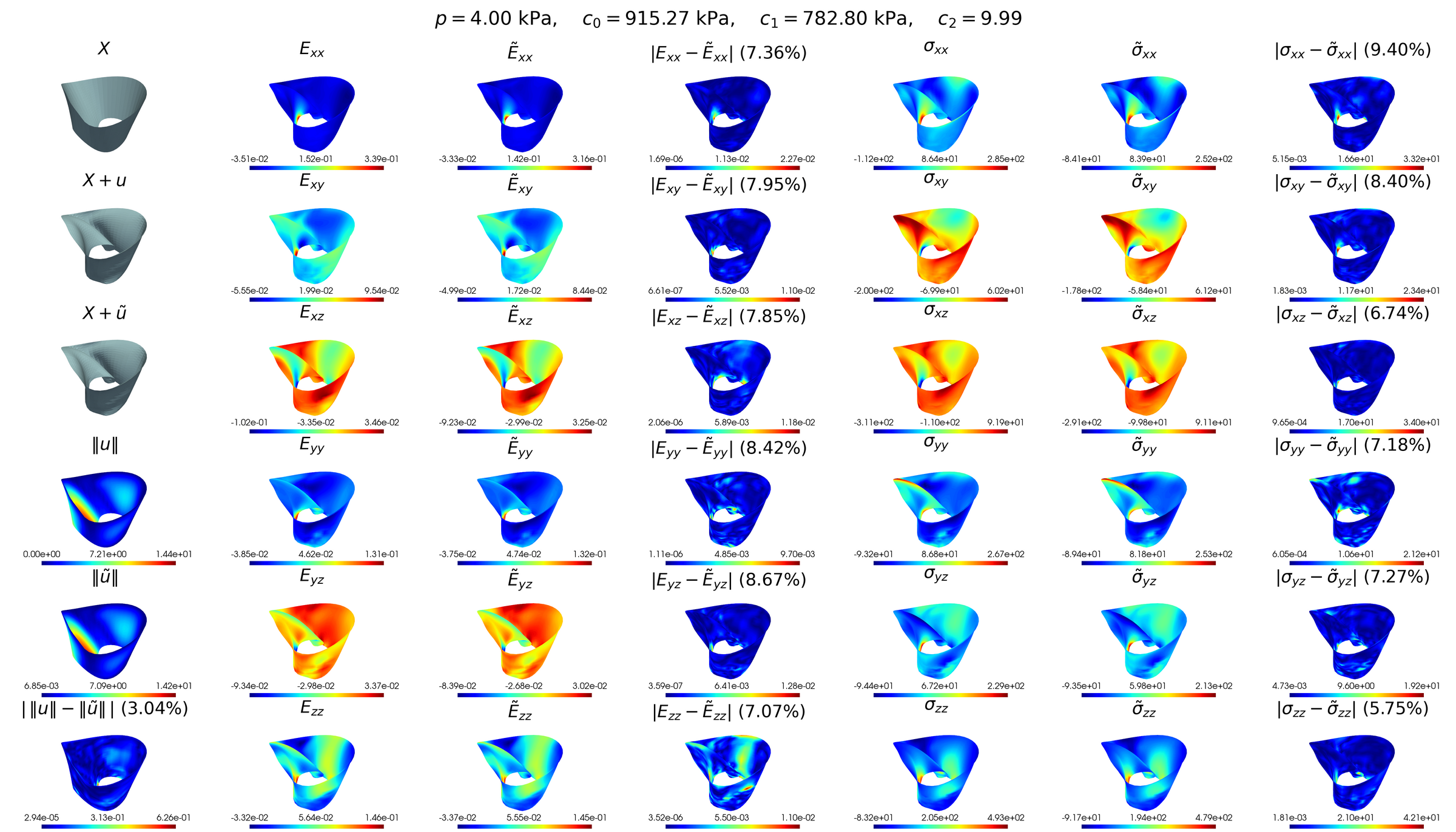}
    \caption{Test example from the trivariate material parameter generalization setting corresponding to a tricuspid valve, showing the individual stress and strain components along with their point-wise absolute errors and relative $L^2$ errors in parentheses.}
    \label{fig:mat_param_example2}
\end{figure}

\begin{figure}[H]
    \centering
    \includegraphics[width=1\linewidth]{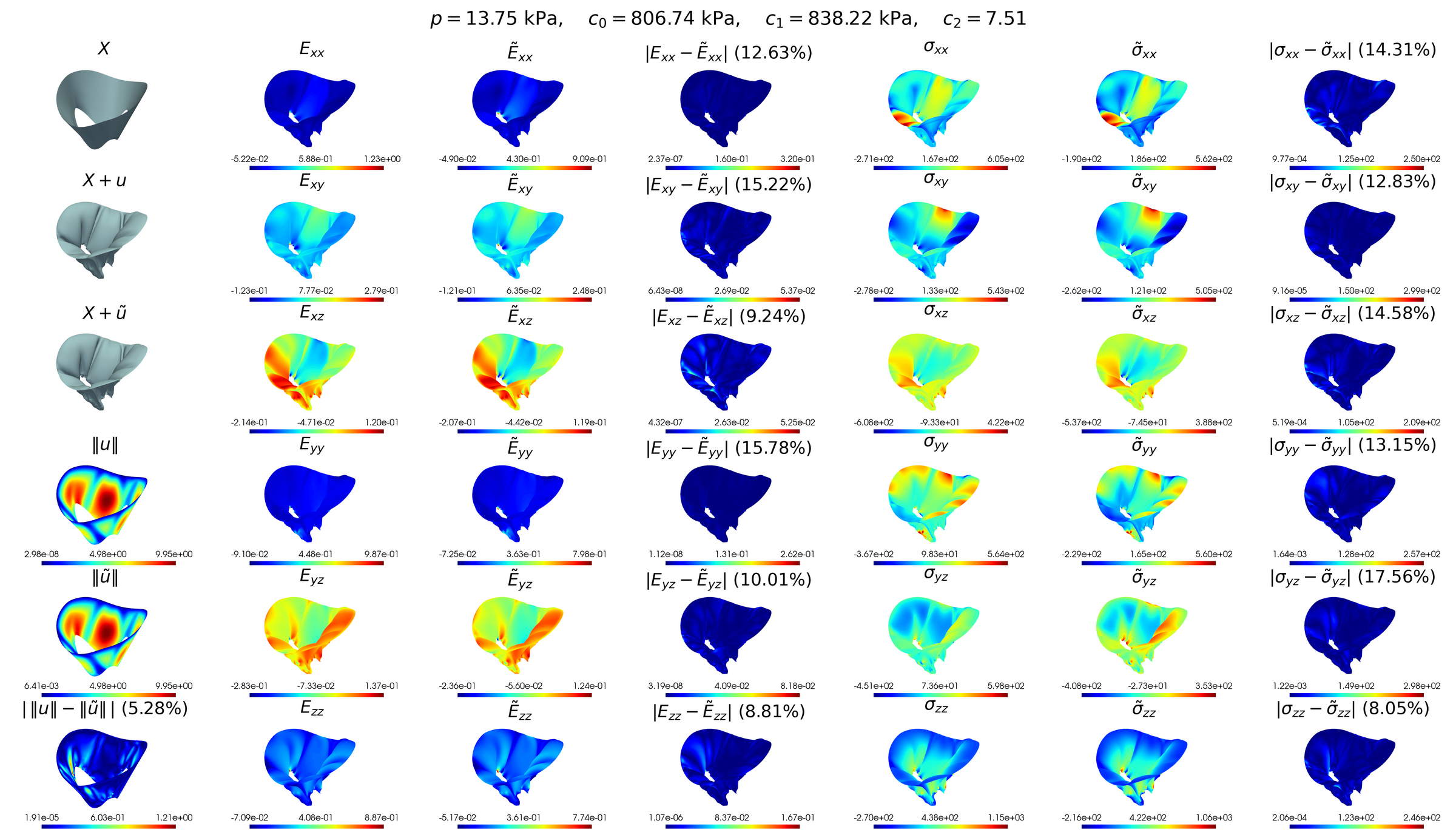}
    \caption{Test example from the joint SBP and material parameter generalization setting corresponding to a tricuspid valve, showing the individual stress and strain components along with their point-wise absolute errors and relative $L^2$ errors in parentheses.}
    \label{fig:mat_param_sbp_example2}
\end{figure}

\begin{figure}[H]
    \centering
    \includegraphics[width=1\linewidth]{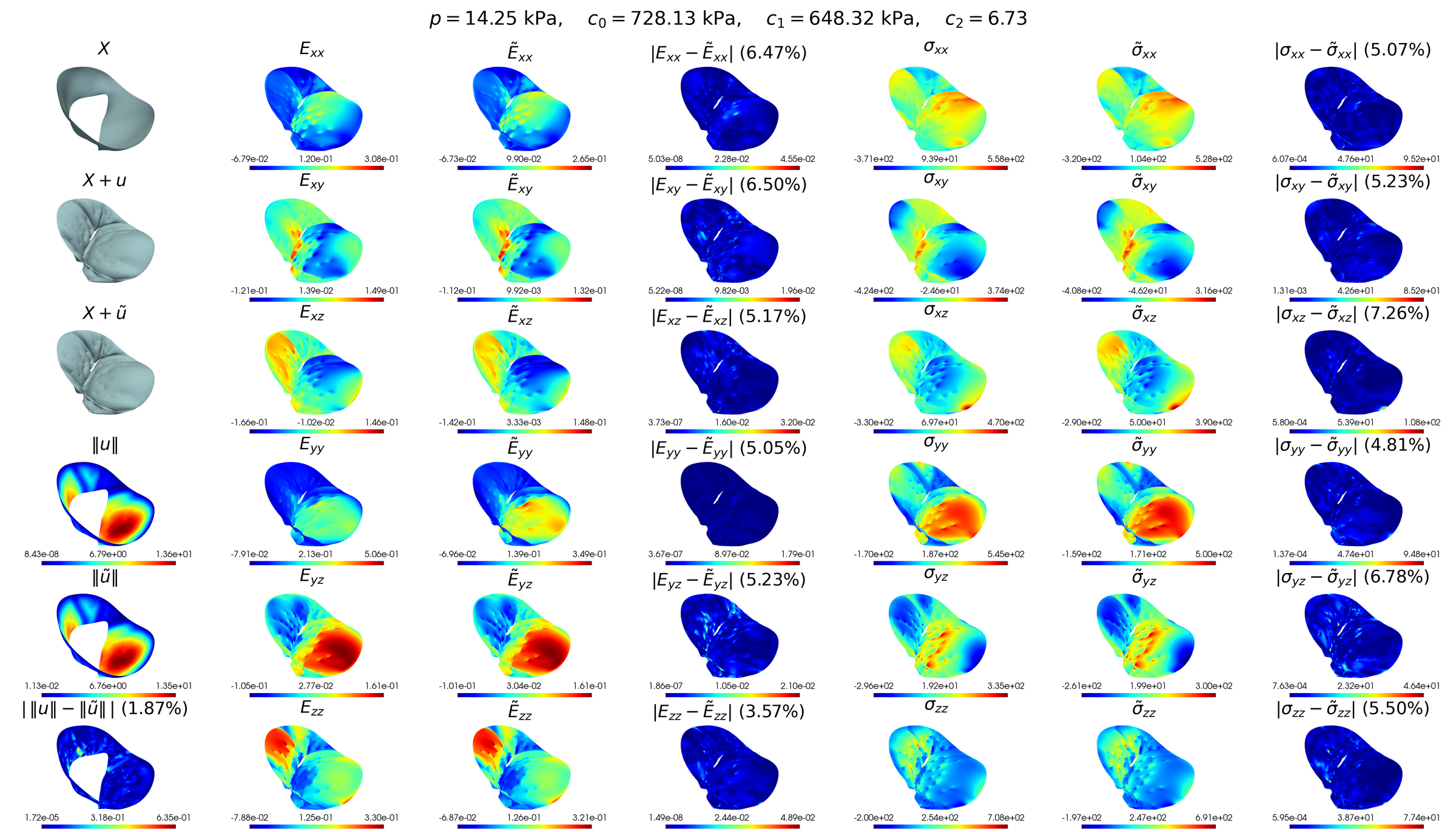}
    \caption{Test example from the diseased joint SBP and material parameter generalization setting corresponding to a mitral valve, showing the individual stress and strain components along with their point-wise absolute errors and relative $L^2$ errors in parentheses.}
    \label{fig:diseases_mat_param_sbp_example2}
\end{figure}

\newpage

\begin{figure}[H]
    \centering
    \includegraphics[width=1\linewidth]{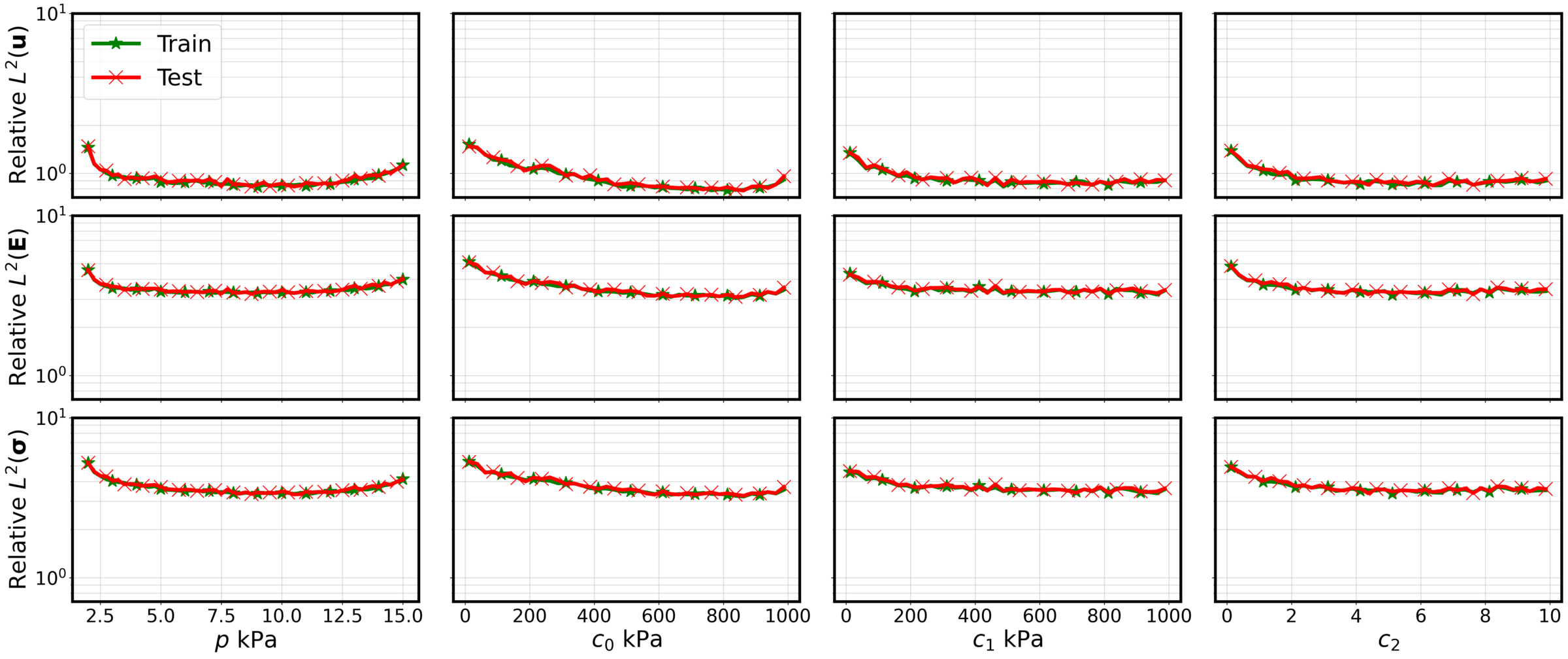}
    \caption{Relative errors as functions of the physical conditioning variables for the in-distribution setting.}
    \label{fig:id_error_space}
\end{figure}

\begin{figure}[H]
    \centering
    \includegraphics[width=1\linewidth]{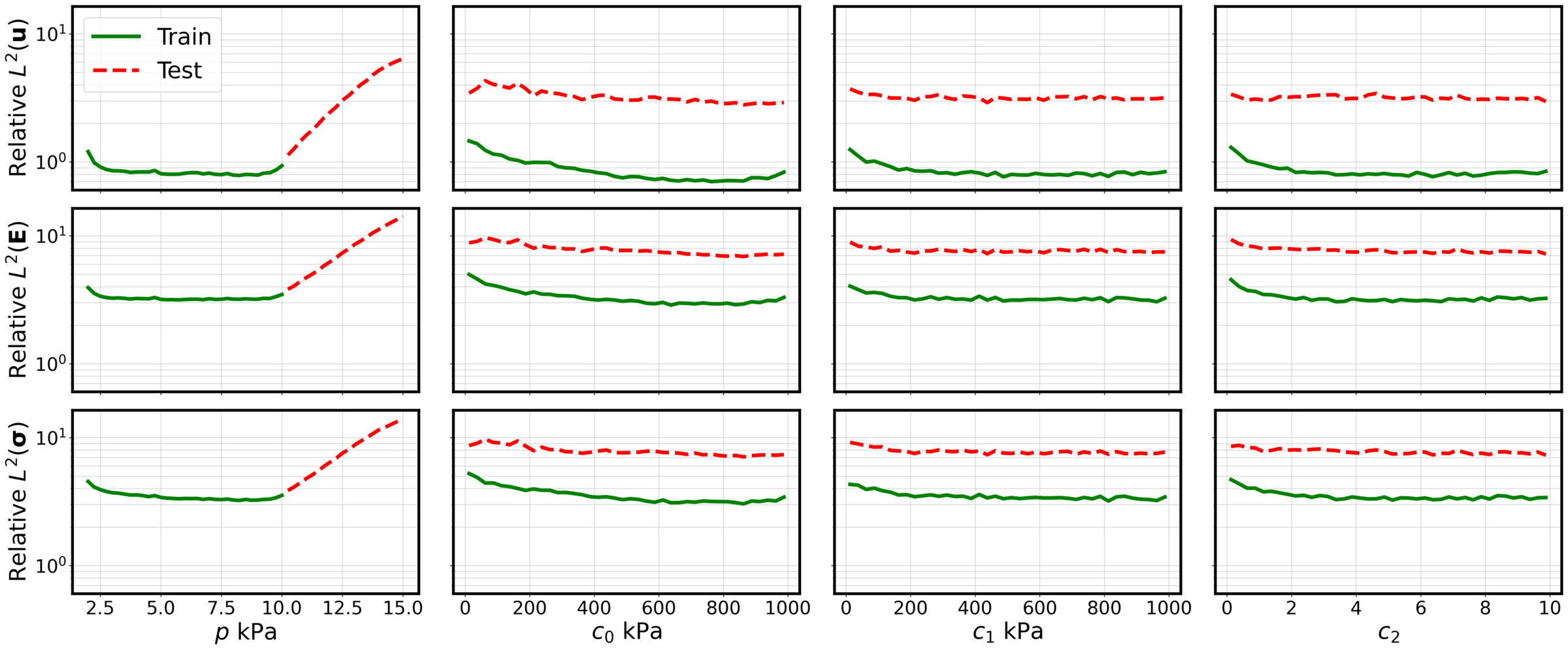}
    \caption{Relative errors as functions of the physical conditioning variables for the SBP generalization setting.}
    \label{fig:sbp_error_space}
\end{figure}

\begin{figure}[H]
    \centering
    \includegraphics[width=1\linewidth]{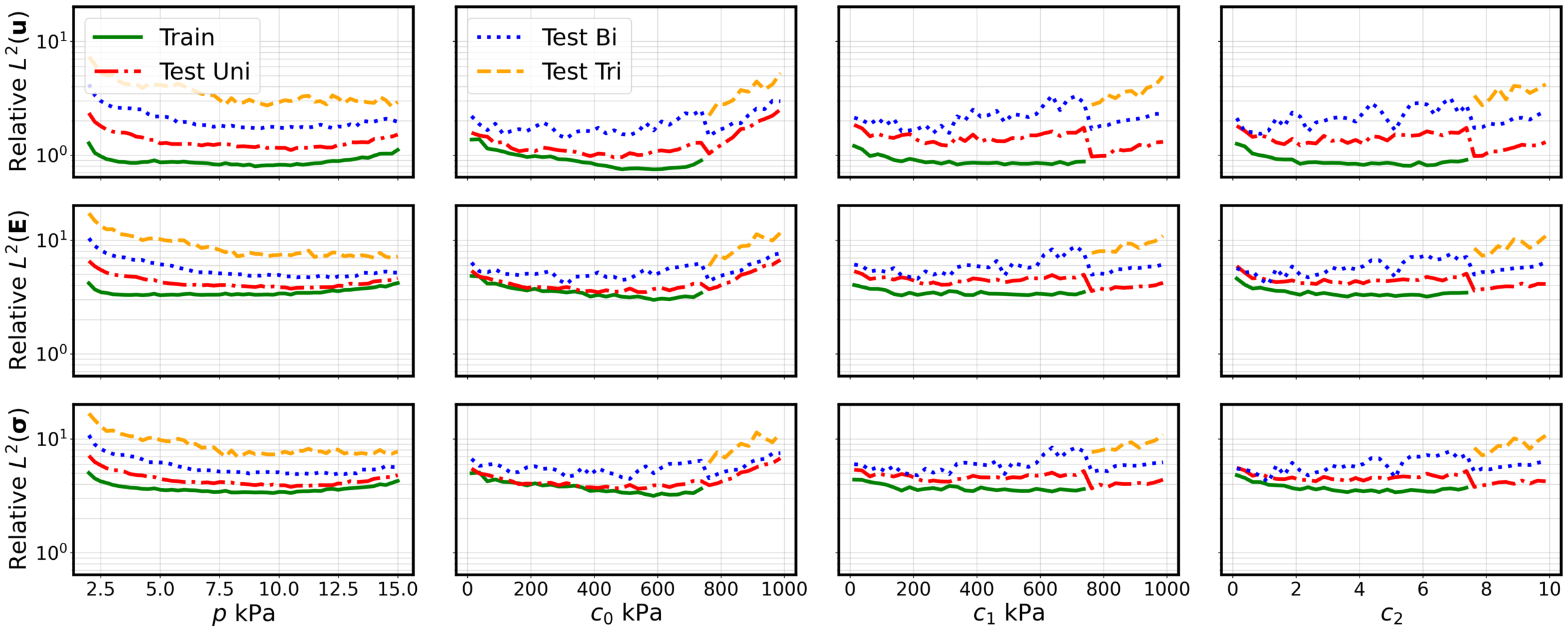}
    \caption{Relative errors as functions of the physical conditioning variables for the Uni, Bi, and Tri material parameter generalization setting.}
    \label{fig:mat_param_error_space}
\end{figure}

\begin{figure}[H]
    \centering
    \includegraphics[width=1\linewidth]{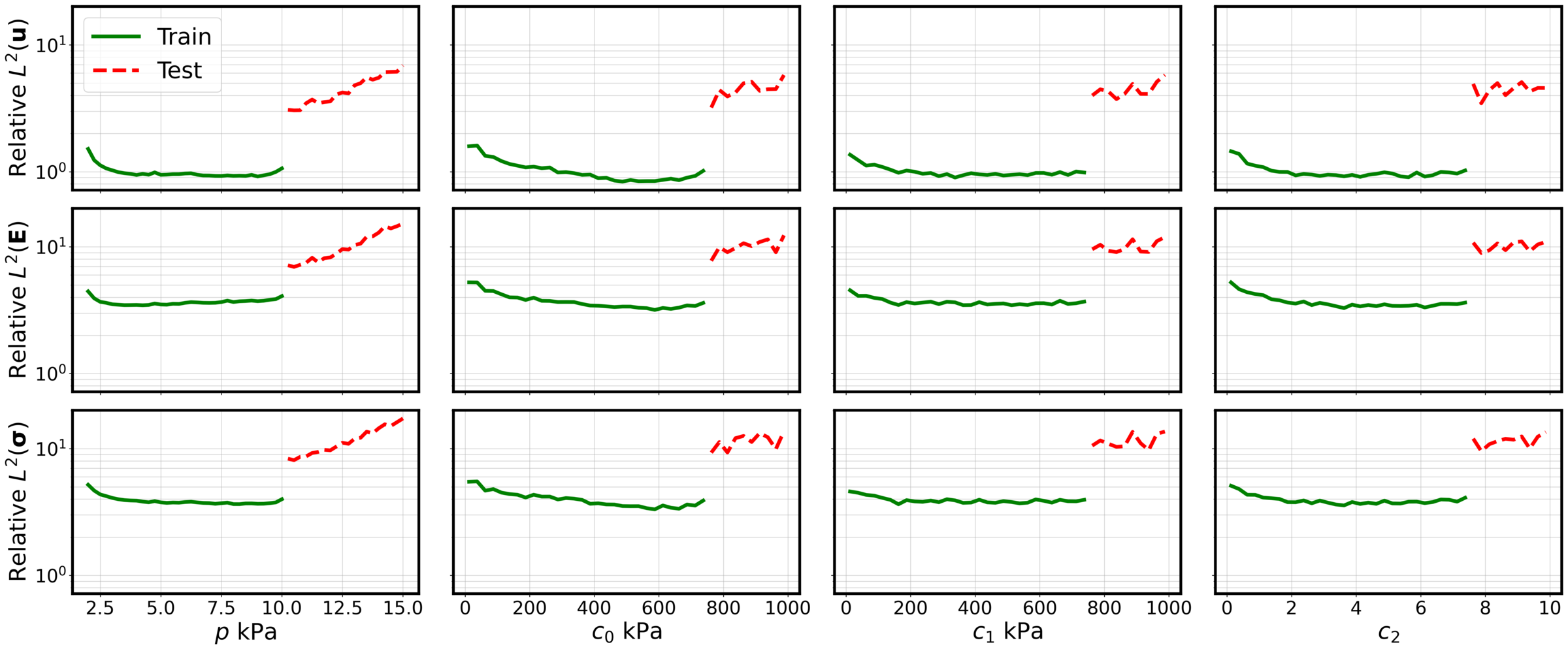}
    \caption{Relative errors as functions of the physical conditioning variables for the joint SBP and material parameter generalization setting.}
    \label{fig:mat_param_sbp_error_space}
\end{figure}

\begin{figure}[H]
    \centering
    \includegraphics[width=1\linewidth]{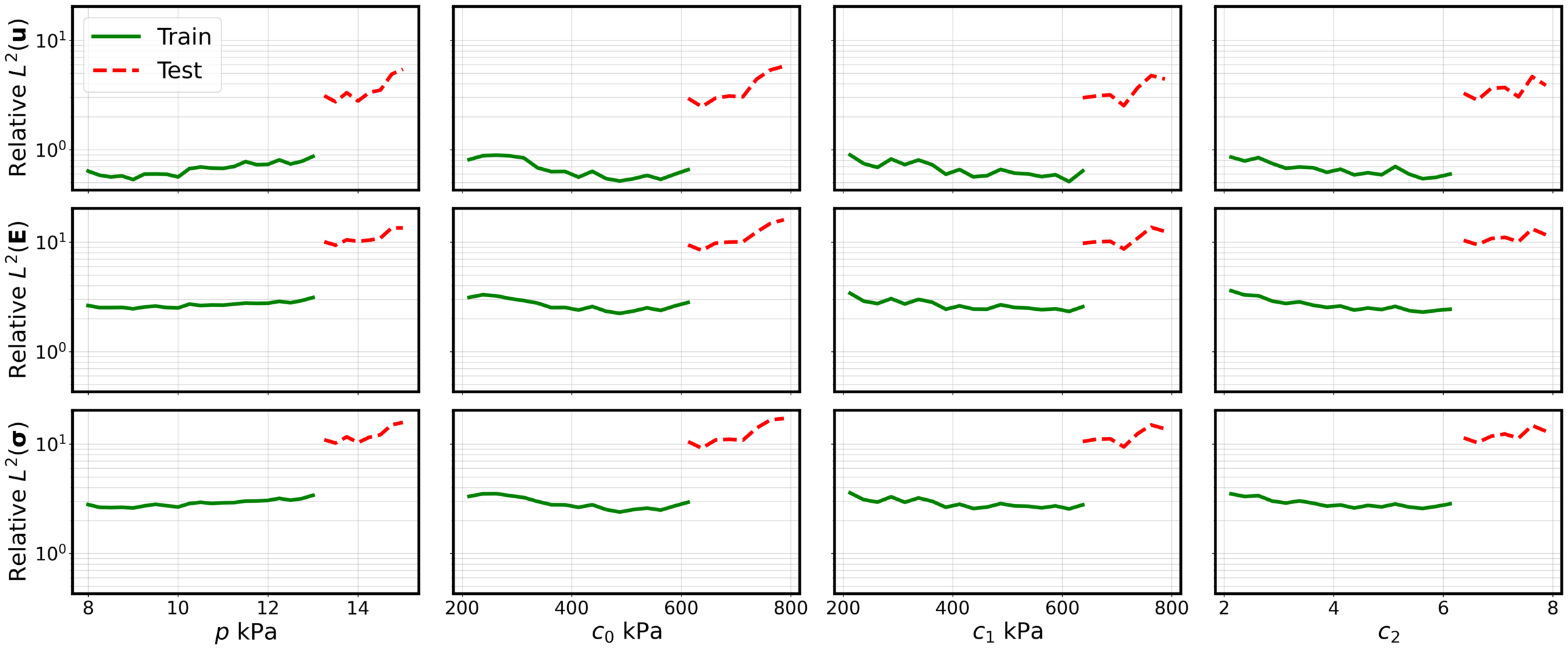}
    \caption{Relative errors as functions of the physical conditioning variables for the diseased joint SBP and material parameter generalization setting.}
    \label{fig:diseased_mat_param_sbp_error_space}
\end{figure}

\begin{table}[H]
\centering
\begin{tabular}{lllccc}
\toprule
\textbf{Dataset} & \textbf{Setting} & \textbf{Split}
  & \multicolumn{3}{c}{\textbf{Geometric variability}} \\
\cmidrule(lr){4-6}
 & & 
  & \makecell{Same \\ geometry}
  & \makecell{Diverse \\ geometry}
  & \makecell{Moving \\ boundary} \\
\midrule
\multirow{11}{*}{Base}
 & \multirow{2}{*}{In-distribution}
   & Train & $59{,}944$  & $113{,}681$ & $91{,}796$ \\
 & & Test  & $14{,}986$  & $28{,}421$  & $22{,}949$ \\
\cmidrule(lr){2-6}
 & \multirow{2}{*}{SBP Generalization}
   & Train & $51{,}803$  & $104{,}637$ & $86{,}418$ \\
 & & Test  & $23{,}127$  & $37{,}465$  & $28{,}327$ \\
\cmidrule(lr){2-6}
 & \multirow{4}{*}{\makecell[l]{Material Parameter \\ Generalization}}
   & Train      & $29{,}951$ & $53{,}124$ & $41{,}521$ \\
 & & Test (Uni) & $32{,}930$ & $62{,}368$ & $52{,}651$ \\
 & & Test (Bi)  & $10{,}581$ & $24{,}063$ & $18{,}764$ \\
 & & Test (Tri) & $1{,}468$  & $2{,}547$  & $1{,}809$  \\
\cmidrule(lr){2-6}
 & \multirow{2}{*}{\makecell[l]{Joint SBP and Material \\ Parameter Generalization}}
   & Train & $21{,}167$ & $40{,}656$ & $33{,}279$ \\
 & & Test  & $520$      & $783$      & $558$      \\
\midrule
 & & & \multicolumn{3}{c}{\textbf{Pathology}} \\
\cmidrule(lr){4-6}
 & & 
  & Tethering
  & \makecell{P2 \\ Prolapse}
  & Dilation \\
\midrule
\multirow{4}{*}{Diseased}
 & \multirow{2}{*}{In-distribution}
   & Train & $11{,}110$ & $10{,}618$ & $11{,}112$ \\
 & & Test  & $2{,}778$  & $2{,}655$  & $2{,}779$  \\
\cmidrule(lr){2-6}
 & \multirow{2}{*}{\makecell[l]{Joint SBP and Material \\ Parameter Generalization}}
   & Train & $3{,}546$ & $3{,}388$ & $3{,}546$ \\
 & & Test  & $103$     & $98$      & $103$     \\
\bottomrule
\end{tabular}
\caption{Number of training and test examples across generalization settings on the base and diseased datasets. The Base dataset is partitioned by geometric variability (Same geometry, Diverse geometry, Moving boundary), while the Diseased dataset is partitioned by pathology (Tethering, P2 prolapse, Dilation). Material parameter generalization cases (Uni, Bi, Tri) share identical training data.}
\label{tab:splits_summary}
\end{table}

\begin{table}[H]
\centering
\begin{tabular}{lccccccc}
\toprule
\textbf{Model} & $h$ & $d$ & $r$ & $L$ & $L_o$ & \textbf{\# Parameters} \\
\midrule 
Tiny & 4 & 128 & 2 & 2 & 1 & 0.6 M \\
Small & 12 & 384 & 4 & 4 & 1 & 12 M \\
Big & 16 & 512 & 6 & 6 & 2 & 38 M \\
Large & 20 & 640 & 8 & 8 & 3 & 93 M \\
\bottomrule
\end{tabular}
\caption{Details of model variants. $h$ denotes the number of heads, $d$ the latent space dimension, $r$ the expansion ratio, $L$ the number of encoder layers, $L_o$ the number of output MLP head layers.}
\end{table}

\begin{table}[H]
\centering
\begin{tabular}{lll ccc}
\toprule
\textbf{Setting} & \textbf{Ablation} & \textbf{Variant}
  & Rel. $L^2(\boldsymbol{u})$ & Rel. $L^2(\boldsymbol{E})$ & Rel. $L^2(\boldsymbol{\sigma})$ \\
\midrule
\multirow{13}{*}{\makecell[l]{In-\\distribution}}
 & \multirow{2}{*}{EMA}
   & No EMA & 1.06 & 3.87 & 4.00 \\
 & & EMA    & $0.93$ \footnotesize{($12.26\%$)}
            & $3.48$ \footnotesize{($10.08\%$)}
            & $3.69$ \footnotesize{($7.75\%$)} \\
\cmidrule(lr){2-6}
 & \multirow{2}{*}{Loss function}
   & MSE & 1.29 & 3.83 & 4.01 \\
 & & MAE & $0.93$ \footnotesize{($27.91\%$)}
         & $3.48$ \footnotesize{($9.14\%$)}
         & $3.69$ \footnotesize{($7.98\%$)} \\
\cmidrule(lr){2-6}
 & \multirow{4}{*}{Model Size}
   & Tiny  & 3.99 & 12.82 & 13.24 \\
 & & Small & 1.39 \footnotesize{($65.16\%$)}
           & 4.62 \footnotesize{($63.96\%$)}
           & 4.81 \footnotesize{($63.67\%$)} \\
 & & Big   & 1.06 \footnotesize{($23.74\%$)}
           & 3.77 \footnotesize{($18.40\%$)}
           & 3.98 \footnotesize{($17.26\%$)} \\
 & & Large & $0.93$ \footnotesize{($12.26\%$)}
           & $3.48$ \footnotesize{($7.69\%$)}
           & $3.69$ \footnotesize{($7.29\%$)} \\
\cmidrule(lr){2-6}
 & \multirow{3}{*}{Attention}
   & PA \cite{wu2024transolver} & 0.97 \footnotesize{($-4.12\%$)} & 3.55 \footnotesize{($-1.97\%$)} & 3.76 \footnotesize{($-1.86\%$)} \\
 & & LA \cite{katharopoulos2020transformers} & 0.96 \footnotesize{($-3.12\%$)} & 3.52 \footnotesize{($-1.14\%$)} & 3.75 \footnotesize{($-1.60\%$)} \\
 & & RALA \cite{fan2025breaking} & 0.93 & 3.48 & 3.69 \\
\midrule
\multirow{13}{*}{\makecell[l]{Joint\\Gen.}}
 & \multirow{2}{*}{EMA}
   & No EMA & $4.65$ & $10.48$ & $11.88$ \\
 & & EMA    & $4.48$ \footnotesize{($3.66\%$)}
            & $10.11$ \footnotesize{($3.53\%$)}
            & $11.56$ \footnotesize{($2.69\%$)} \\
\cmidrule(lr){2-6}
 & \multirow{2}{*}{Loss function}
   & MSE & $4.91$ & $9.59$ & $10.15$ \\
 & & MAE & $4.48$ \footnotesize{($8.76\%$)}
         & $10.11$ \footnotesize{($-5.42\%$)}
         & $11.56$ \footnotesize{($-13.89\%$)} \\
\cmidrule(lr){2-6}
 & \multirow{4}{*}{Model Size}
   & Tiny  & $6.52$ & $17.76$ & $18.96$ \\
 & & Small & $4.57$ \footnotesize{($29.91\%$)}
           & $10.34$ \footnotesize{($41.78\%$)}
           & $11.49$ \footnotesize{($39.40\%$)} \\
 & & Big   & $3.96$ \footnotesize{($13.35\%$)}
           & $8.69$ \footnotesize{($15.96\%$)}
           & $9.99$ \footnotesize{($13.05\%$)} \\
 & & Large & $4.48$ \footnotesize{($-13.13\%$)}
           & $10.11$ \footnotesize{($-16.34\%$)}
           & $11.56$ \footnotesize{($-15.72\%$)} \\
\cmidrule(lr){2-6}
 & \multirow{3}{*}{Attention}
   & PA \cite{wu2024transolver} & $5.61$ \footnotesize{($-20.14\%$)} & $11.04$ \footnotesize{($-8.42\%$)} & $12.74$ \footnotesize{($-9.26\%$)} \\
 & & LA \cite{katharopoulos2020transformers} & $7.39$ \footnotesize{($-39.38\%$)} & $13.91$ \footnotesize{($-27.32\%$)} & $14.81$ \footnotesize{($-21.94\%$)} \\
 & & RALA \cite{fan2025breaking} & $4.48$ & $10.11$ & $11.56$ \\
\bottomrule
\end{tabular}
\caption{Ablation studies on EMA, loss function, model size, and attention mechanism (relative errors, \%) on the base dataset for the in-distribution and joint SBP and trivariate material parameter generalization (Joint Gen.) test settings. Values in parentheses show the relative improvement over the previous row of each ablation block. For the attention mechanism ablation, values in parentheses show the signed relative difference with respect to RALA; negative values indicate the baseline is worse than RALA. LA denotes Linear Attention, PA Physics-Attention, and RALA Rank-Augmented Linear Attention.}
\label{tab:ablations}
\end{table}

\begin{table}[H]
\centering
\begin{tabular}{lcccc}
\toprule
 & \textbf{Tiny} & \textbf{Small} & \textbf{Big} & \textbf{Large} \\
\midrule
\# Parameters   & 0.6 M    & 12 M  & 38 M  & 93 M  \\
Training time h & 1.20   & 2.06  & 6.06  & 12.29 \\
\midrule
Inference~ms ($N = 12{,}436$) & 1.25 & 9.81 & 25.78 & 55.57 \\
Inference~ms ($N = 11{,}917$) & 1.23 & 9.78 & 24.59 & 53.38 \\
Inference~ms ($N = 8{,}651$)  & 0.88 & 7.07 & 18.71 & 40.35 \\
Inference~ms ($N = 7{,}471$)  & 0.78 & 6.30 & 15.45 & 36.51 \\
Inference~ms ($N = 7{,}440$)  & 0.77 & 6.24 & 15.48 & 36.17 \\
Inference~ms ($N = 1{,}029$)  & 0.28 & 1.20 & 2.70  & 6.01  \\
\bottomrule
\end{tabular}
\caption{Comparison of parameter count, training time, and inference time (where $N$ represents the number of nodes). All timing measurements are performed on a single NVIDIA A6000 GPU and are averaged over 5{,}000 independent runs. Note that training and inference times were performed on our model with only a mechanical field MLP head (i.e., no classification head).}
\label{tab:times}
\end{table}

\begin{figure}[H]
    \centering
    \includegraphics[width=1\linewidth]{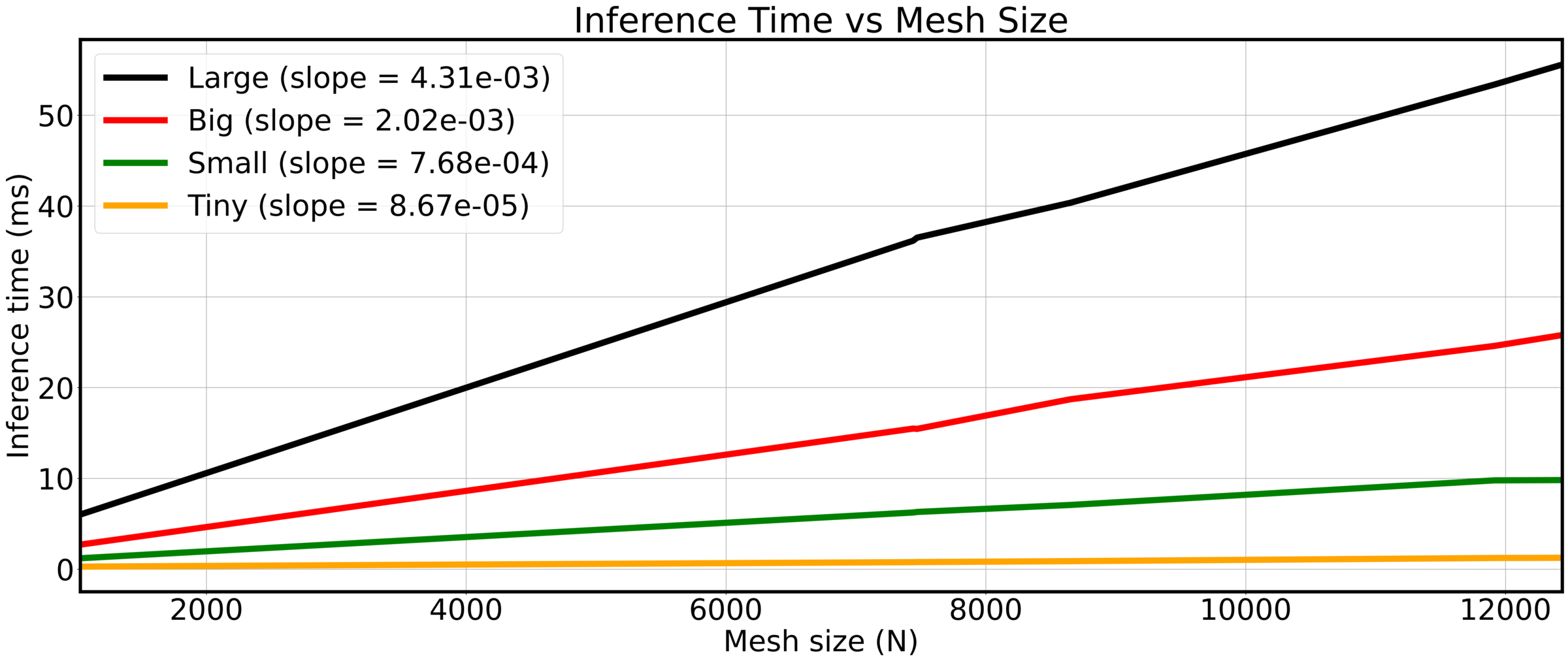}
    \caption{Inference times plotted against various mesh sizes apparent in the base and diseased datasets, highlighting the model's $\mathcal{O}(N)$ complexity.}
    \label{fig:inference_v_mesh}
\end{figure}

\begin{figure}[H]
    \centering
    \includegraphics[width=1\linewidth]{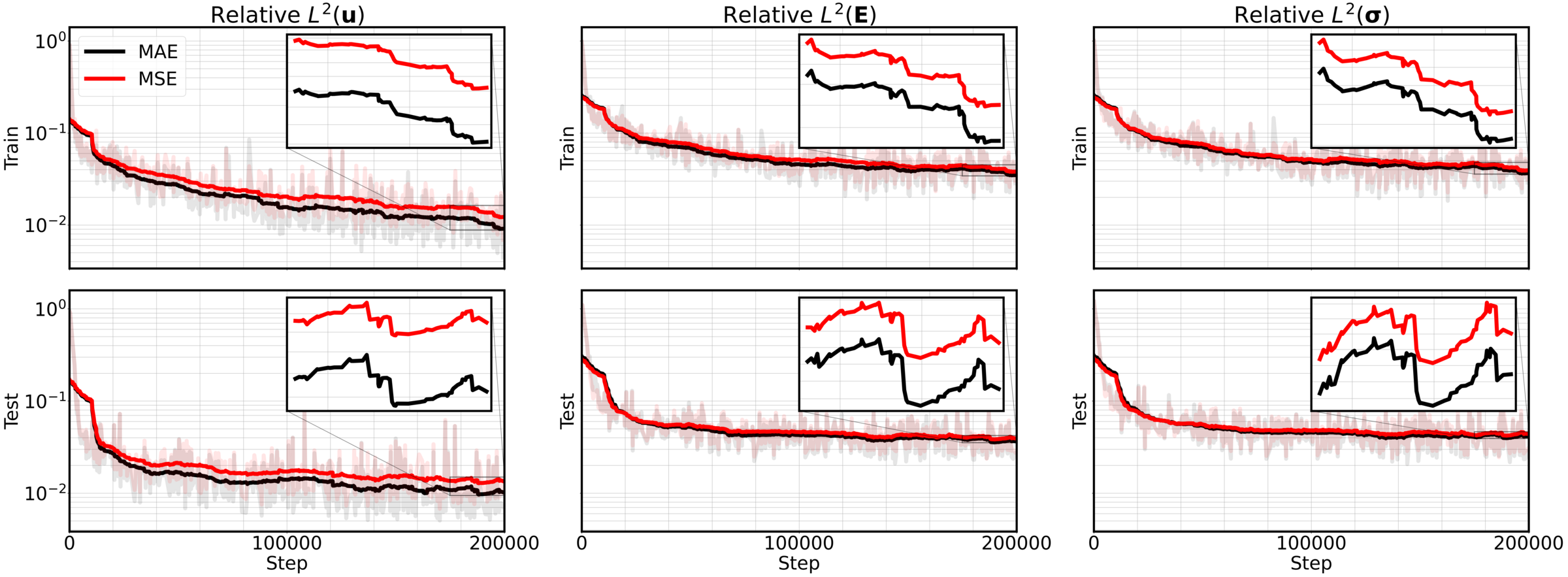}
    \caption{Comparison of relative $L^2$ error convergence over training steps using a MAE and MSE loss functions for the in-distribution setting. Errors are plotted as fractions rather than percentages (10$^{-2}$ corresponds to 1\%).}
    \label{fig:mae_v_mse}
\end{figure}

\begin{figure}[H]
    \centering
    \includegraphics[width=1\linewidth]{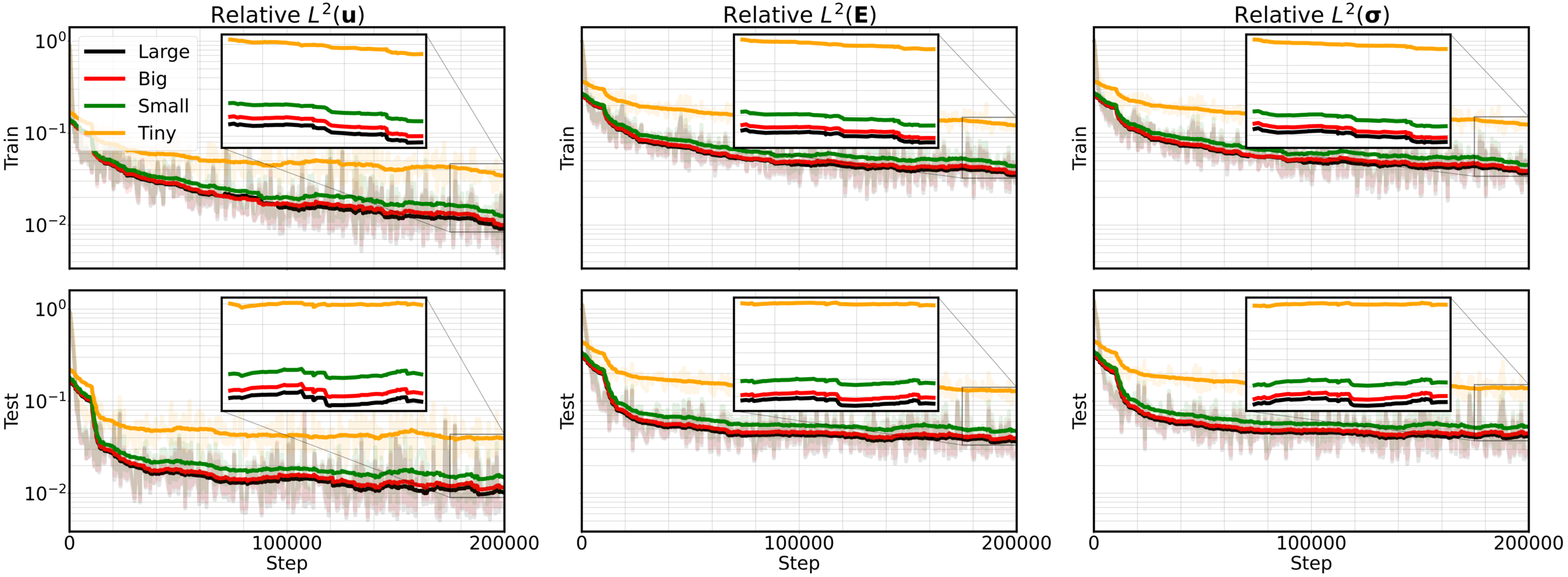}
    \caption{Comparison of relative $L^2$ error convergence over training steps using different model sizes for the in-distribution setting. Errors are plotted as fractions rather than percentages (10$^{-2}$ corresponds to 1\%).}
    \label{fig:model_size_loss}
\end{figure}

\begin{figure}[H]
    \centering
    \includegraphics[width=1\linewidth]{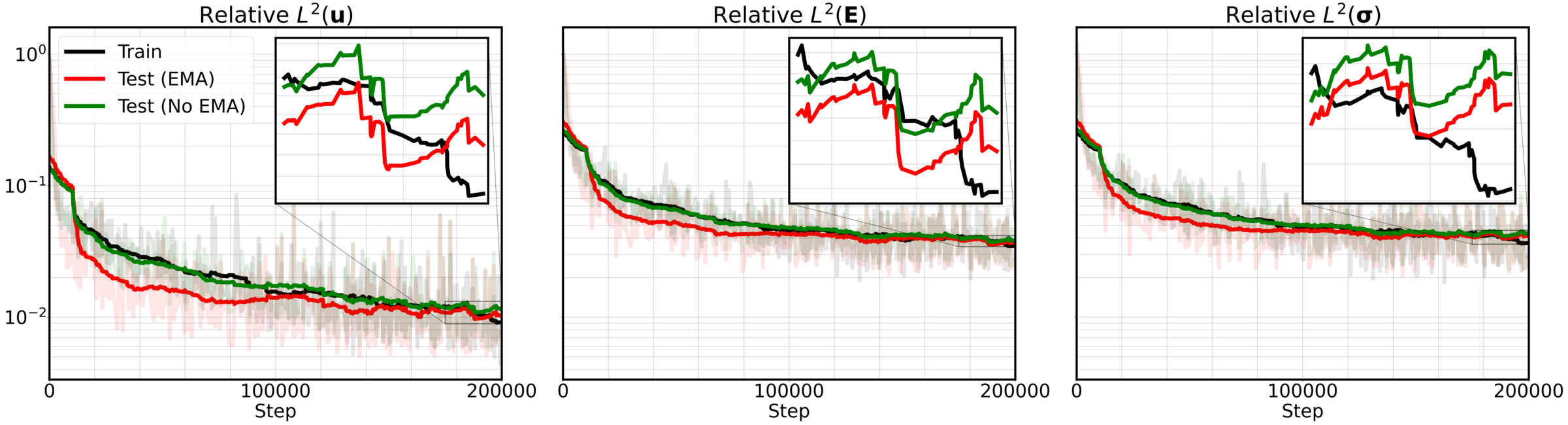}
    \caption{Comparison of relative $L^2$ error convergence over training steps using the EMA and non-EMA parameters for test evaluation for the in-distribution setting. Errors are plotted as fractions rather than percentages (10$^{-2}$ corresponds to 1\%).}
    \label{fig:no_ema_v_ema}
\end{figure}

\begin{figure}[H]
    \centering
    \includegraphics[width=1\linewidth]{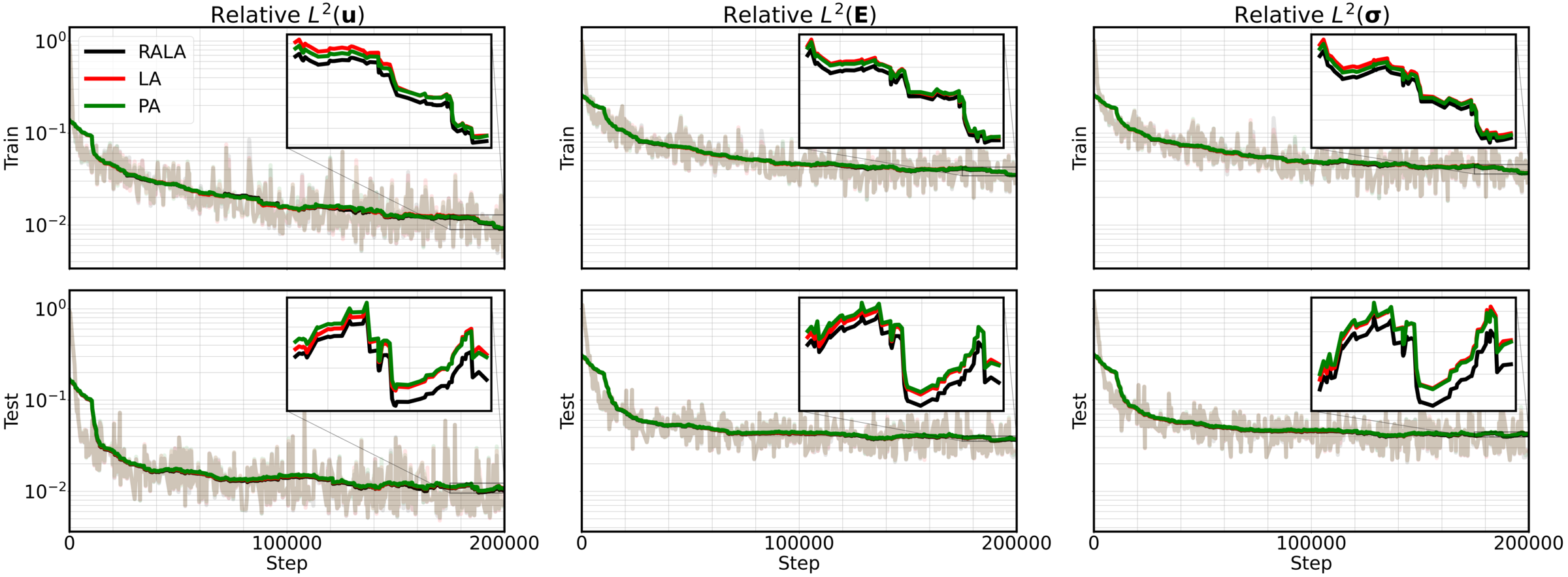}
    \caption{Comparison of relative $L^2$ error convergence over training steps using Linear Attention (LA) \cite{katharopoulos2020transformers}, Physics-Attention \cite{wu2024transolver}, and RALA \cite{fan2025breaking} in the encoder for the in-distribution setting. Errors are plotted as fractions rather than percentages (10$^{-2}$ corresponds to 1\%).}
    \label{fig:rala_v_la}
\end{figure}

\begin{figure}[H]
    \centering
    \includegraphics[width=1\linewidth]{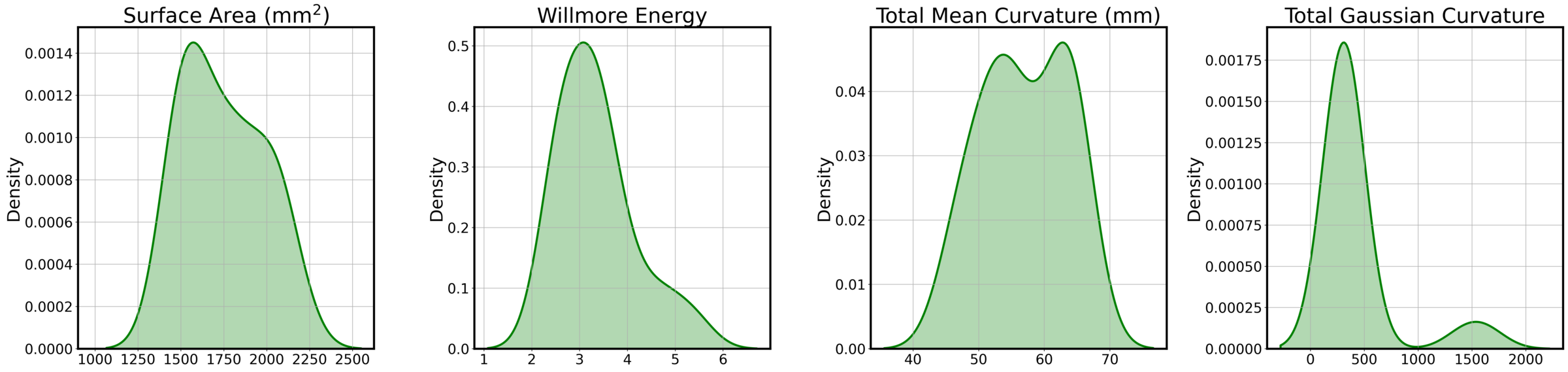}
    \caption{Distribution of global shape conditioning variables in the base dataset. The secondary modes correspond to the two finely meshed geometries.}
    \label{fig:geometric_dist}
\end{figure}

\begin{table}[H]
\centering
\begin{tabular}{c cc cc}
\toprule
\multirow{2}{*}{\textbf{Geometry ID}} 
& \multicolumn{2}{c}{\textbf{Diverse Geometry}} 
& \multicolumn{2}{c}{\textbf{Moving Boundary}} \\
\cmidrule(lr){2-3}\cmidrule(lr){4-5}
& \textbf{Mitral} & \textbf{Tricuspid} & \textbf{Mitral} & \textbf{Tricuspid} \\
\midrule
$1$  & $9{,}943$  & $3{,}883$ & $8{,}567$ & $4{,}572$ \\
$2$  & $7{,}811$  & $5{,}804$ & $8{,}163$ & $4{,}311$ \\
$3$  & $4{,}357$  & $8{,}235$ & $2{,}392$ & $5{,}073$ \\
$4$  & $9{,}248$  & $9{,}610$ & $7{,}977$ & $8{,}894$ \\
$5$  & $10{,}021$ & $8{,}795$ & $357$     & $8{,}516$ \\
$6$  & $4{,}854$  & $7{,}680$ & $3{,}602$ & $7{,}606$ \\
$7$  & $5{,}961$  & $9{,}721$ & $7{,}612$ & $7{,}997$ \\
$8$  & $8{,}061$  & $9{,}108$ & $7{,}280$ & $8{,}194$ \\
$9$  & $3{,}875$  & $4{,}018$ & $3{,}414$ & $5{,}003$ \\
$10$ & $5{,}426$  & $5{,}691$ & $4{,}923$ & $292$     \\
\bottomrule
\end{tabular}
\caption{Per-geometry counts for the Diverse Geometry and Moving Boundary datasets.}
\label{tab:geometry_counts}
\end{table}

\begin{table}[H]
\centering
\begin{tabular}{lllccc}
\toprule
\textbf{Dataset} & \textbf{Setting} & \textbf{Split} & Rel. $L^2(\boldsymbol{u})$ & Rel. $L^2(\boldsymbol{E})$ & Rel. $L^2(\boldsymbol{\sigma})$ \\
\midrule
\multirow{11}{*}{Base}
 & \multirow{2}{*}{In-distribution}
   & Train & 1.01 & 3.60 & 3.82 \\
 & & Test  & 1.02 & 3.63 & 3.86 \\
\cmidrule(lr){2-6}
 & \multirow{2}{*}{SBP Generalization}
   & Train & 0.94 & 3.41 & 3.67 \\
 & & Test  & 3.33 & 7.85 & 7.98 \\
\cmidrule(lr){2-6}
 & \multirow{4}{*}{\makecell[l]{Material Parameter \\ Generalization}}
   & Train      & 0.96 & 3.56 & 3.83 \\
 & & Test (Uni) & 1.48 & 4.51 & 4.69 \\
 & & Test (Bi)  & 2.22 & 5.88 & 6.06 \\
 & & Test (Tri) & 3.72 & 9.11 & 9.07 \\
\cmidrule(lr){2-6}
 & \multirow{2}{*}{\makecell[l]{Joint SBP and Material \\ Parameter Generalization}}
   & Train & 1.07 & 3.79 & 4.06 \\
 & & Test  & 4.68 & 10.34 & 11.85 \\
\midrule
\multirow{4}{*}{Diseased}
 & \multirow{2}{*}{In-distribution}
   & Train & 0.28 & 1.08 & 1.26 \\
 & & Test  & 0.30 & 1.14 & 1.31 \\
\cmidrule(lr){2-6}
 & \multirow{2}{*}{\makecell[l]{Joint SBP and Material \\ Parameter Generalization}}
   & Train & 0.72 & 2.77 & 3.00 \\
 & & Test  & 3.80 & 11.19 & 12.38 \\
\bottomrule
\end{tabular}
\caption{Average relative $L^2$ errors (\%) across generalization settings on the base and diseased datasets, computed over all samples without the 1\% trimming used in the main text (compare Table~1). Material parameter generalization cases (Uni, Bi, Tri) share identical training data.}
\label{tab:results_summary2}
\end{table}

\end{document}